\documentclass[10pt,twocolumn,letterpaper]{article}

\usepackage{iccv}
\usepackage{times}
\usepackage{epsfig}
\usepackage{graphicx}
\usepackage{amsmath}
\usepackage{amssymb}
\usepackage{amsthm}
\usepackage{mathrsfs}
\usepackage{multirow}
\usepackage{enumerate}
\usepackage{subfigure}
\usepackage{amssymb}
\usepackage{mathrsfs}
\usepackage{amsmath}
\usepackage{booktabs}

\usepackage{color}
\usepackage[accsupp]{axessibility}

\newtheorem{theorem}{Theorem}
\newtheorem{lemma}{Lemma}
\newtheorem{definition}{Definition}

\usepackage[breaklinks=true,bookmarks=false]{hyperref}

\iccvfinalcopy % *** Uncomment this line for the final submission

\def\iccvPaperID{6058} % *** Enter the ICCV Paper ID here

\ificcvfinal\fi

\begin{document}

%%%%%%%%% TITLE

\title{Learning with Noisy Labels via Sparse Regularization}
\author{Xiong Zhou$^{1,2}$ \  Xianming Liu$^{1,2}$\thanks{Correspondence to: Xianming Liu (csxm@hit.edu.cn)}\ \ \ Chenyang Wang$^{1}$ \  Deming Zhai$^{1}$ \  Junjun Jiang$^{1,2}$ \  Xiangyang Ji$^{3}$\\
$^1$Harbin Institute of Technology\quad $^2$Peng Cheng Laboratory\quad $^3$Tsinghua University\\
{\tt\small \{\href{mailto:cszx@hit.edu.cn}{cszx},\href{mailto:csxm@hit.edu.cn}{csxm},\href{mailto:cswcy@hit.edu.cn}{cswcy},\href{mailto:zhaideming@hit.edu.cn}{zhaideming},\href{mailto:junjunjiang@hit.edu.cn}{junjunjiang}\}@hit.edu.cn}\quad \tt\small \href{mailto:xyji@tsinghua.edu.cn}{xyji@tsinghua.edu.cn}

}
% \author{Xiong Zhou \  Xianming Liu \  Chenyang Wang \  Deming Zhai \  Junjun Jiang \  Xiangyang Ji\\
% Harbin Institute of Technology\\
% \and
% Xianming Liu\\
% Harbin Institute of Technology\\
% \and
% Chenyang Wang\\
% Harbin Institute of Technology

% }

\maketitle
\ificcvfinal\thispagestyle{empty}\fi

%%%%%%%%% ABSTRACT
\begin{abstract}
    Learning with noisy labels is an important and challenging task for training accurate deep neural networks. Some commonly-used loss functions, such as Cross Entropy (CE), suffer from severe overfitting to noisy labels. Robust loss functions that satisfy the symmetric condition were tailored to remedy this problem, which however encounter the underfitting effect. In this paper, we theoretically prove that \textbf{any loss can be made robust to noisy labels} by restricting the network output to the set of permutations over a fixed vector. When the fixed vector is one-hot, we only need to constrain the output to be one-hot, which however produces zero gradients almost everywhere and thus makes gradient-based optimization difficult. In this work, we introduce the sparse regularization strategy to approximate the one-hot constraint, which is composed of network output sharpening operation that enforces the output distribution of a network to be sharp and the $\ell_p$-norm ($p\le 1$) regularization that promotes the network output to be sparse. This simple approach guarantees the robustness of arbitrary loss functions while not hindering the fitting ability. Experimental results demonstrate that our method can significantly improve the performance of commonly-used loss functions in the presence of noisy labels and class imbalance, and outperform the state-of-the-art methods. The code is available at \href{https://github.com/hitcszx/lnl_sr}{\textcolor{magenta}{https://github.com/hitcszx/lnl\_sr}}.
\end{abstract}

%%%%%%%%% BODY TEXT
\section{Introduction}
Deep neural networks (DNNs) have achieved remarkable success on various computer vision tasks, such as image classification, segmentation, and object detection \cite{goodfellow2016deep}. The most widely used paradigm for DNN training is the end-to-end supervised manner, whose performance largely relies on massive high-quality annotated data. However, collecting large-scale datasets with fully precise annotations (or called clean labels) is usually expensive and time-consuming, and sometimes even impossible. Noisy labels, which are systematically corrupted from ground-truth labels, are ubiquitous in many real-world applications, such as online queries \cite{liu2011noise}, crowdsourcing \cite{arpit2017closer}, adversarial attacks \cite{zhou2020dast}, and medical images analysis \cite{KARIMI2020101759}. On the other hand, it is well-known that over-parameterized neural networks have enough capacity to memorize large-scale data with even completely random labels, leading to poor performance in generalization  \cite{2017Understanding, arpit2017closer, jiang2018mentornet}. Therefore, robust learning with noisy labels has become an important and challenging task in computer vision \cite{2020Learning, Han2020ASO, KARIMI2020101759, zhou2021asymmetric}.
\begin{figure}[t]
    \centering
    \subfigure[FL]{
        \label{CE}
        \includegraphics[scale=0.125]{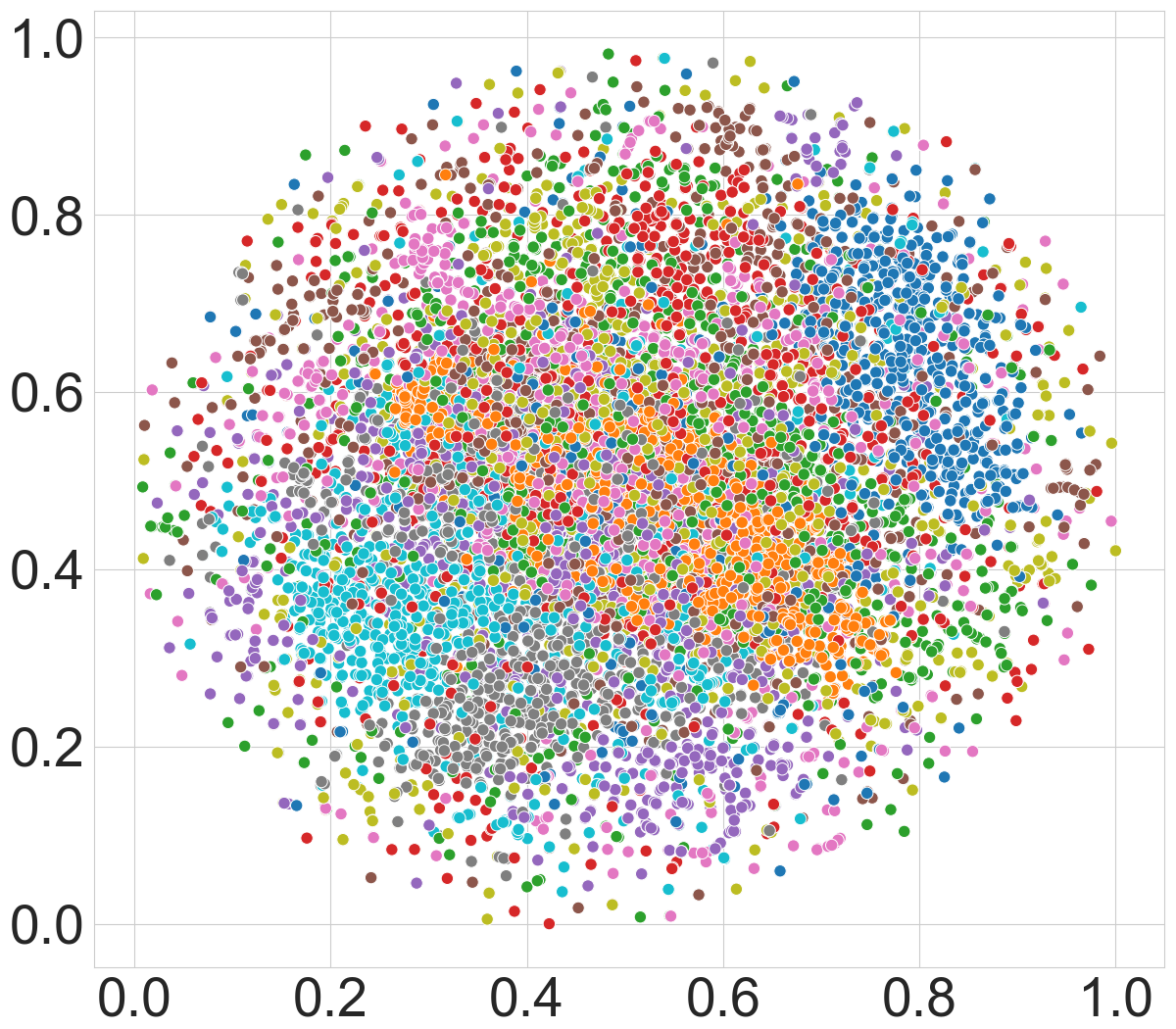}
    }
    \subfigure[FL+SR]{
        \label{CE+SR}
        \includegraphics[scale=0.125]{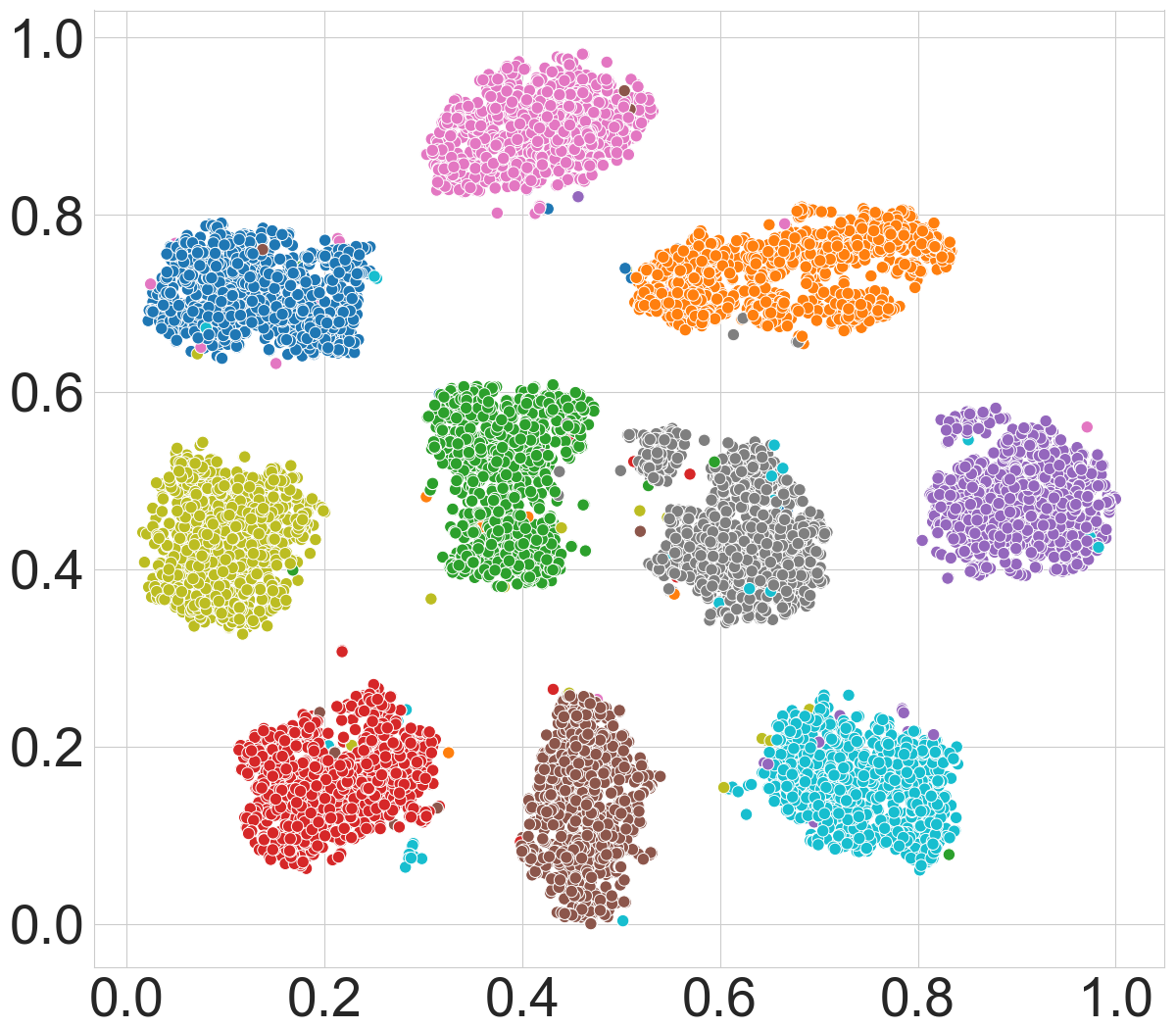}
    }
    \\
    \subfigure[GCE]{
        \label{GCE}
        \includegraphics[scale=0.125]{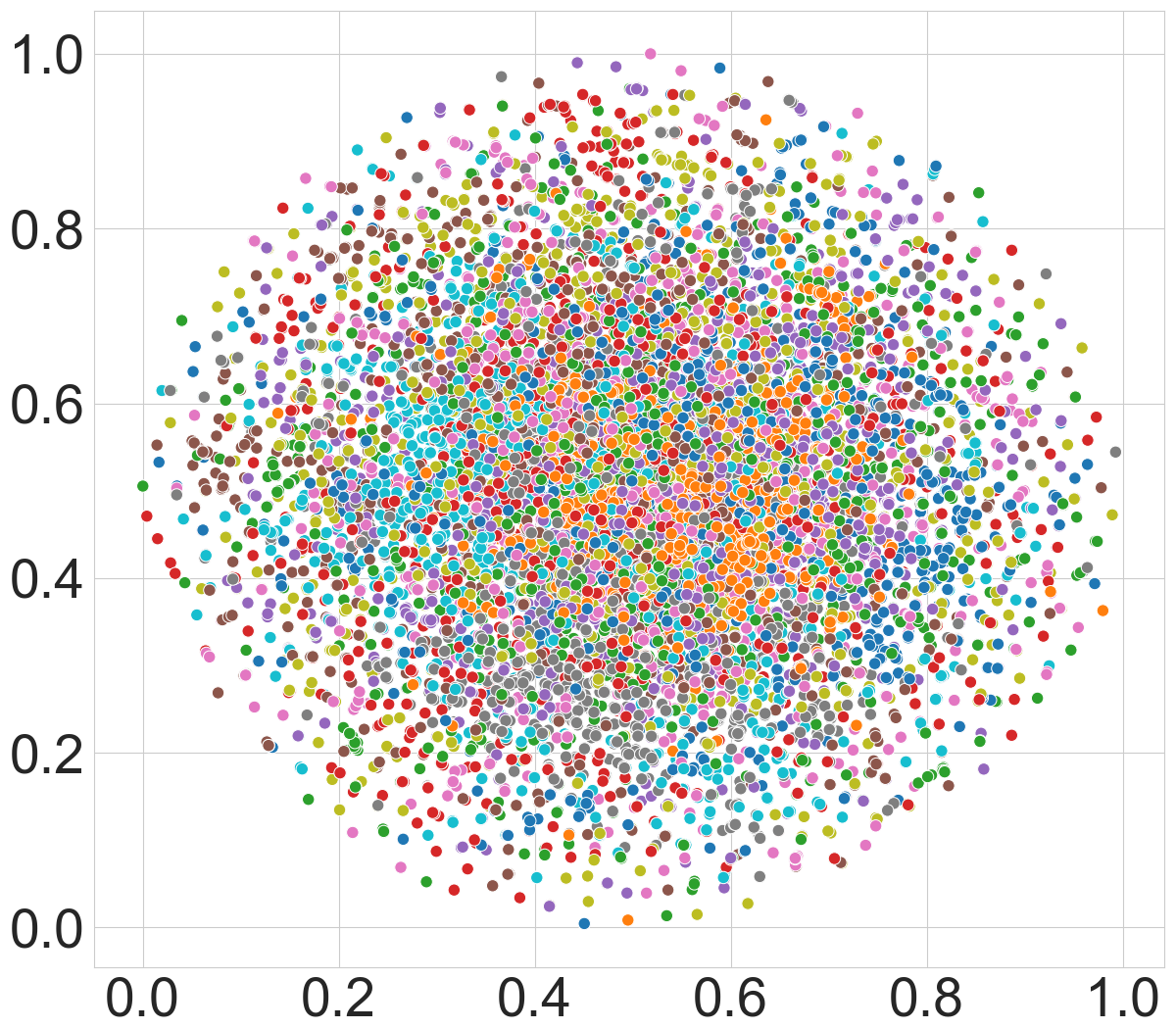}
    }
    \subfigure[GCE+SR]{
        \label{GCE+SR}
        \includegraphics[scale=0.125]{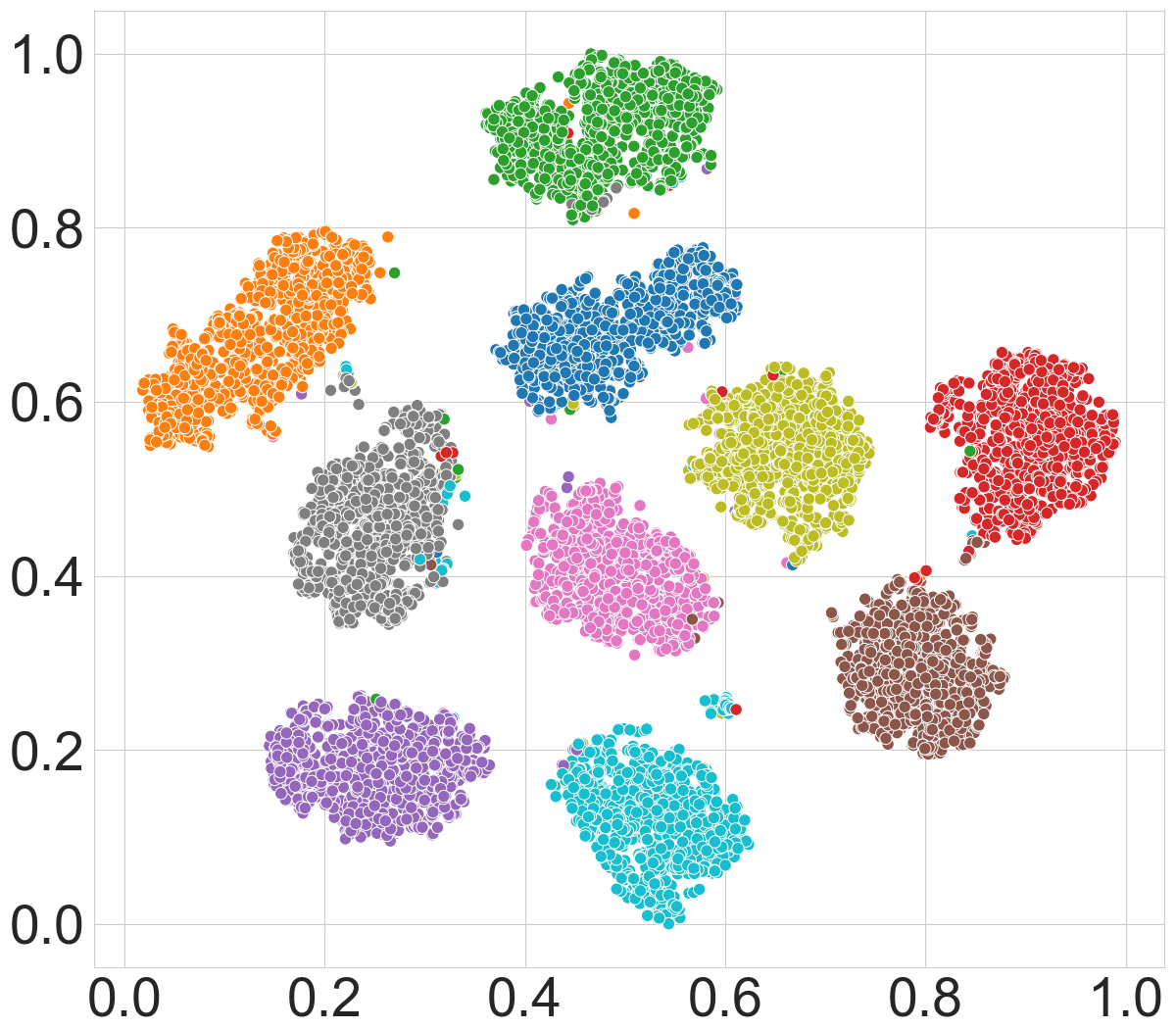}
    }
    \caption{Visualization of learned representations on MNIST with 0.8 symmetric label noise. The representations learned by the proposed sparse regularization (SR)-enhanced methods are significantly discriminative than that learned by original losses, which are with more separated and clearly bound margin.}
    \label{tsne-mnist}
    \vskip-5pt
\end{figure}

To prevent over-fitting to mislabeled data, many strategies have been presented in the literature, among which robust loss function design is one of the most popular approaches since it enjoys simplicity and universality. Ghosh \textit{et al.} \cite{symmetric} theoretically proved that a loss function would be inherently tolerant to symmetric label noise as long as it satisfies the symmetric condition. However, the derived loss functions according to this design principle, such as MAE \cite{symmetric} and Reverse Cross Entropy (RCE) \cite{SCE}, suffer from the underfitting effect on complicated datasets \cite{BERAUC, ma2020normalized}. As demonstrated in \cite{GCE}, the robustness of MAE can concurrently cause increased difficulty in training, leading to performance drops. On the other hand, the commonly-used CE and focal loss (FL) \cite{lin2017focal} enjoy the advantage of sufficient learning where the optimizer puts more emphasis on ambiguous samples, but they tend to overfit on noisy labels. 

How to achieve robustness and learning sufficiency simultaneously? This question has motivated a large amount of work to design new loss functions that are robust to noisy labels as well as easy to fit clean ones. 
For instance, Zhang \textit{et al.} proposed a generalization of cross entropy (GCE) \cite{GCE}, which behaves like a generalized mixture of MAE and CE. Wang \textit{et al.} proposed the Symmetric Cross Entropy (SCE) loss \cite{SCE}, which combines RCE with CE. However, both GCE and SCE just perform the trade-off between symmetric loss and CE, which are only partially robust to noisy labels. Ma \textit{et al.} \cite{ma2020normalized} theoretically proved that by applying a simple normalization, any loss can be made robust to noisy labels. However, the normalization operation actually changes the form of loss functions, which no longer 
preserve the original fitting ability. The authors further proposed the Active Passive Loss (APL) to remedy this problem, which is the combination of two symmetric losses.

As reviewed above, all existing methods attempt to design new loss functions that satisfy the symmetric condition  \cite{symmetric} to achieve robustness and achieve learning sufficiency meanwhile by combining multiple forms of loss function, such as symmetric loss and CE in \cite{GCE,SCE} and two symmetric losses in \cite{ma2020normalized}. In this work, we propose a novel perspective to understand the symmetric condition, and prove in theory that any loss can be made robust through
restricting the hypothesis class. Specifically, we demonstrate that label noise under risk minimization can be mitigated by restricting the network output to a permutation set of a fixed one-hot vector instead of modifying the loss function.  This discrete process, however would result in many zero gradients, making the gradient-based optimization difficult.  We then propose a sparse regularization strategy to approximate the one-hot constraint, which includes network output sharpening operation and $\ell_p$-norm ($p\le 1$) regularization in risk minimization. Experimental results on synthesis and real-world datasets demonstrate that our method can significantly improve the performance of commonly-used loss functions in the presence of noisy labels, as illustrated in Fig. \ref{tsne-mnist}, and outperform the state-of-the-art robust loss functions. Moreover, we evaluate the sparse regularization strategy on the long-tailed and step-imbalanced image classification, which demonstrates that SR can also mitigate class imbalance well.

The main contributions of our work are highlighted as follows:
\begin{itemize}
    \item To the best of our knowledge, we are the first work in the literature to meet the symmetric condition by restricting the hypothesis class. It offers a novel perspective of understanding the symmetric condition and an alternative approach for robust learning. 
    \vskip-3pt
    \item We theoretically prove that any loss function can be made robust to noisy labels by restricting the network output to the set of permutations over a fixed vector.
    \vskip-3pt
    \item We propose a simple but effective approach for robust training through sparse regularization.
    \vskip-3pt
    \item  We provide a principled approach to simultaneously achieve robust training and preserve the fitting ability of commonly-used losses such as CE.
\end{itemize}

\section{The Proposed Method}
In this section, we first introduce some preliminaries about robust learning. Subsequently, we present our finding in theory that any loss function can achieve noise tolerance via network output permutation. Furthermore, we offer a simple but effective approach for robust learning by introducing sparse regularization in network training. Finally, we provide analysis about the merit of our scheme---achieve better tradeoff between robustness and sufficient learning.

\subsection{Preliminaries}
\noindent\textbf{Risk Minimization.}\quad Assume $\mathcal X\subset \mathbb R^d$ is the feature space from which the examples are drawn, and $\mathcal Y=[k]=\{1,...,k\}$ is the class label space, \textit{i.e.}, we consider a $k$-classification problem. In a typical classifier learning problem, we are given a training set, $\mathcal S=\{(\mathbf x_1,y_1),...,(\mathbf x_{N}, y_{N})\}$, where $(\mathbf x_i, y_i)$ is drawn i.i.d. according to an unknown distribution, $\mathcal D$, over $\mathcal X\times\mathcal Y$. The classifier is a mapping function from feature space to label space $h(\mathbf x)=\mathop{\text{arg}}\max_i f(\mathbf x)_i$, where $f:\mathcal X\rightarrow \mathcal C$ denotes an approximation of $p(\cdot|\mathbf x)$, and $\mathcal C\subset [0,1]^k$, $\forall \mathbf c\in\mathcal C$, $\mathbf 1^T \mathbf c=1$. In deep learning, $f(\mathbf x)$ is usually modeled by a neural network ending with a softmax layer.

The loss function is defined as a mapping $L:\mathcal C\times \mathcal Y\rightarrow \mathbb{R}^+$, where $\arg\min_{\mathbf u\in \mathcal C}L(\mathbf u,y)=\mathbf{e}_y$, and $\mathbf e_y$ denotes the one-hot vector. In this work, we consider the loss functional $\mathcal L$, where $\forall L\in \mathcal L$, $L(\mathbf u, i)=\ell_1(u_i)+\sum_{j\not =i}\ell_2(u_j)$, and $\ell_1$ as well as $\ell_2$ are two basic functions. For example, CE can be expressed by $\ell_1(u_i)=-\log u_i$ and $\ell_2(u_j)=0$. Given any loss function $L$, and a classifier $f$, the $L$-risk of $f$ is
\begin{equation}
\label{L-risk}
    R_L(f)=\mathbb{E}_{\mathcal D}[L(f(\mathbf x),y)]=\mathbb E_{\mathbf x,y}[L(f(\mathbf x), y)],
\end{equation}
where $\mathbb{E}$ denotes expectation. Under the risk minimization framework, the objective is to learn an optimal classifier, $f^*$, which is a global minimum of $R_L(f)$.

\noindent\textbf{Noise Tolerance.}\quad We define the noise corruption process as that a clean label $y$ is flipped into a noisy version $\tilde{y}$ with probability $\eta_{\mathbf x, \tilde{y}}=p(\tilde{y}|y,\mathbf x)$ (more label noise settings can be found in \cite{symmetric}). The corresponding noisy $L$-risk is 
\vskip-5pt
\begin{equation}
\notag
    R_L^\eta(f)=\mathbb{E}_{\mathcal D}[(1-\eta_{\mathbf x})L(f(\mathbf x), y)+\sum_{i\not =y}\eta_{\mathbf x,i} L(f(\mathbf x),i)],
\end{equation}
\vskip-5pt
where $\eta_{\mathbf x}=\sum_{i\not=y}\eta_{\mathbf x, i}$ denotes the noise rate. Risk minimization under a given loss function is noise-tolerant if  $R_L^\eta(f)$ shares the same global minimum as $R_L(f)$.

\noindent\textbf{Symmetric Loss Functions.}\quad 
A symmetric loss function \cite{Manwani, symmetric} is proved to be noise-tolerant for a $k$-class classification under symmetric noise if the noise rate $\eta<\frac{k-1}{k}$ and the loss function satisfies
\vskip-5pt
\begin{equation}
\label{symmetric-condition}
    \sum_{i=1}^k L(f(\mathbf x), i)=C,\ \forall \mathbf x\in\mathcal X, \forall f\in\mathcal H,
\end{equation}
where $C$ is a constant, and $\mathcal H$ is the hypothesis class.

The symmetric condition (\ref{symmetric-condition}) stated above guarantees the noise tolerance by risk minimization on a symmetric loss function, \textit{i.e.}, the classifier trained in noisy case has the same misclassification probability as that trained in noise-free case under the specified assumption. Moreover, if $R_L(f^*)=0$, $L$ is also noise-tolerant under an asymmetric noise, where $f^*$ is a global risk minimum of $R_L$. 

\subsection{Noise Tolerance via Output Permutation}
The symmetric condition (\ref{symmetric-condition}) theoretically guarantees that a symmetric loss can lead to robust training. However, the derived loss functions according to this design principle usually suffer from underfitting \cite{SCE, BERAUC, ma2020normalized}. On the other hand, the existing methods all pay attention to designing new loss functions, but never attempt to restrict the hypothesis class $\mathcal H$ to satisfy the symmetric condition. In this work, we propose to restrict the hypothesis class $\mathcal H$ such that any losses satisfy the symmetric condition (\ref{symmetric-condition}) and thus become robust to label noise. Furthermore, we provide a theoretical analysis to demonstrate the noise tolerance of our scheme. The proofs can be seen in the supplementary materials.

First of all, we provide the definition of permutation operation, which plays an important role in our derivation.
\begin{definition}
\label{permutation}
For a vector $\mathbf v\in\mathbb{R}^k$, the permutation operation on it is defined as \cite{pmlr-v119-prillo20a}:
\begin{equation}
    \mathbf v_{\pi}= P_{\pi}\mathbf v,
\end{equation}
where $P_{\pi}=[\mathbf e_{\pi_1},\mathbf e_{\pi_2},...,\mathbf e_{\pi_k}]^T$ is the permutation matrix, and $\{\pi_1,\pi_2,...,\pi_k\}=[k]$. 
\end{definition}
According to this definition, it is easy to find that $\mathbf v$ and $\mathbf v_{\pi}$ share the same space, \textit{i.e.},
% \begin{equation}
% \label{equal}
    $\{v_{\pi_1},v_{\pi_2},...,v_{\pi_k}\}=\{v_1,v_2,...,v_k\}$.
For instance, when $k=3$, $\mathbf v = \{v_1,v_2,v_3\}$ and $\pi=[3,1,2]$, then $P_{\pi}=[\mathbf e_3,\mathbf e_1,\mathbf e_2]^T$, and the vector after permutation operation is $\mathbf v_{\pi}=(v_3,v_1,v_2)^T$. 

More generally, let $\mathcal P_{\mathbf v}$ denote the permutation set over $\mathbf v$, we have
\vskip-5pt
\begin{equation}
\sum_{i=1}^k\ell(u_i)=\sum_{i=1}^k \ell(v_i), \forall\ \mathbf u\in\mathcal P_{\mathbf v}.
\end{equation}

According to the above definition and derivation, we arrive at Lemma \ref{symmetric-lemma}:
\begin{lemma}
\label{symmetric-lemma}
Given a vector $\mathbf v$, $\forall L \in \mathcal L$, we have
\vskip-5pt
\begin{equation}
    \sum_{i=1}^k L(\mathbf u,i)=C,\quad \forall\ \mathbf u\in\mathcal P_{\mathbf v},
\end{equation}
where $C=\sum_{i=1}^k L(\mathbf v, i)$ is a constant when $\mathbf v$ is fixed.
\end{lemma}

Lemma \ref{symmetric-lemma} indicates that, when the network output $\mathbf u$ is restricted to belong to a permutation set $\mathcal P_{\mathbf v}$ of a fixed vector $\mathbf v$, any loss functions in $\mathcal L$ satisfy the symmetric condition.
We further have the following theorems for symmetric and asymmetric noise that can be proved similarly as \cite{symmetric}:

\begin{theorem}[Noise tolerance under symmetric noise]
\label{symmetric-robust}
In a multi-class classification problem, $\forall L\in\mathcal L$, $L$ is noise-tolerant under symmetric label noise if $\eta<1-\frac{1}{k}$ and $f:\mathcal X\rightarrow \mathcal P_{\mathbf v}$, i.e.,
\vskip-5pt
\begin{equation}
    \mathop{\arg\min}_{f:\mathcal X\rightarrow \mathcal P_{\mathbf v}} R_{L}(f)=\mathop{\arg\min}_{f:\mathcal X\rightarrow \mathcal P_{\mathbf v}} R^\eta_{L}(f),
\end{equation}
where $\mathbf v$ is a fixed vector.
\end{theorem}

\begin{theorem}[Noise tolerance under asymmetric noise]
\label{asymmetric-robust}
In a multi-class classification problem, let $f:\mathcal X\rightarrow \mathcal P_{\mathbf v}$, where $\mathbf v$ is a fixed vector, and suppose $L\in\mathcal L$ satisfy $0\le L(f(\mathbf x),i)\le \frac{C}{k-1}$, $\forall i\in[k]$. If $R_L(f^*)=0$, then $L$ is noise-tolerant under asymmetric or class-conditional noise when $\eta_{y,i}<1-\eta_y$ with $\sum_{k\not =y}\eta_{y,i}=\eta_y$, $\forall \mathbf x$.
\end{theorem}

Theorem \ref{symmetric-robust} and \ref{asymmetric-robust} inspire us that label noise under risk minimization can be mitigated by restricting the network output to a permutation set instead of changing the loss function. This offers an alternative principle approach to achieve robust learning. However, the optimization is non-trivial when using gradient-based strategy because the constraint that $\mathcal P_{\mathbf v}$ is a discrete mapping produces many zero gradients. Instead, we turn to approximate the constraint by relaxing the output restriction of the hypothesis class with an error bound $\varepsilon$, \textit{i.e.}, $\mathcal H_{\mathbf v,\varepsilon}=\{f:\min_{\mathbf u\in\mathcal P_{\mathbf v}}\|f(\mathbf x)-\mathbf u\|_2\le \varepsilon, \forall \mathbf x\}$. We can derive the risk bound as follows:
\begin{theorem}
\label{risk-bound}
In a multi-class classification problem, if the loss function $L\in \mathcal L$ satisfies $|\sum_{i=1}^k(L(\mathbf u_1,i)-L(\mathbf u_2,i))|\le \delta$ when $\| \mathbf u_1-\mathbf u_2\|_2\le \varepsilon$, and $\delta \rightarrow 0$ as $\varepsilon\rightarrow 0$, then for symmetric label noise satisfying $\eta<1-\frac{1}{k}$, the risk bound for $f\in\mathcal H_{\mathbf v,\varepsilon}$ can be expressed as
$$
    R_L(f^*_\eta)-R_L(f^*)\le 2c\delta, 
$$
where $c=\frac{\eta}{(1-\eta)k-1}$, $f^*_\eta$ and $f^*$ denote the global minimum of $R_L^\eta(f)$ and $R_L(f)$, respectively.
\end{theorem}

Theorem \ref{risk-bound} indicates that when restricting the output of the network to belong to $\mathcal H_{\mathbf v, \varepsilon}$, the noisy minimum $f^*_{\eta}$, compared to the clean minimum $f^*$, has a risk error bound $2c\delta$. And when $\varepsilon \rightarrow 0$, the bound also tends to $0$. This implies that, by shrinking $\mathcal H_{\mathbf v,\varepsilon}$,  $f^*_\eta=\arg\min_{f\in\mathcal H_{\mathbf v,\varepsilon}}R_L^\eta(f)$ converges to $f^*$.

\subsection{Robust Learning via Sparse Regularization}
Upon the above theoretical analysis, we propose a simple but effective approach for robust learning, which can make any losses robust to noisy labels by introducing sparse regularization on network output. Specifically, we consider the fixed vector $\mathbf v$ as a one-hot vector, \textit{i.e.}, we restrict the network output to one-hot vectors $\mathcal P_{0-1}$. This discrete process would result in many zero gradients, making the optimization difficult. To approximate the one-hot constraint, we propose the sparse regularization strategy, which is composed of two modules: network output sharpening and $\ell_p$-norm ($p\le 1$) regularization in risk minimization.

\noindent\textbf{Network Output Sharpening.} The output sharpening module is to make the network output closer to a one-hot vector. One popular way to approximate a one-hot vector by the continuous mapping is to use a temperature-dependent softmax function, \textit{i.e.},
\begin{equation}
    \sigma_{\tau}(\mathbf z)_i = \frac{\exp(z_i/\tau)}{\sum_{j=1}^k \exp(z_j/\tau)},
\end{equation}
where $0\le\tau\le 1$, and $\sigma_\tau(\mathbf z)_i$ is a point in the probability simplex. Note that, in the limit situation where $\tau\rightarrow 0$, $\sigma_\tau(\mathbf z)$ converges to a one-hot vector. In other words, with low temperatures, the distribution spends essentially all of its probability mass in the most probable state. Meanwhile, we may limit the value of $z_i$ to the range $[-1, 1]$ by performing $\ell_2$ normalization before the output sharpening to prevent it from trivial scaling solution.
% the numerical overflow when the network produces a particularly large value.

\noindent\textbf{$\ell_p$-norm Regularization}. We further introduce $\ell_p$-norm regularization into risk minimization to promote the sparsity of network output. Specifically, we perform the following constrained risk minimization in network training:
\vskip-5pt
\begin{equation}
    \min_{f\in \mathcal H} R_L(f)\quad\text{s.t.}\quad \|f\|_p \le \gamma,
\end{equation}
where $0\le p\le1$, and $\gamma$ is an appropriately selected parameter. In practice, we can convert to train a neural network by minimizing the following form:
\vskip-5pt
\begin{equation}
    \label{norm}
    \sum_{i=1}^N L(f(\mathbf x_i), \tilde{y}_i) + \lambda \|f(\mathbf x_i)\|_p^p.
\end{equation}

It is worth noting that, if we define $L(\mathbf u,i)=-\alpha\log u_i - \lambda u_i$ and $p=1$, then Eq. (\ref{norm}) is equivalent to the Symmetric Cross Entropy (SCE) loss \cite{SCE}. Thus, SCE can be regarded as a special case of a loss function with the $\ell_p$-norm regularization.

\begin{figure}
    \centering
    \subfigure[CE+SR]{
        \label{lambda-CE}
        \includegraphics[width=1.5in]{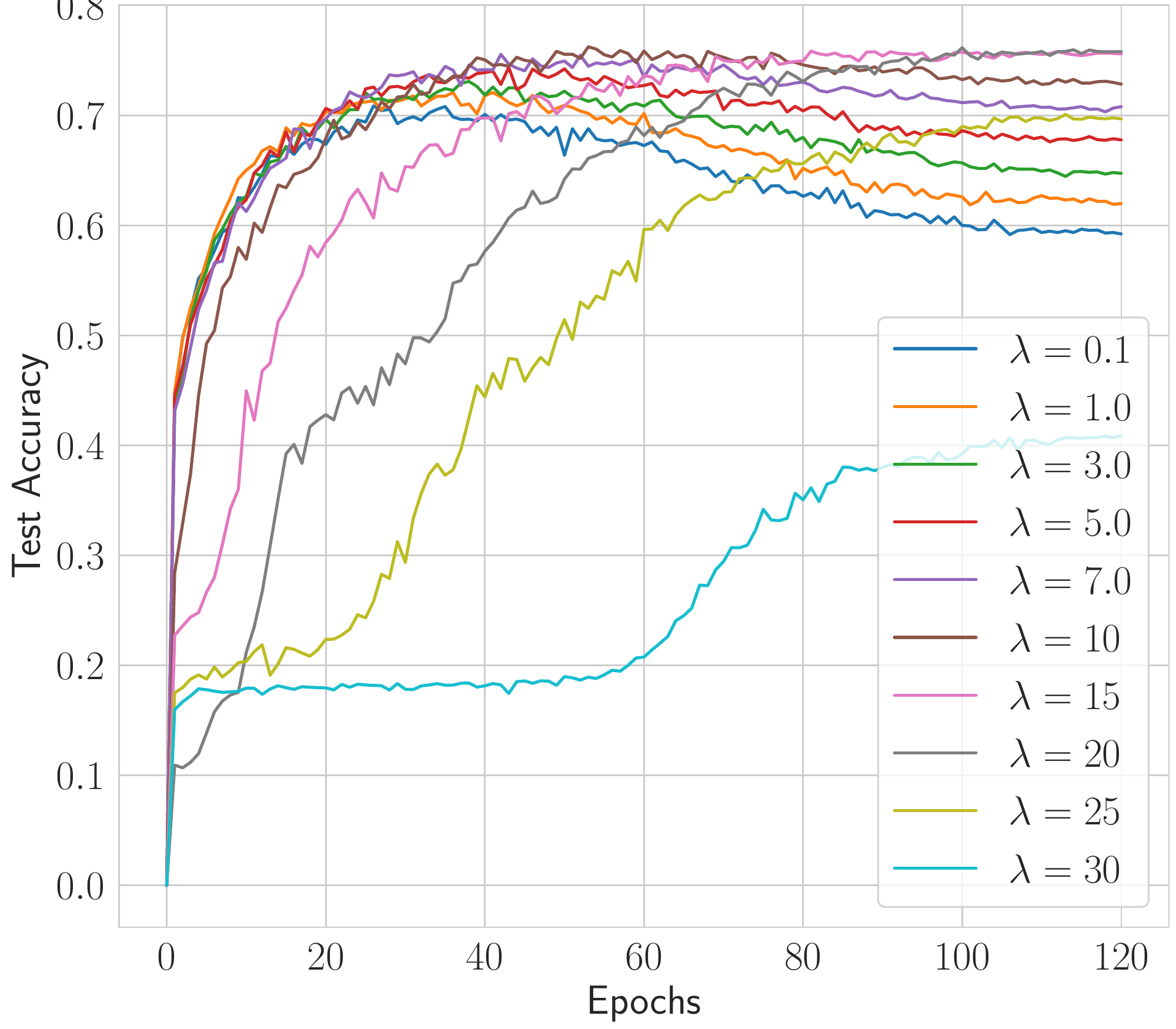}
    }
    \subfigure[FL+SR]{
        \label{lambda-FL}
        \includegraphics[width=1.5in]{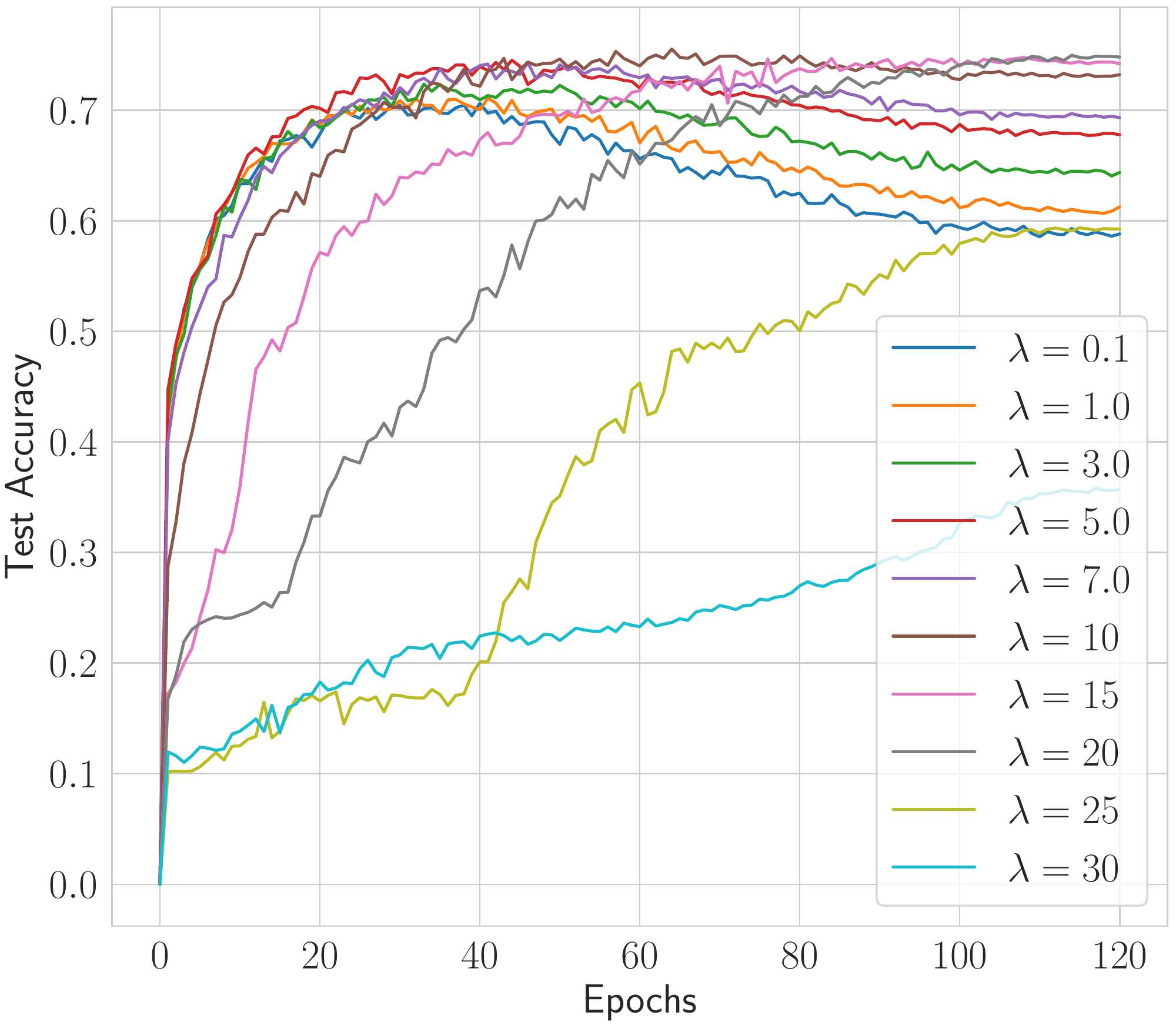}
    }
    \caption{Test accuracy curve of different $\lambda$ on CIFAR-10 with 0.6 symmetric label noise. Robustness is improved when increasing the weight of $\ell_p$-norm for both CE and FL. }
    \label{lambda}
    \vskip-20pt
\end{figure}

For effective learning, the regularization parameter $\lambda$ in Eq. (\ref{norm}) cannot be set too large, since the network would tend to minimize $\lambda \|f(\mathbf x_i)\|_p^p$ rather than $L(f(\mathbf x_i), \tilde{y}_i)$. As shown in Fig. \ref{lambda-CE}, when $\lambda=25$ and 30, although the curves look robust, they suffer from the underfitting effect. On the other hand, $\lambda$ should not be set too small, otherwise robustness cannot be guaranteed (see $\lambda\le 7$ in Fig. \ref{lambda-CE}). We need a large enough $\lambda$ to maintain robustness. An effective strategy in practical implementation is to gradually increase the value of $\lambda$ during training, \textit{i.e.}, $\lambda_t = \lambda_0 \cdot \rho^{\lfloor t/r\rfloor}$ ($\rho\ge1$), where $t$ denotes the training epoch, and $r$ denotes the updating rate of $\lambda$.

\subsection{On the Robustness and Learning Sufficiency}
In the following, we provide analysis about the robustness and learning sufficiency of the proposed scheme.

To obtain enough robustness, we restrict the output of the network to be one-hot, which naturally satisfies the symmetric condition. As for the output sharpening process, the derivative of $\sigma_\tau(\mathbf z)_j$ with respect to $z_i$ can be derived as
\begin{equation}
    \label{derivative-ol}
    \frac{\partial \sigma_\tau(\mathbf z)_j}{\partial z_i}=\frac{1}{\tau}\sigma_\tau(\mathbf z)_i(\delta_{ij}-\sigma_\tau(\mathbf z)_j),
\end{equation}
where $\delta_{ij}=\mathbb I(i=j)$, and $\mathbb I(\cdot)$ is the identity function. We can see that the derivative is a scaled derivative of the original softmax function, so it would not change the optimization direction but change the step size. A larger step size in Eq. (\ref{derivative-ol}) would speed up the convergence to one-hot vectors. We can achieve this purpose by choosing an appropriate value for $\tau$. On the other hand, we have $\lim_{\tau\rightarrow 0^+}\frac{\partial \sigma_\tau(\mathbf z)_j}{\partial z_i}=0$, which shows that the gradient would disappear if $\tau$ is small, so $\tau$ cannot be overly small to prevent underfitting. We fix $\tau$ in our implementation for simplicity, but we suggest to gradually decay $\tau$ in training, which can be regarded as an early-stopping strategy \cite{li2020gradient}.

Moreover, consider the loss $L(\sigma_{\tau}(\mathbf z), y)=-\log \sigma_{\tau}(\mathbf z)_y$, we have the derivative of $-\log \sigma_{\tau}(\mathbf z)_y +\lambda \|\sigma_{\tau}(\mathbf z)\|_p^p$ with respect to $\mathbf z$ as follows
$$
\begin{aligned}
&\frac{\partial -\log \sigma_{\tau}(\mathbf z)_y}{\partial \mathbf z}+\lambda \frac{\partial \|\sigma_\tau(\mathbf z)\|_p^p}{\partial \mathbf z}\\
% =&-\frac{1}{\sigma_{\tau}(\mathbf z)_y}\cdot \frac{\partial \sigma_{\tau}(\mathbf z)_y}{\partial \mathbf z}+\lambda\sum_{i=1}^k \frac{\partial [\sigma_{\tau}(\mathbf z)_i]^p}{\partial \mathbf z}\\
=&-\frac{1}{\sigma_{\tau}(\mathbf z)_y}\cdot \frac{\partial \sigma_{\tau}(\mathbf z)_y}{\partial \mathbf z}+\lambda p\sum_{i=1}^k\frac{1}{[\sigma_{\tau}(\mathbf z)_i]^{1-p}}\cdot \frac{\partial\sigma_{\tau}(\mathbf z)_i}{\partial \mathbf z}\\
=&\underbrace{-(\frac{1}{\sigma_{\tau}(\mathbf z)_y}-\frac{\lambda p}{[\sigma_{\tau}(\mathbf z)_y]^{1-p}})\cdot \frac{\partial \sigma_{\tau}(\mathbf z)_y}{\partial \mathbf z}}_{\text{fitting term}}\\
&+\underbrace{\lambda p\sum_{i\neq y}\frac{1}{[\sigma_{\tau}(\mathbf z)_i]^{1-p}}\cdot \frac{\partial\sigma_{\tau}(\mathbf z)_i}{\partial \mathbf z}}_{\text{complementary term}}\\
\end{aligned}
$$

The fitting term denotes the gradient of learning towards the target $y$, while the complementary term limits the increase of $\sigma_{\tau}(\mathbf z)_i$, $\forall i\neq y$. In the early phase of training, we guarantee enough fitting power by setting $\lambda p<1$. As $\lambda$ increases, the fitting term becomes weaker to mitigate label noise, but the complementary term still maintains a certain amount of fitting power through minimizing $\sigma_{\tau}(\mathbf z)_i$, $\forall i\neq y$ to passively maximize $\sigma_{\tau}(\mathbf z)_y$.

\begin{figure}[htb]
    \centering
    \subfigure[Sparse Rate]{
        \label{sparse-rate}
        \includegraphics[width=1.5in]{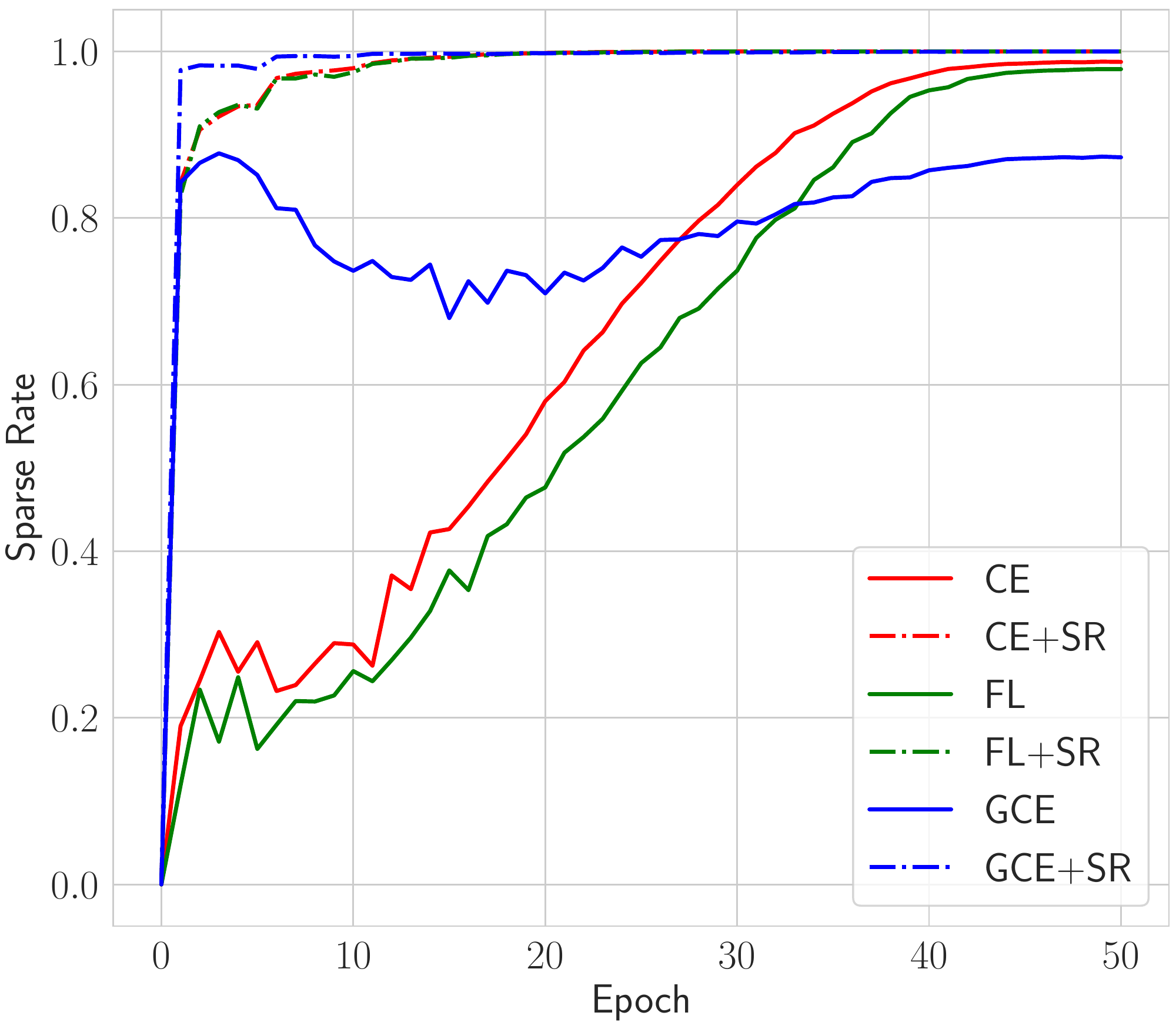}
    }
    \subfigure[Test Accuracy]{
        \label{test-acc}
        \includegraphics[width=1.5in]{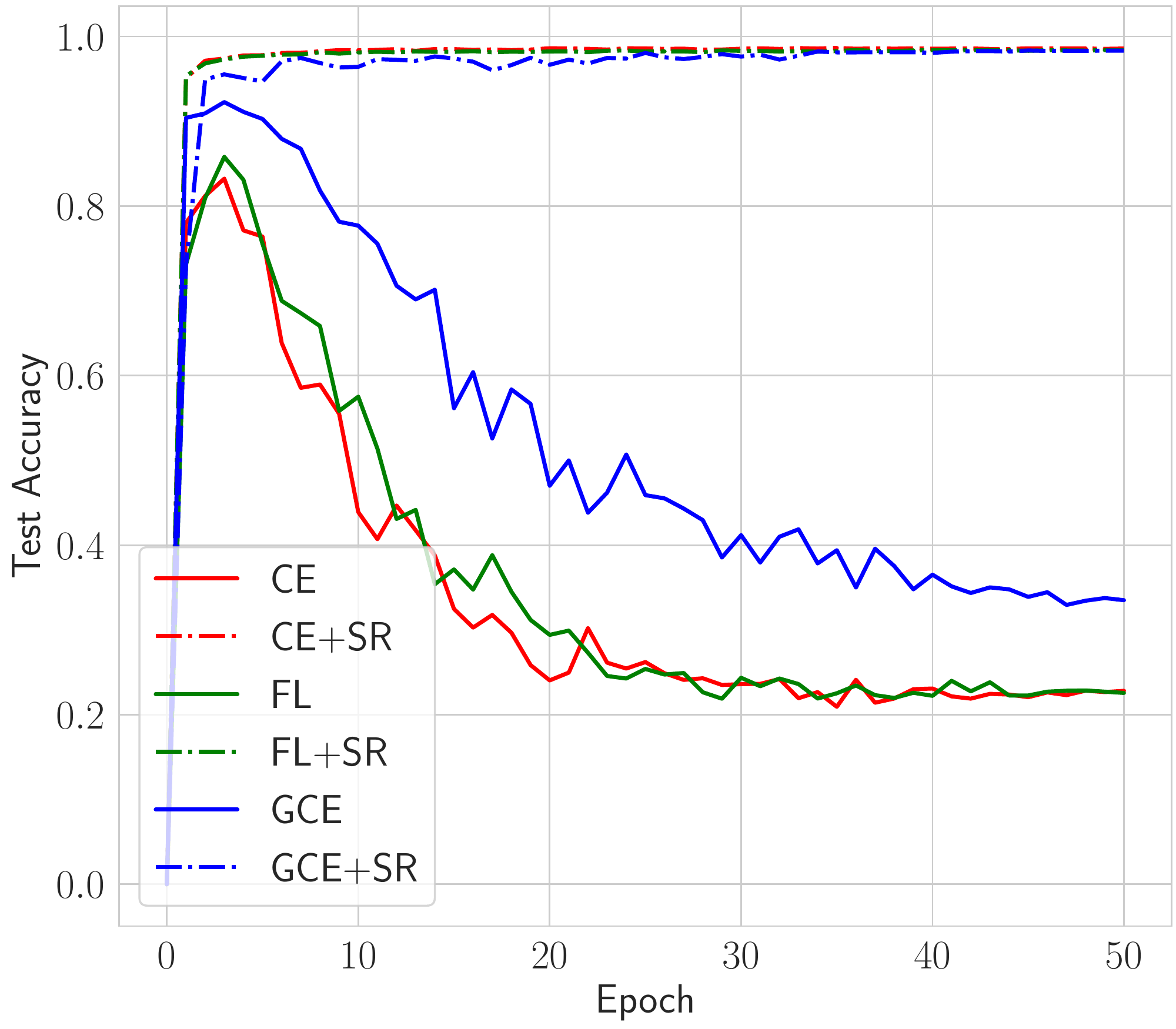}
    }
    \caption{Sparse rate and test accuracy of different methods on MNIST with 0.8 symmetric label noise. One-hot constraint significantly improves the robustness while not hindering fitting ability.}
    \label{sparse}
    \vskip-5pt
\end{figure}

On the other hand, we can regard the $\ell_p$-norm regularized loss function $L(\mathbf u, i)+\lambda \|\mathbf u\|_p^p$ as a new loss function. If there exists $\alpha>0$, such that $L(\mathbf u,i)+\alpha u_i^p$ is monotonically decreasing on $u_i$, then the new loss can be divided into the active loss $L(\mathbf u,i)+\alpha u_i^p$ and the passive loss $(\lambda-\alpha)u_i^p+\lambda\sum_{j\not=i} u_j^p$. In fact, $\alpha$ always exists for commonly-used loss functions, for example, when $L(\mathbf u,i)=-\log u_i$, we have $\alpha<\frac{1}{p}$ such that $-\log u_i +\alpha u_i^p$ keeps monotonically decreasing on $u_i$. Therefore, our proposed $\ell_p$-norm regularization coincides with the Active Passive Loss proposed in \cite{ma2020normalized}. The analysis demonstrates that our scheme achieves the best of both worlds of robustness and sufficient learning.

\section{Experiments}
In this section, we empirically investigate the effectiveness of sparse regularization on synthetic datasets, including MNIST \cite{MNIST}, CIFAR-10/-100 \cite{CIFAR}, and a real-world noisy dataset WebVision \cite{Webvision}.

\subsection{Empirical Analysis}
% We first run a set of experiments to verify that sparse regularization can mitigate label noise.
\noindent\textbf{One-hot Constraint Means Robustness.}\quad We first run a set of experiments on MNIST with 0.8 symmetric label noise to analyse the sparse rate and test accuracy during training, where the sparse rate is formulated as $\frac{\sum_{i=1}^N \mathbb{I}(\max_j f(\mathbf x_i)_j>1-0.01)}{N}$,
% test accuracy is calculated by $\frac{\sum_{i=1}^N\mathbb{I}(\arg\min_{j}f(\mathbf x_i)_j=y_i)}{N}$,
and $f(\mathbf x_i)$ is performed by the output sharpening with $\tau=0.1$. If $\max_j f(\mathbf x_i)_j>1-0.01$, then $\min_{\mathbf u\in\mathcal P_{0-1}}\|f(\mathbf x_i)-\mathbf u\|_2\le \sqrt{2}\cdot 0.01=\varepsilon$. We add the sparse regularization strategy to CE, FL and GCE, the results are shown in Fig. \ref{sparse}. As we can see, with the role of SR, the sparse rate usually maintains a high value after several epochs, and the test accuracy curves of CE+SR, FL+SR and GCE+SR show enough robustness and learning efficiency for model to mitigate label noise, while the original losses have low sparse rate and poor accuracy. This validates that the noise tolerance can be obtained by restricting the output of a network to one-hot vectors.
\begin{figure}[t]
    \centering
    \subfigure[CE with different $p$]{
        \label{p-CE}
        \includegraphics[width=1.5in]{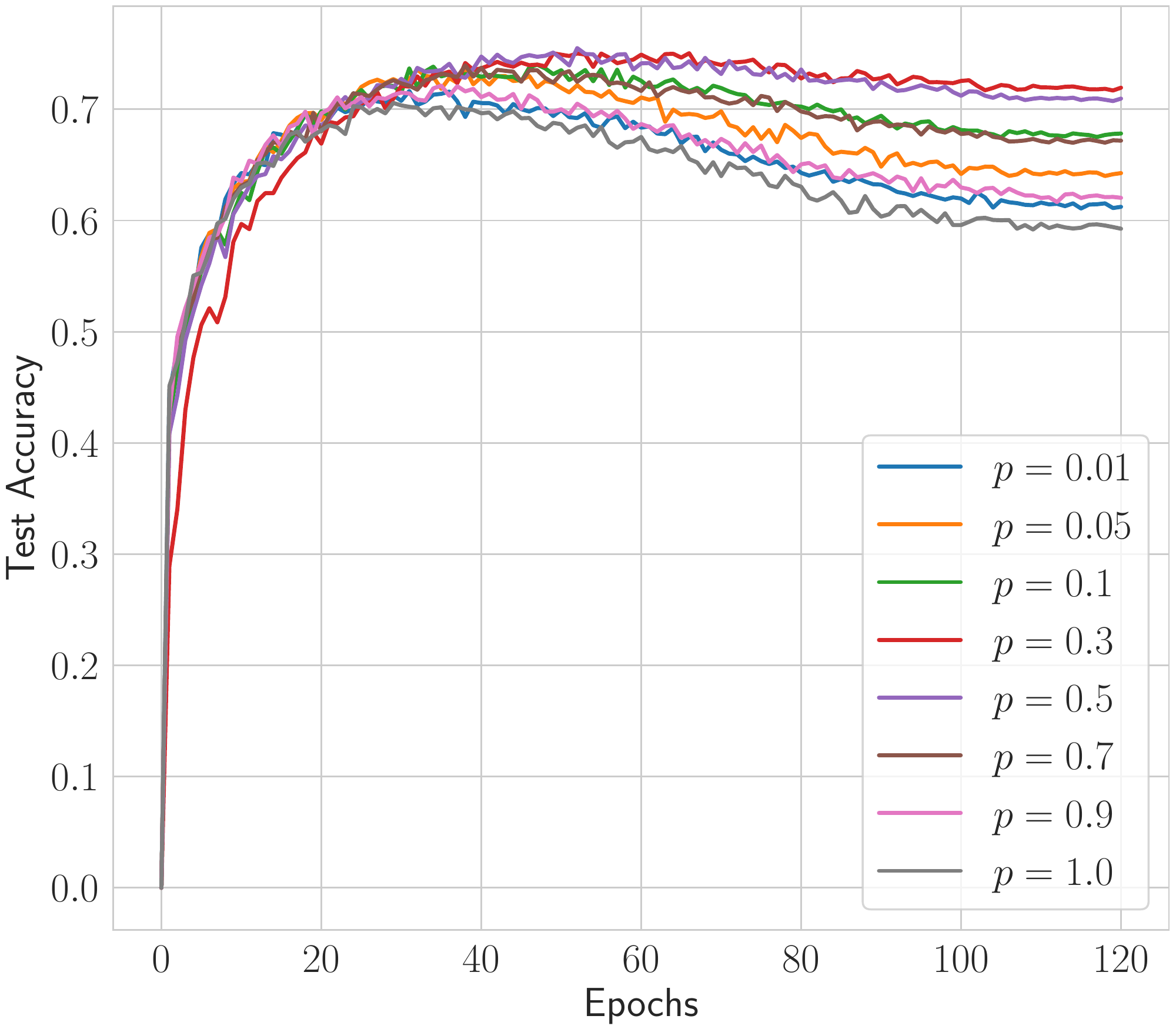}
    }
    \subfigure[CE with different $\tau$]{
        \label{t-CE}
        \includegraphics[width=1.5in]{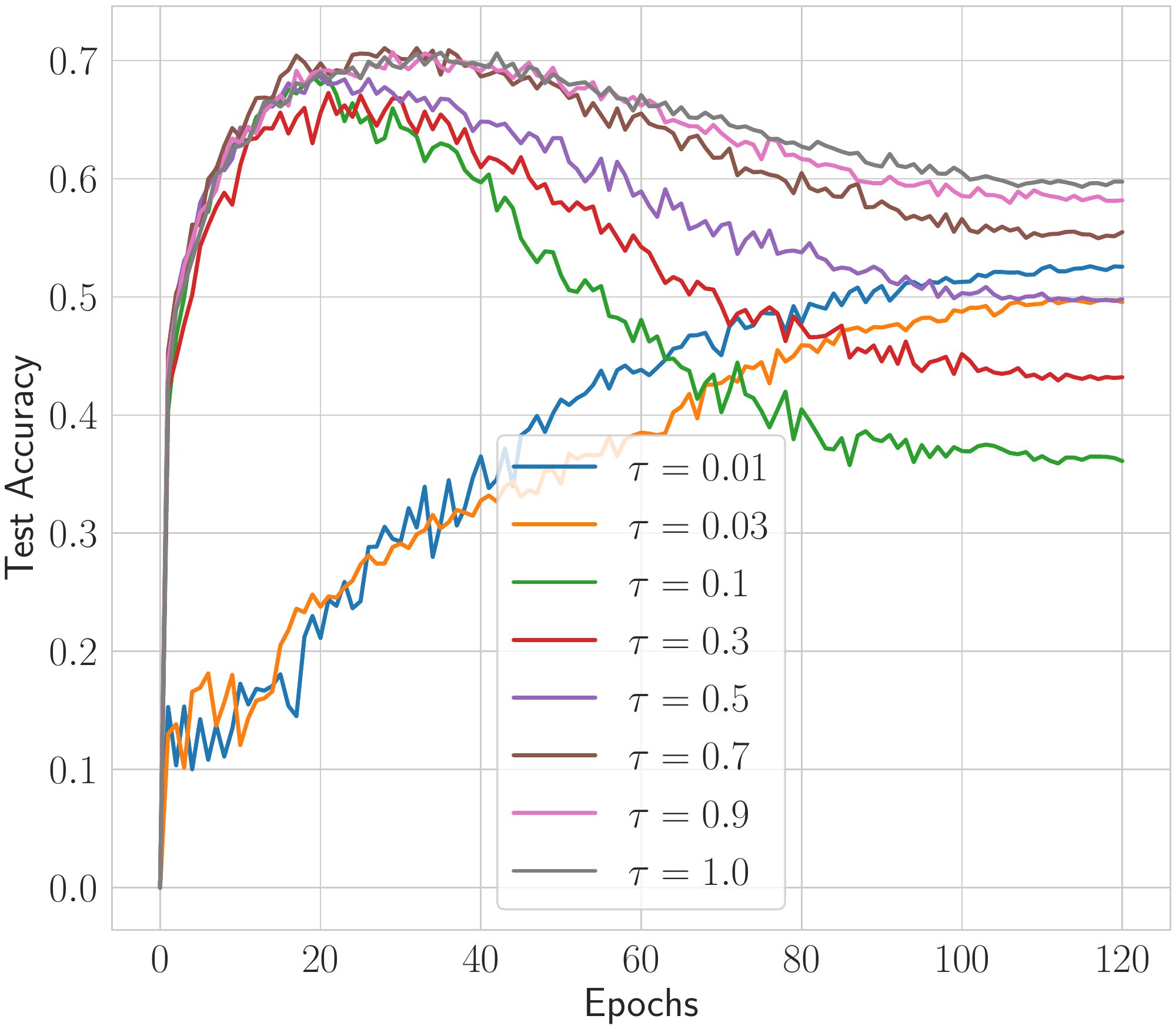}
    }
    \caption{Test accuracy curve of different parameters on CIFAR-10 with 0.6 symmetric label noise.}
    \label{p}
    \vskip-15pt
\end{figure}

\begin{figure*}[htb]
    \centering
    \subfigure[CE with $\eta=0.0$]{
    \label{fig:CE0.0}
    \includegraphics[width=1.2in]{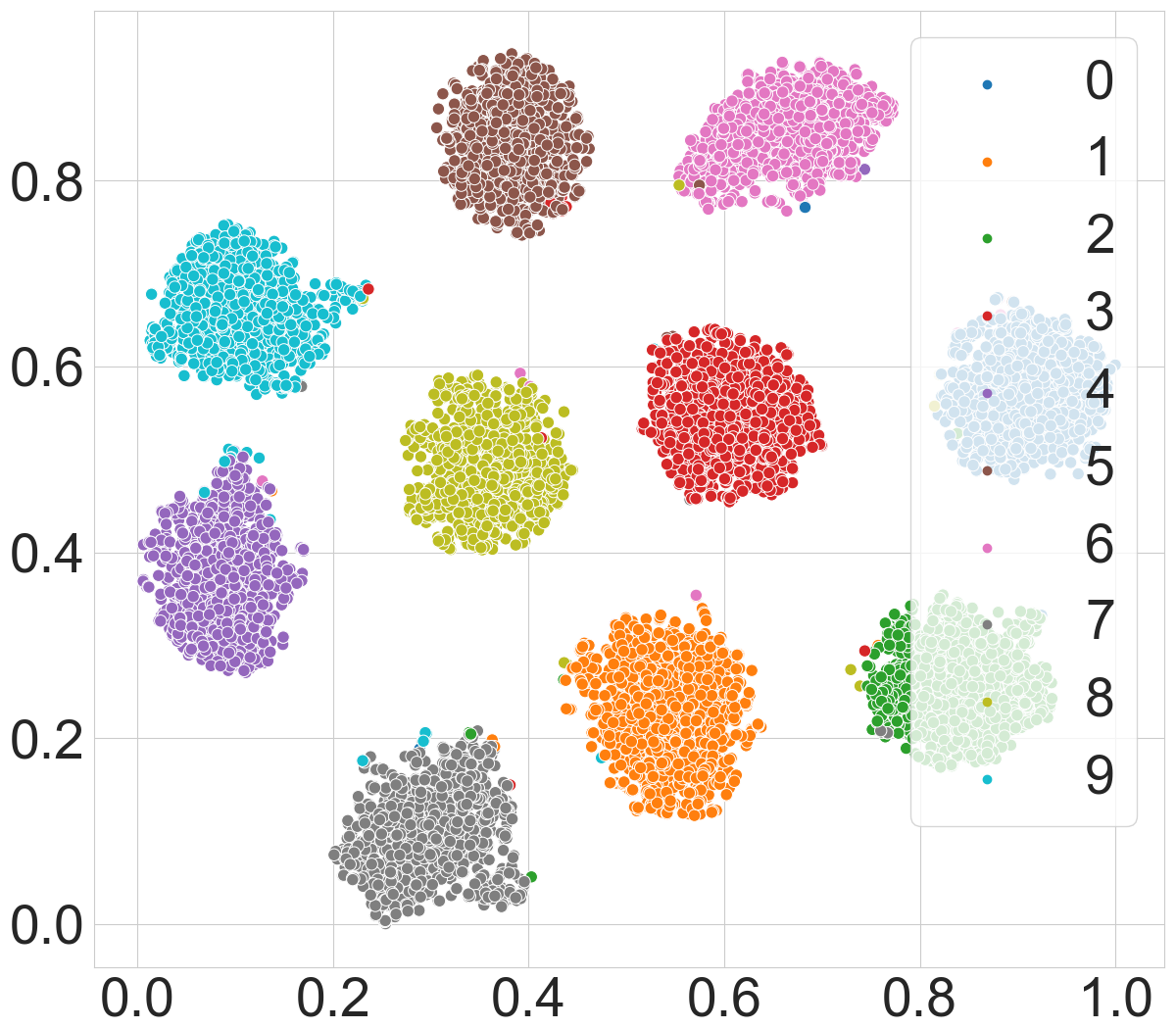}
    }
    \subfigure[CE with $\eta=0.2$]{
    \label{fig:CE0.2}
    \includegraphics[width=1.2in]{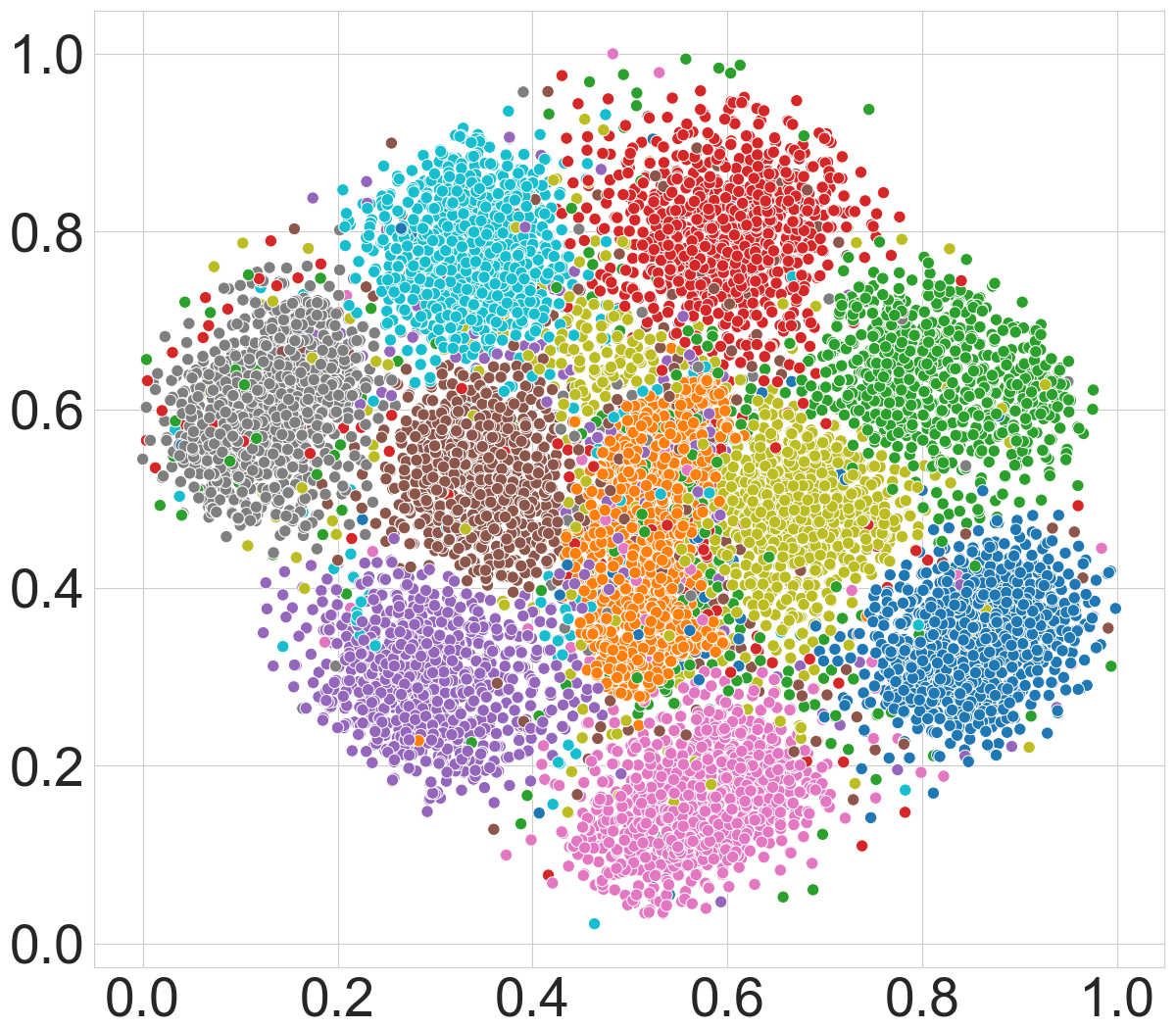}
    }
    \subfigure[CE with $\eta=0.4$]{
    \label{fig:CE0.4}
    \includegraphics[width=1.2in]{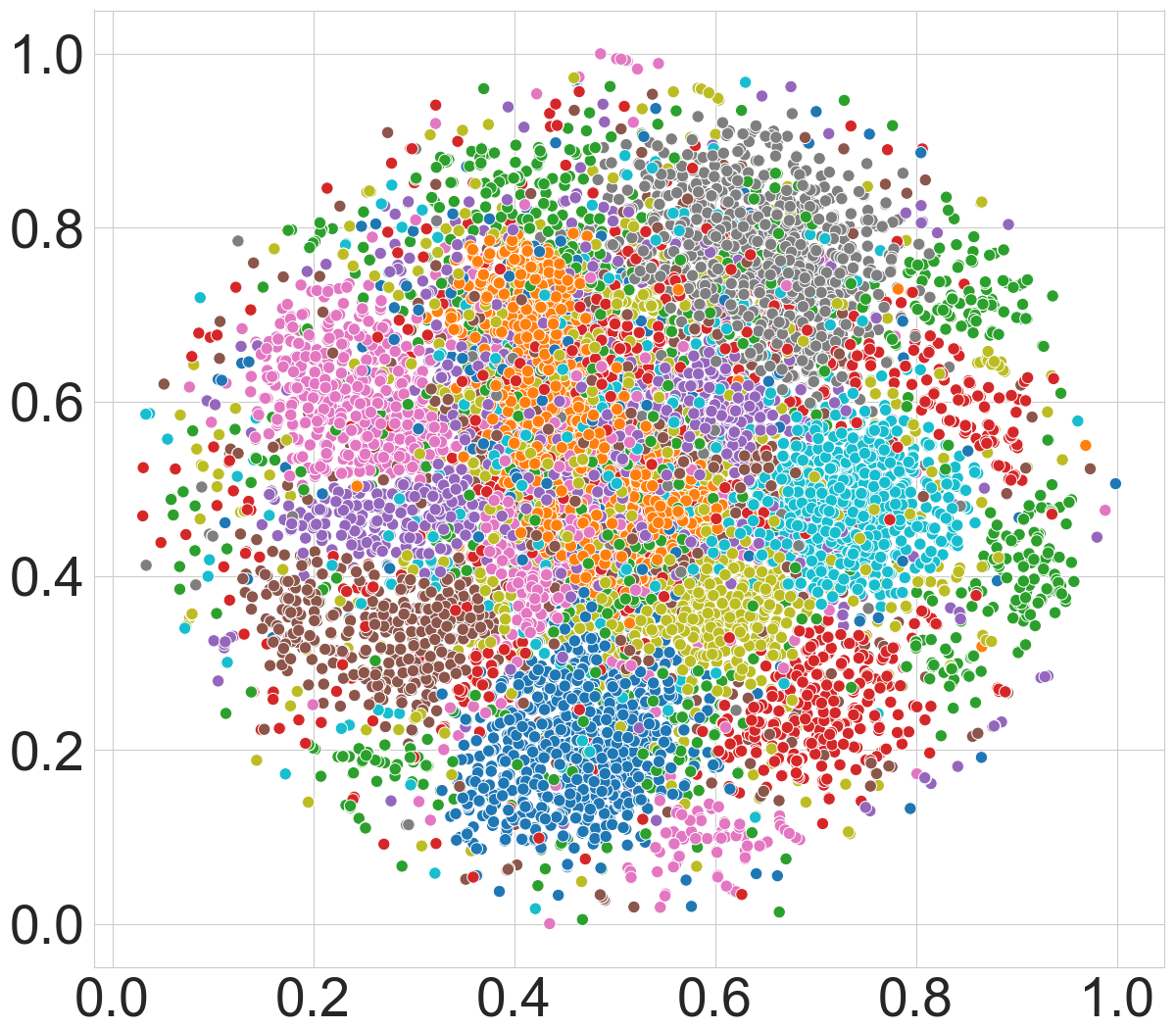}
    }
    \subfigure[CE with $\eta=0.6$]{
    \label{fig:CE0.6}
    \includegraphics[width=1.2in]{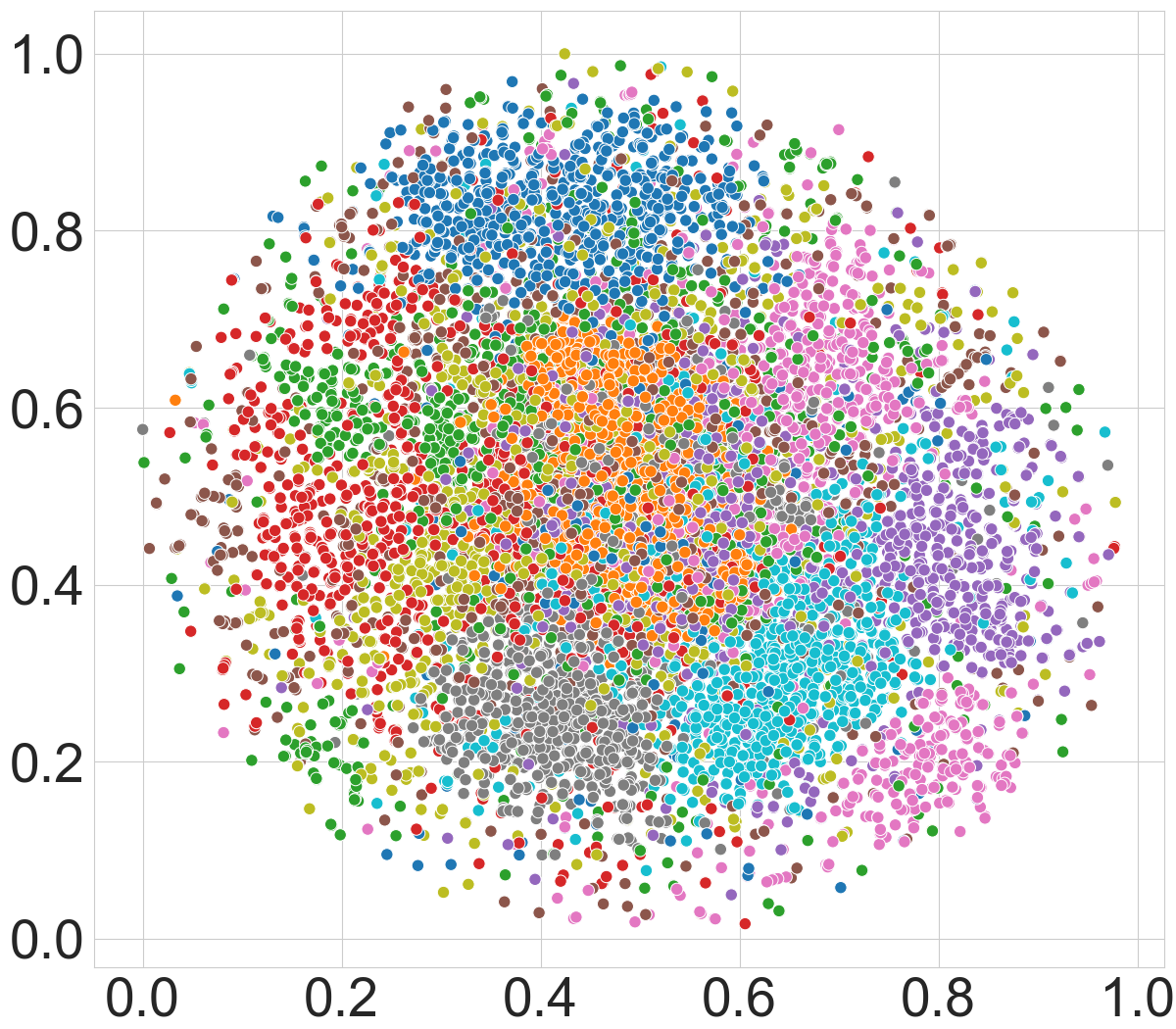}
    }
    \subfigure[CE with $\eta=0.8$]{
    \label{fig:CE0.8}
    \includegraphics[width=1.2in]{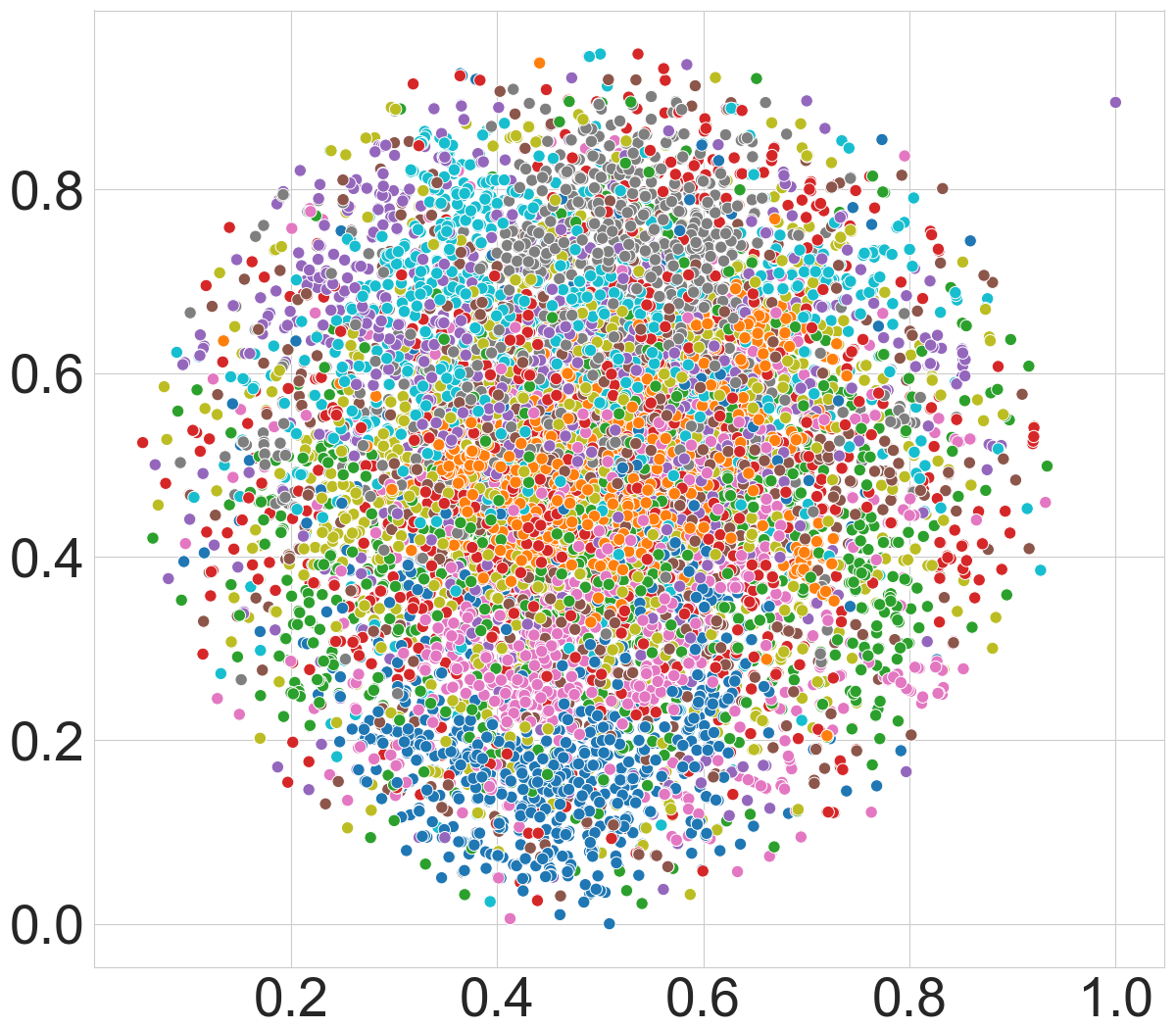}
    }
    \subfigure[CE+SR with $\eta=0.0$]{
    \label{fig:CE+SR0.0}
    \includegraphics[width=1.2in]{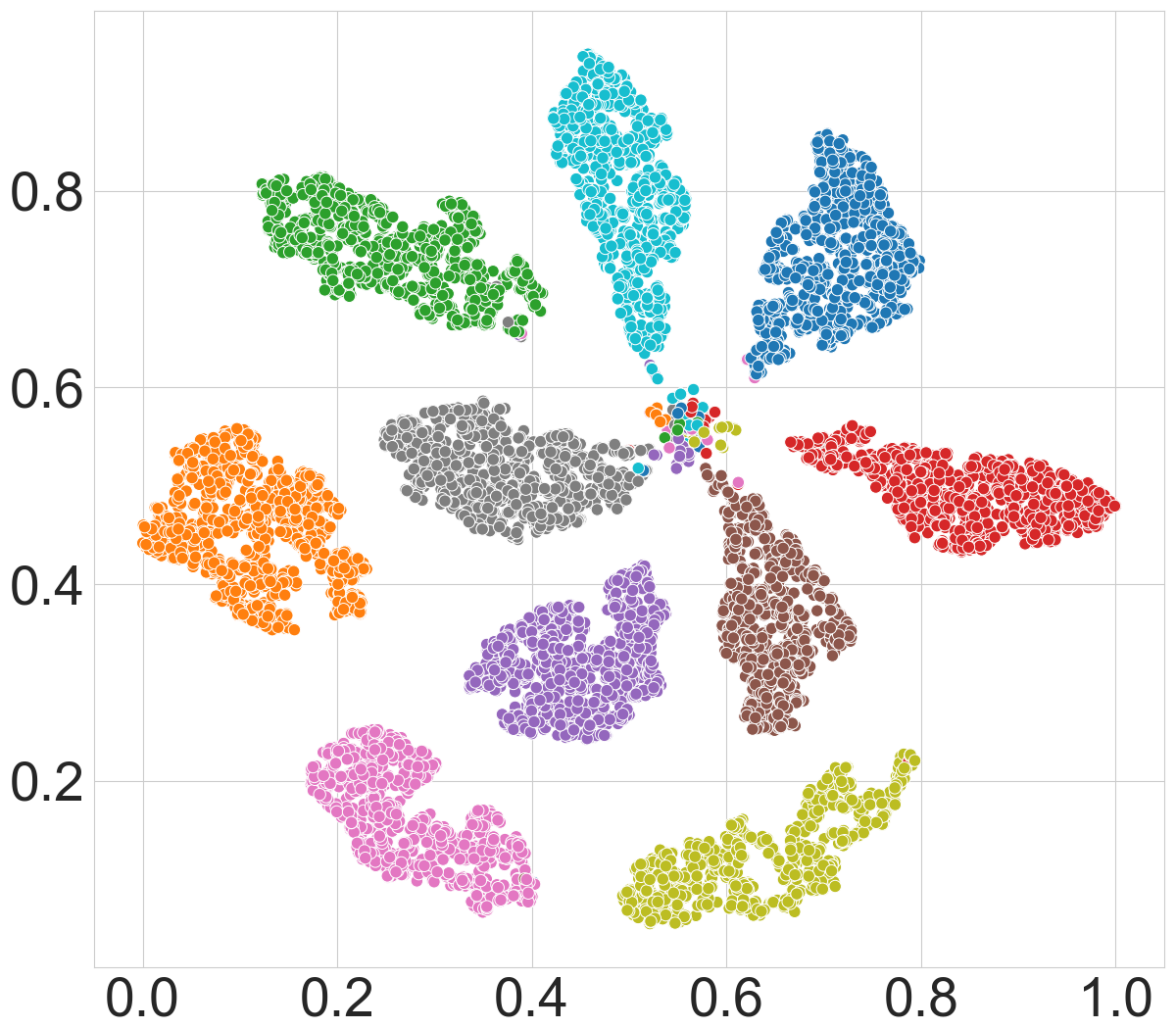}
    }
    \subfigure[CE+SR with $\eta=0.2$]{
    \label{fig:CE+SR0.2}
    \includegraphics[width=1.2in]{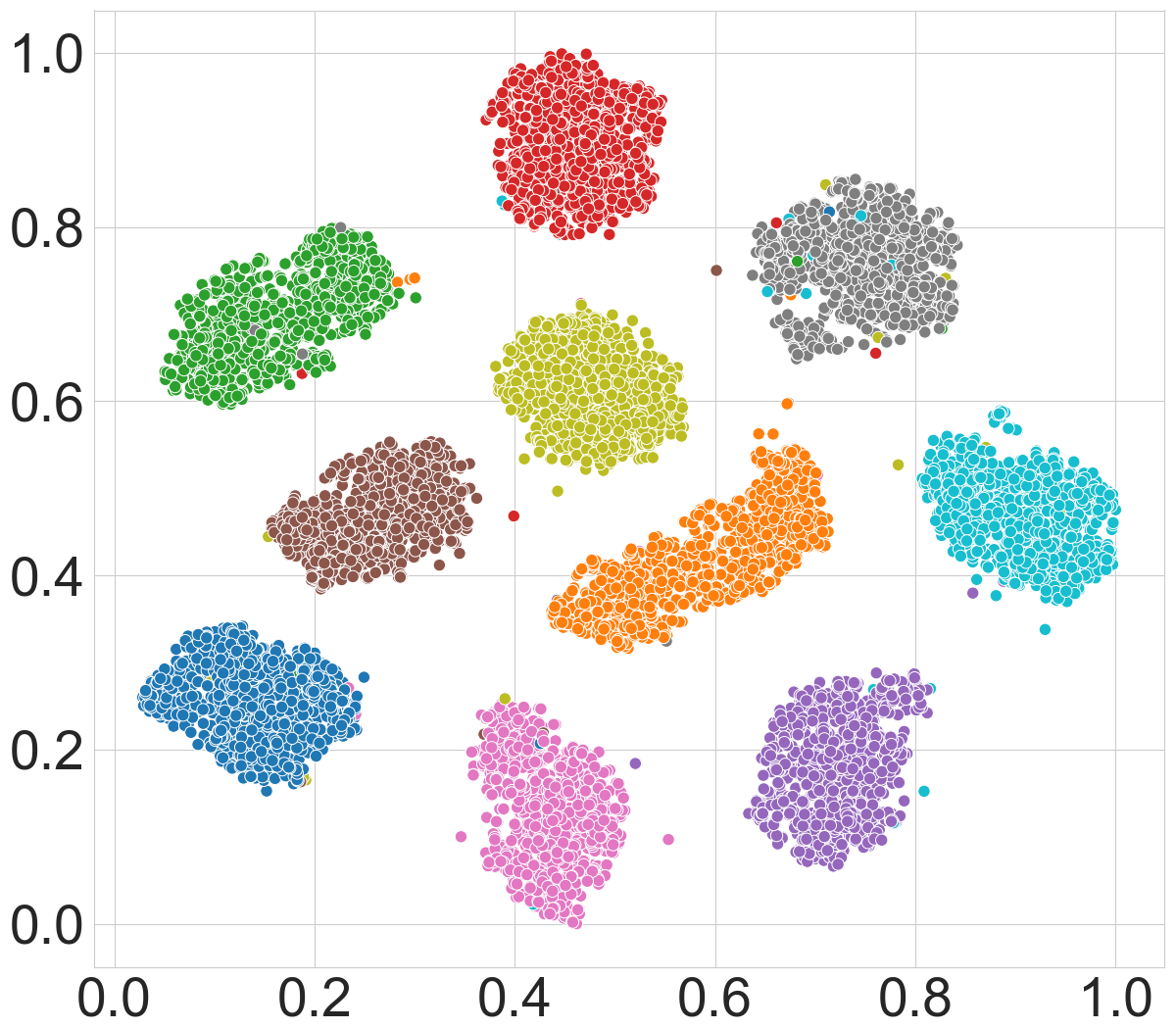}
    }
    \subfigure[CE+SR with $\eta=0.4$]{
    \label{fig:CE+SR0.4}
    \includegraphics[width=1.2in]{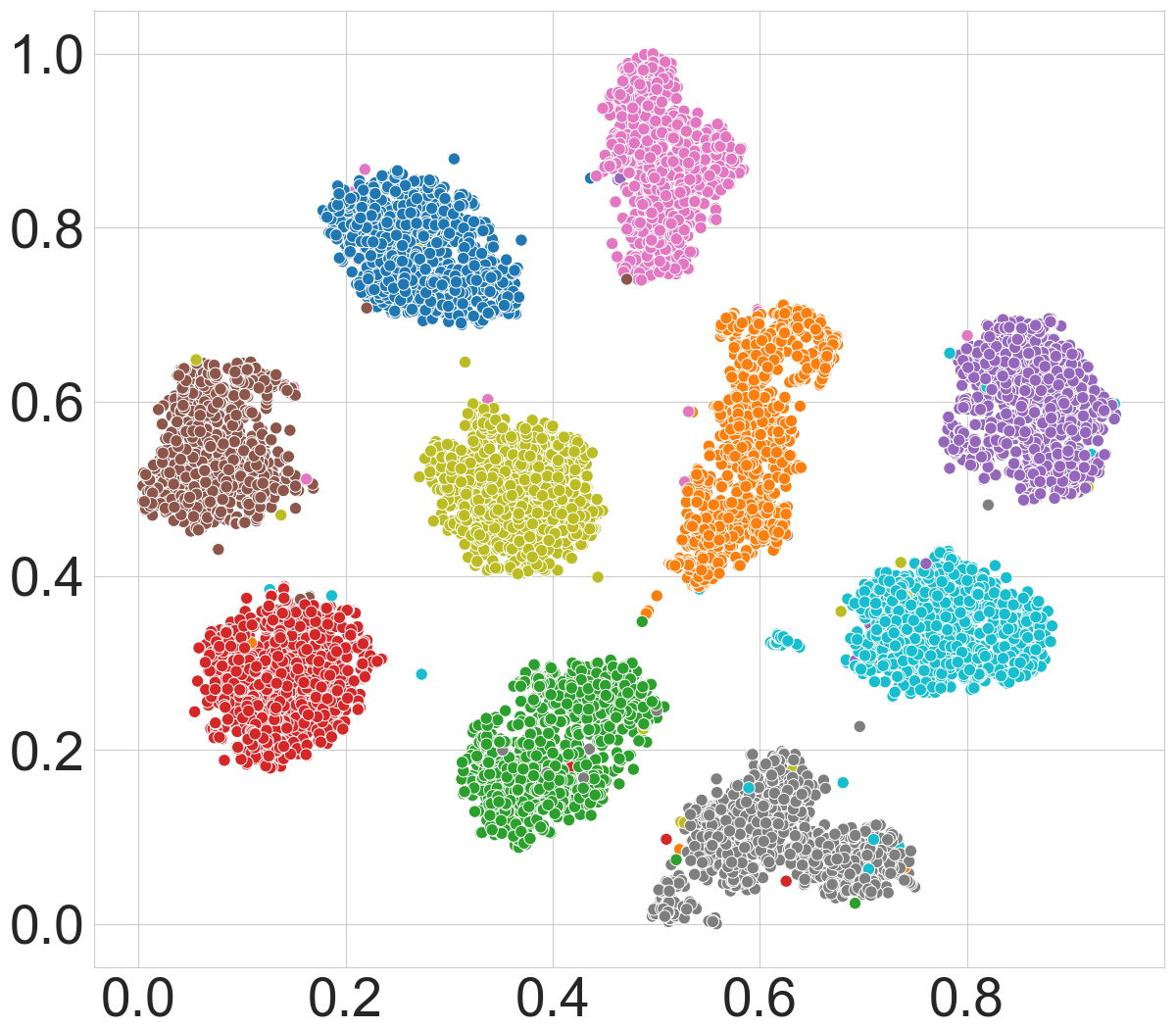}
    }
    \subfigure[CE+SR with $\eta=0.6$]{
    \label{fig:CE+SR0.6}
    \includegraphics[width=1.2in]{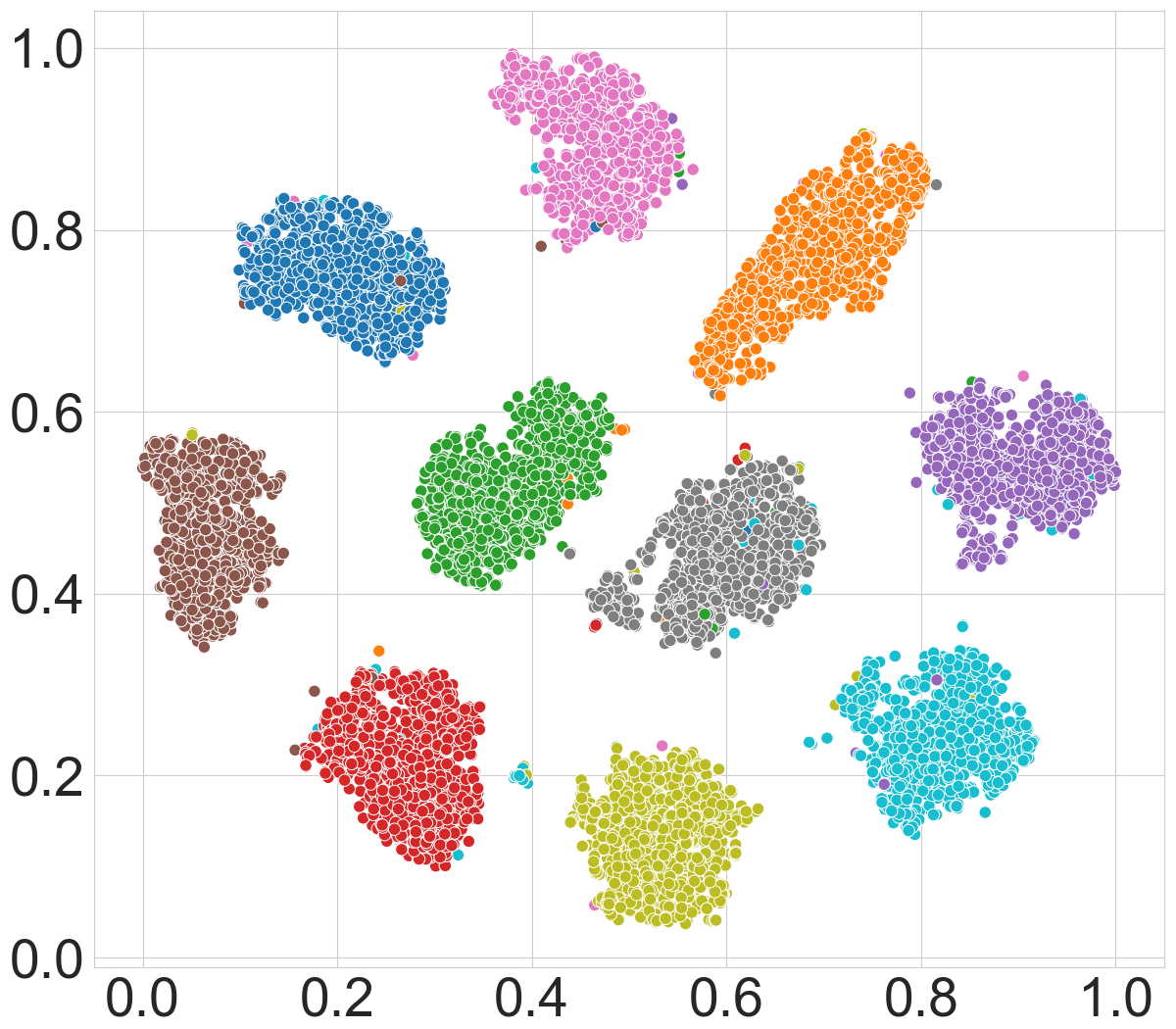}
    }
    \subfigure[CE+SR with $\eta=0.8$]{
    \label{fig:CE+SR0.8}
    \includegraphics[width=1.2in]{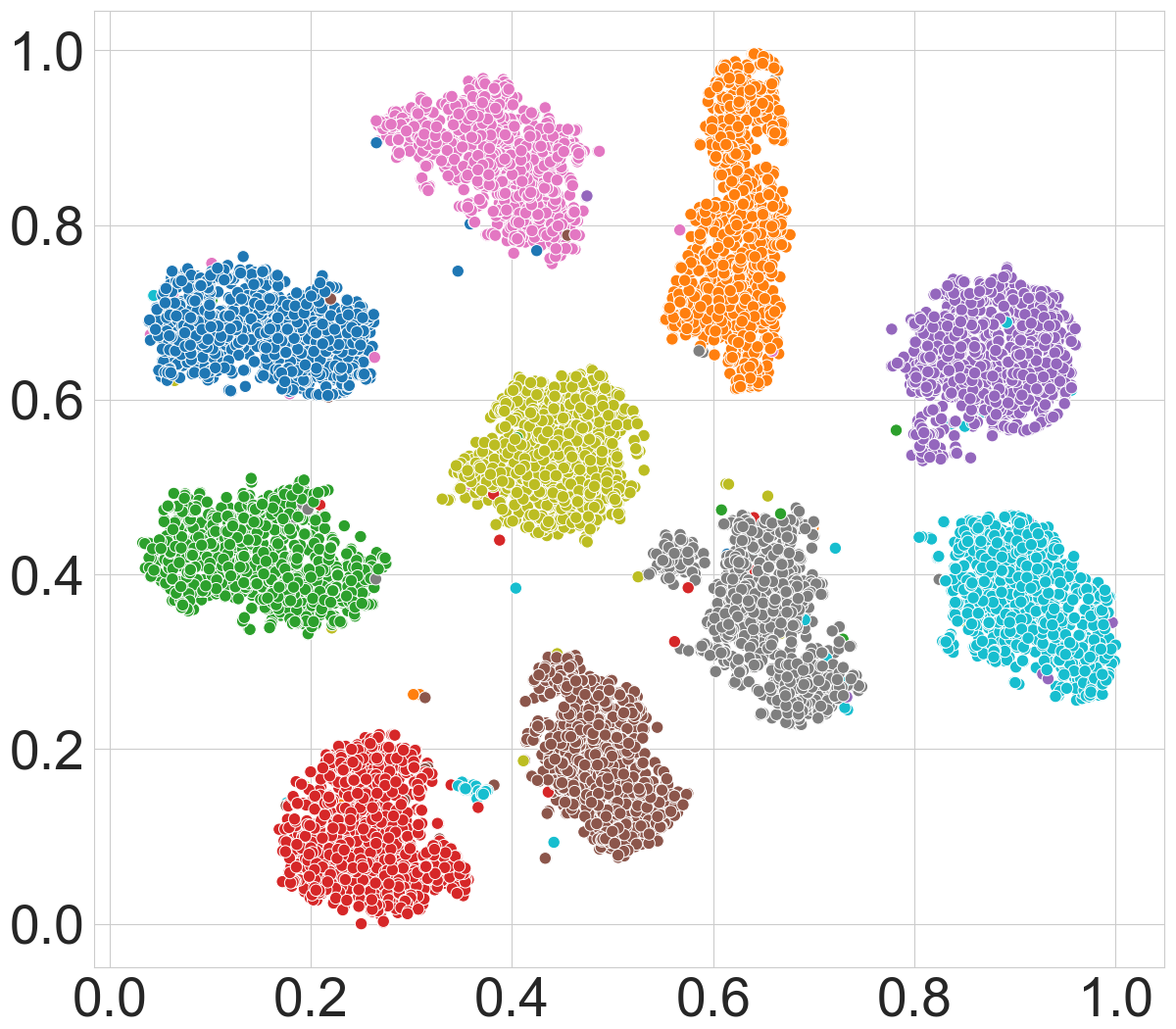}
    }
    \caption{Visualization of learned representations on MNIST with different symmetric label noise ($\eta\in[0.0, 0.2, 0.4, 0.6, 0.8]$).}
    \label{fig:tsne}
    \vskip-10pt
\end{figure*}
\noindent\textbf{Sparse Regularization can Mitigate Label Noise.}\quad As shown in Fig. \ref{lambda}, when we add SR to enhance the performance of CE and FL, the training process is increasingly robust as $\lambda$ increases while not hindering the fitting ability ($\lambda\le 20$). This demonstrates that learning with sparse regularization can be both robust and effective when mitigating label noise. The projected representations on MNIST are illustrated in Fig. \ref{tsne-mnist}. Under both settings, the representations learned by the SR-enhanced method are of significantly better quality than those learned by the original losses with more separated and clearly bounded clusters.

\noindent\textbf{Parameter Analysis.}\quad We choose the different parameters $\tau$, $\lambda$ and $p$ for sparse regularization to CE. The experiments are conducted on CIFAR-10 with 0.6 symmetric noise. We first tested $\tau\in\{0.01, 0.03, 0.1, 0.3, 0.5, 0.7, 0.9, 1\}$ while $\lambda=0$, and the results can be show in Fig. \ref{t-CE}. When $\tau$ is small ($\le 0.003$), the curve is very robust, but it also suffer from significant underfitting problem. As $\tau$ increases, the curve falls into overfitting. Then we tested $\lambda\in\{0.1, 0.3, 1.0, 3.0, 5.0, 7.0, 10, 15, 20, 25, 30\}$ while $p=0.1$ and $\tau=1$. As shown in Fig. \ref{lambda}, curve gets more robustness as $\lambda$ increases, but when $\lambda\ge 7$, it encounters severe underfitting since the optimization pays more attention on minimizing $\lambda\|f(\mathbf x)\|_p^p$. Moreover, we tuned $p\in\{0.01, 0.05, 0.1, 0.3, 0.5, 0.7, 0.9, 1\}$ while $\tau=1$ and $\lambda=5$. The results in Fig. \ref{p-CE} show that both small $p$ and large $p$ tend to cause more overfitting to label noise, so we need choose an appropriate value. 

\noindent\textbf{Remark.} As for parameter tunings, a simple principled approach for parameter setting is that parameters with strong regularization are selected for simple datasets, and parameters with weak regularization are selected otherwise. Specifically, this can be achieved by setting proper $\lambda$: the larger $\lambda$, the stronger regularization effect. However, as shown in Fig. \ref{lambda}, the large initial $\lambda$ leads to underfitting since much attention is paid on minimizing $\lambda\|f(\mathbf x)\|_p^p$, especially when $\lambda \cdot p>1$. Instead, we turn to gradually increase $\lambda$, \textit{i.e.}, $\lambda_t = \lambda_0 \cdot \rho^{\lfloor t/r\rfloor}$, where $t$ is the iteration number and $r=1$; $p=0.1$ and $0.01$, $\rho=50^{1/120}\approx 1.03$ and $50^{1/200} \approx 1.02$ for CIFAR-10 and CIFAR-100 (where 120 and 200 represent the training epochs). The only parameter that requires careful tuning is $\lambda_0$, which is set as $\lambda_0 < \frac{1}{p}$. We use the similar parameter setting strategy for CIFAR-10, CIFAR-100, WebVision, and all achieve satisfactory results.

\subsection{Evaluation on Benchmark Datasets}
\label{benchmark}

\noindent \textbf{Experimental Details.} The benchmark datasets, noise generation, networks, training details, parameter settings, more comparisons, and more experimental results can be found in the supplementary materials.

% \FloatBarrier
\begin{table*}
\small
\centering
\caption{Test accuracies (\%) of different methods on benchmark datasets with clean or symmetric label noise ($\eta\in[0.2, 0.4, 0.6, 0.8]$). The results (mean$\pm$std) are reported over 3 random runs and the top 3 best results are \textbf{boldfaced}.}
\label{symmetric-noise}
\begin{tabular}{c|c|c|cccc}
    \toprule
      \multirow{2}*{Datasets} &\multirow{2}*{Methods} & \multirow{2}*{Clean ($\eta=0.0$)} & \multicolumn{4}{c}{Symmetric Noise Rate ($\eta$)} \\
      ~ & ~ & ~ & 0.2 & 0.4 & 0.6 & 0.8\\
     \hline\hline
      \multirow{9}*{MNIST} & CE & 99.15 $\pm$ 0.05 & 91.62 $\pm$ 0.39 & 73.98 $\pm$ 0.27 & 49.36 $\pm$ 0.43 & 22.66 $\pm$ 0.61\\
     ~ & FL &99.13 $\pm$ 0.09 &91.68 $\pm$ 0.14 &74.54 $\pm$ 0.06 &50.39 $\pm$ 0.28 &22.65 $\pm$ 0.26 \\
     ~ & GCE &99.27 $\pm$ 0.05 &98.86 $\pm$ 0.07 &97.16 $\pm$ 0.03 &81.53 $\pm$ 0.58 &33.95 $\pm$ 0.82\\
    % ~ & MAE &99.16 $\pm$ 0.07 &99.01 $\pm$ 0.05 &98.46 $\pm$ 0.04 &95.03 $\pm$ 0.17 &62.57 $\pm$ 1.31\\
    ~ & SCE & 99.23 $\pm$ 0.10 &98.92 $\pm$ 0.12 &97.38 $\pm$ 0.15 &88.83 $\pm$ 0.55 &48.75 $\pm$ 1.54\\
     ~ & NLNL & 98.85 $\pm$ 0.05 & 98.33 $\pm$ 0.03 & 97.80 $\pm$ 0.07 & 96.18 $\pm$ 0.11 &86.34 $\pm$ 1.43\\
     ~ & APL & 99.34 $\pm$ 0.02 & \textbf{99.14 $\pm$ 0.05} &98.42 $\pm$ 0.09 &95.65 $\pm$ 0.13 &72.97 $\pm$ 0.34\\
     \cline{2-7}
     ~ & \textbf{CE+SR} & {99.33 $\pm$ 0.02} & \textbf{99.22 $\pm$ 0.06} & \textbf{99.16 $\pm$ 0.04} & \textbf{98.85 $\pm$ 0.02} & \textbf{98.06 $\pm$ 0.86}\\
     ~ & \textbf{FL+SR} & {99.35 $\pm$ 0.05} & \textbf{99.25 $\pm$ 0.01} & \textbf{99.10 $\pm$ 0.10} & \textbf{98.81 $\pm$ 0.06} & \textbf{97.00 $\pm$ 1.28}\\
     ~ & \textbf{GCE+SR} & 99.27 $\pm$ 0.06 & 99.13 $\pm$ 0.07 & \textbf{99.06 $\pm$ 0.02} & \textbf{98.84 $\pm$ 0.09} & \textbf{98.37 $\pm$ 0.26}\\
    \hline\hline
    \multirow{9}*{CIFAR-10} & CE & 90.48 $\pm$ 0.11 & 74.68 $\pm$ 0.25 & 58.26 $\pm$ 0.21 & 38.70 $\pm$ 0.53 & 19.55 $\pm$ 0.49\\
     ~ & FL & 89.82 $\pm$ 0.20 & 73.72 $\pm$ 0.08 & 57.90 $\pm$ 0.45 & 38.86 $\pm$ 0.07 & 19.13 $\pm$ 0.28\\
    %  ~ & MAE & 89.94 $\pm$ 0.13 & 88.06 $\pm$ 0.07 & 82.83$\pm$ 0.24 & 77.93$\pm$ 0.58 & 49.37$\pm$ 0.33\\
     ~ & GCE & 89.59 $\pm$ 0.26 & 87.03 $\pm$ 0.35 & 82.66 $\pm$ 0.17 & 67.70 $\pm$ 0.45 & 26.67 $\pm$ 0.59\\
     ~ & SCE & 91.61 $\pm$ 0.19 & 87.10 $\pm$ 0.25 & 79.67 $\pm$ 0.37 & 61.35 $\pm$ 0.56 & 28.66 $\pm$ 0.27\\
     ~ & NLNL & 90.73 $\pm$ 0.20 & 73.70 $\pm$ 0.05 & 63.90 $\pm$ 0.44 & 50.68 $\pm$ 0.47 & 29.53 $\pm$ 1.55\\
     ~ & APL & 89.17 $\pm$ 0.09 & 86.98 $\pm$ 0.07 &83.74 $\pm$ 0.10 &76.02 $\pm$ 0.16 & {46.69 $\pm$ 0.31}\\
     \cline{2-7}
     ~ & \textbf{CE+SR} & 90.06 $\pm$ 0.02 & \textbf{87.93 $\pm$ 0.07} & \textbf{84.86 $\pm$ 0.18} & \textbf{78.18 $\pm$ 0.36} & \textbf{51.13 $\pm$ 0.51}\\
      ~ & \textbf{FL+SR} & 89.86 $\pm$ 0.11	& \textbf{87.94 $\pm$ 0.19}	& \textbf{84.65 $\pm$ 0.05}	& \textbf{77.85 $\pm$ 0.74}	& \textbf{52.42 $\pm$ 0.76}\\
      ~ & \textbf{GCE+SR} &90.02 $\pm$ 0.40	&\textbf{87.93 $\pm$ 0.27}	&\textbf{84.82 $\pm$ 0.06}	&\textbf{77.65 $\pm$ 0.05}	&\textbf{51.97 $\pm$ 1.13}\\
    \hline\hline
    \multirow{9}*{CIFAR-100} & CE &71.33 $\pm$ 0.43	&56.51 $\pm$ 0.39	&39.92 $\pm$ 0.10	&21.39 $\pm$ 1.17	&\ \  7.59 $\pm$ 0.20\\
     ~ & FL &70.06 $\pm$ 0.70	&55.78 $\pm$ 1.55	&39.83 $\pm$ 0.43	&21.91 $\pm$ 0.89	&\ \ 7.51 $\pm$ 0.09\\
    %  ~ & MAE &4.80 $\pm$ 0.40	&5.03 $\pm$ 1.08	&3.89 $\pm$ 0.25	&2.22 $\pm$ 0.55	&1.79 $\pm$ 0.33\\
     ~ & GCE &63.09 $\pm$ 1.39	&61.57 $\pm$ 1.06	&56.11 $\pm$ 1.35	&\textbf{45.28 $\pm$ 0.61}	&\textbf{17.42 $\pm$ 0.06}\\
     ~ & SCE & 70.64 $\pm$ 0.05	& 56.07 $\pm$ 0.26	& 39.88 $\pm$ 0.67	& 21.16 $\pm$ 0.65	&\ \ 7.63 $\pm$ 0.15\\
     ~ & NLNL & 68.72 $\pm$ 0.60	&46.99 $\pm$ 0.91	& 30.29 $\pm$ 1.64	&16.60 $\pm$ 0.90	&11.01 $\pm$ 2.48\\
    ~ & APL & 67.95 $\pm$ 0.21	& 64.21 $\pm$ 0.24	& 57.70 $\pm$ 0.64	& \textbf{45.20 $\pm$ 0.75}	& \textbf{24.91 $\pm$ 0.42}\\
    \cline{2-7}
    ~ & \textbf{CE+SR} & 72.19 $\pm$ 0.06 & \textbf{67.51 $\pm$ 0.29} & \textbf{60.70 $\pm$ 0.25} & \textbf{44.95 $\pm$ 0.65} & \textbf{17.35 $\pm$ 0.13} \\
     ~ & \textbf{FL+SR} & 72.08 $\pm$ 0.31 & \textbf{67.64 $\pm$ 0.10} & \textbf{60.67 $\pm$ 0.48} & {44.76 $\pm$ 0.08} & {17.16 $\pm$ 0.24}\\
     ~ & \textbf{GCE+SR} & 72.11 $\pm$ 0.26 & \textbf{67.03 $\pm$ 0.46} & \textbf{60.68 $\pm$ 0.90} & 44.66 $\pm$ 0.84 & {17.35 $\pm$ 0.42}\\
    \bottomrule
\end{tabular}
\vskip-12pt
\end{table*}

\noindent \textbf{Baselines.}\quad We experiment with the state-of-the art methods GCE \cite{GCE}, SCE \cite{SCE}, NLNL \cite{kim2019nlnl}, APL \cite{ma2020normalized}, and two effective loss functions CE and Focal Loss (FL) \cite{lin2017focal} for classification. Moreover, we add the proposed sparse regularization mechanism to CE, FL and GCE, i.e., CE+SR, FL+SR and GCE+SR. All the implementations and experiments are based on PyTorch.

\begin{table}
\small
\setlength{\tabcolsep}{0.8mm}
\centering
\caption{Test accuracies (\%) of different methods on benchmark datasets with clean or asymmetric label noise ($\eta\in[0.1, 0.2, 0.3, 0.4]$). The results (mean$\pm$std) are reported over 3 random runs and the top 3 best results are \textbf{boldfaced}.}
\label{asymmetric-noise}
\begin{tabular}{c|c|ccc}
    \toprule
     \multirow{2}*{Datasets} & \multirow{2}*{Methods} & \multicolumn{3}{c}{Asymmetric Noise Rate ($\eta$)}  \\
     ~ & ~ & 0.2 & 0.3 & 0.4\\
     \hline\hline
     \multirow{9}*{MNIST} & CE  & 94.56 $\pm$ 0.22 & 88.81 $\pm$ 0.10 & 82.27 $\pm$ 0.40\\
    ~ & FL  &94.25 $\pm$ 0.15 &89.09 $\pm$ 0.25 &82.13 $\pm$ 0.49\\
    ~ & GCE  &96.69 $\pm$ 0.12 &89.12 $\pm$ 0.24 &81.51 $\pm$ 0.19\\
    % ~ & MAE &99.06 $\pm$ 0.05 &98.92 $\pm$ 0.08 &97.01 $\pm$ 0.06 &90.27 $\pm$ 0.44\\
    ~ & SCE  &98.03 $\pm$ 0.05 &93.68 $\pm$ 0.43 &85.36 $\pm$ 0.17\\
    ~ & NLNL  &98.35 $\pm$ 0.01 &97.51 $\pm$ 0.15 &95.84 $\pm$ 0.26\\
    ~ & APL  &98.89 $\pm$ 0.04 &96.93 $\pm$ 0.17 &91.45 $\pm$ 0.40\\
    \cline{2-5}
    ~ & \textbf{CE+SR}  & \textbf{99.27 $\pm$ 0.06} & \textbf{99.24 $\pm$ 0.08} & \textbf{99.23 $\pm$ 0.07}\\
    ~ & \textbf{FL+SR}  & \textbf{99.31 $\pm$ 0.02} & \textbf{99.23 $\pm$ 0.02} & \textbf{99.36 $\pm$ 0.05}\\
    ~ & \textbf{GCE+SR} & \textbf{99.22 $\pm$ 0.02} & \textbf{99.13 $\pm$ 0.05} & \textbf{99.09 $\pm$ 0.02}\\
    \hline\hline
     \multirow{9}*{CIFAR-10} & CE & 83.32 $\pm$ 0.12 & 79.32 $\pm$ 0.59 & 74.67 $\pm$ 0.38\\
     ~ & FL & 83.37 $\pm$ 0.07 & 79.33 $\pm$ 0.08 & 74.28 $\pm$ 0.44\\
    %  ~ & MAE & 89.26 $\pm$ 0.15 & 80.58 $\pm$ 5.81 & 63.62 $\pm$ 0.44 & 55.97 $\pm$ 0.06\\
     ~ & GCE & 85.93 $\pm$ 0.23 & 80.88 $\pm$ 0.38 & 74.29 $\pm$ 0.43\\
     ~ & SCE & 86.20 $\pm$ 0.37 & 81.38 $\pm$ 0.35 & 75.16 $\pm$ 0.39\\
     ~ & NLNL & 84.74 $\pm$ 0.08 & 81.26$\pm$ 0.43 & 76.97 $\pm$ 0.52\\
     ~ & APL &86.50 $\pm$ 0.31 & 83.34 $\pm$ 0.39 & 77.14 $\pm$ 0.33\\
     \cline{2-5}
     ~ & \textbf{CE+SR}  & \textbf{87.70 $\pm$ 0.19} & \textbf{85.63 $\pm$ 0.07} & \textbf{79.29 $\pm$ 0.20}\\
     ~ & \textbf{FL+SR}	& \textbf{87.56 $\pm$ 0.29}	& \textbf{85.10 $\pm$ 0.23}	& \textbf{79.07 $\pm$ 0.50}\\
     ~ & \textbf{GCE+SR} & \textbf{87.55 $\pm$ 0.08}	& \textbf{84.69 $\pm$ 0.46}	& \textbf{79.01 $\pm$ 0.18}\\
     \hline\hline
     \multirow{9}*{CIFAR-100} & CE &58.11 $\pm$ 0.32	&50.68 $\pm$ 0.55	&40.17 $\pm$ 1.31\\
     ~ & FL &58.05 $\pm$ 0.42	&51.15 $\pm$ 0.84	&41.18 $\pm$ 0.68\\
    %  ~ & MAE &3.53 $\pm$ 1.62	&4.40 $\pm$ 0.21	&3.33 $\pm$ 1.23	&2.73 $\pm$ 0.12\\
     ~ & GCE &59.35 $\pm$ 1.10	&53.83 $\pm$ 0.64	&40.91 $\pm$ 0.57\\
     ~ & NLNL & 50.19 $\pm$ 0.56 & 42.81 $\pm$ 1.13 & 35.10 $\pm$ 0.20\\
     ~ & SCE &58.16 $\pm$ 0.73	& 50.98 $\pm$ 0.33	&41.54 $\pm$ 0.52\\
     ~ & APL &62.80 $\pm$ 0.05	&56.74 $\pm$ 0.53	&42.61 $\pm$ 0.24\\
     \cline{2-5}
     ~ & \textbf{CE+SR} & \textbf{64.79 $\pm$ 0.01} & \textbf{59.09 $\pm$ 2.10} & \textbf{49.51 $\pm$ 0.59}\\
     ~ & \textbf{FL+SR} & \textbf{64.61 $\pm$ 0.67} & \textbf{58.94 $\pm$ 0.33} & \textbf{46.94 $\pm$ 1.68}\\
      ~ & \textbf{GCE+SR}  & \textbf{64.35 $\pm$ 0.78} &  \textbf{57.22 $\pm$ 0.80} & \textbf{49.51 $\pm$ 1.31}\\
     \bottomrule
\end{tabular}
\vskip-5pt
\end{table}

\noindent\textbf{Results.}\quad The test accuracies (mean $\pm$ std) under symmetric label noise are reported in Table \ref{symmetric-noise}. As we can see, our proposed SR mechanism significantly improves the robustness of CE, FL, and GCE, which achieve the top 3 best results in most test cases across all datasets. In the scenarios of serious noise, our CE+SR, FL+SR have a very obvious improvement over the original losses. For example, on MNIST with 0.8 symmetric noise, CE+SR outperforms CE by more than 52\%. On CIFAR-10 with 0.8 symmetric noise, CE+SR outperforms CE by more than 31\%. As for CIFAR-100, GCE and APL outperform our method by a small gap for $\eta=0.6$ and $0.8$, but they failed in the cases with small noise rates. The reason is that the fitting ability is not enough, which can be derived according to the experiments in the clean case. When APL and GCE meet a complicated dataset CIFAR-100 in a clean setting, their test accuracies are worse than the commonly used losses CE and FL, while the SR-enhanced methods outperform the original losses and achieve an improvement of $\ge 0.86$. Therefore, our methods not only have good robustness, but also guarantee and even improve the fitting ability.

Results for asymmetric noise are reported in Table \ref{asymmetric-noise}. Again, our methods significantly improve the robustness of the original version across all datasets and achieve top 3 best results in most cases. On MNIST, CE+SR and FL+SR outperform all the state-of-the-art methods over all asymmetric noise by a clear margin. More Surprisingly, the test accuracy (99.36 $\pm$ 0.05) of FL+SR under 0.4 asymmetric noise is higher than the clean case. On CIFAR-10 with 0.1 asymmetric noise, SCE has the best accuracy, but it loses the superiority in the other three cases where our method works much better than all other baselines with at least 1\% increase. On CIFAR-100, the enhanced loss functions show particularly superior performance in all cases.

\noindent\textbf{Representations.}\quad We further investigate the representations learned by CE+SR compared to those learned by CE. We extract the high-dimensional features at the second last full-connected layer, then project all test samples' features into 2D embeddings by t-SNE \cite{tsne}. The projected representations on MNIST with different symmetric noise are illustrated in Fig. \ref{fig:tsne}. As can be observed, CE encounters severe overfitting on label noise, and the embeddings look completely mixed when $\eta=0.8$. On the contrary, CE+SR learns good representations with more separated and clearly bounded clusters in all noisy cases.

\subsection{Evaluation on Real-world Noisy Dataset}
Here, we evaluate our sparse regularization method on large-scale real-world noisy dataset WebVision 1.0 \cite{Webvision}. It contains 2.4 million images with real-world noisy labels, which crawled from the web using 1,000 concepts in ImageNet ILSVRC12 \cite{deng2009imagenet}. Since the dataset is very big, for quick experiments, we follow the training setting in \cite{jiang2018mentornet} that only takes the first 50 classes of the Google resized image subset. We evaluate the trained network on the same 50 classes of WebVision 1.0 validation set, which can be considered as a clean validation set. We add sparse regularization to CE and GCE. The training details follow \cite{ma2020normalized}, where for each loss , we train a ResNet-50 \cite{he2016deep} by using SGD for 250 epochs with initial learning rate 0.4, nesterov momentum 0.9, weight decay $3\times 10^{-5}$, and batch size 512. The learning rate is multiplied by 0.97 after every epoch of training. All the images are resized to $224\times 224$. Typical data agumentations including random width/height shift, color jittering, and random horizontal flip are applied. As shown in in Table \ref{webvision}, our proposed SR mechanism obviously enhances the performance of CE and FL, which outperform the existing loss functions SCE and APL with a clear margin ($\ge 2.2\%$). This verifies the effectiveness of SR against real-world label noise.
\begin{table}[htb]
\small
\setlength{\tabcolsep}{1.5mm}
\centering
\caption{Top-1 validation accuracies (\%) on WebVision validatoin set of ResNet-50 models trained on WebVision using different loss functions, under the Mini setting in \cite{jiang2018mentornet}.}

\label{webvision}
\begin{tabular}{c|cccccc}
    \toprule
    Loss & CE & FL & SCE & APL & \textbf{CE+SR}& \textbf{FL+SR}\\
    \midrule
    Acc & 66.96 & 63.80 & 66.92 & 66.32 & \textbf{69.12} & \textbf{70.28}\\ 
    \bottomrule
\end{tabular}
\vskip-6pt
\end{table}

\noindent \textbf{More Comparison.}
We also compare with Co-teaching\cite{han2018co}, which is the representative work of sample selection, and PHuber-CE \cite{menon2010can}, which is a simple variant of gradient clipping. As shown in Table \ref{tab:comp}, our method works better than Co-teaching and PHuber-CE.

\begin{table}[h]
\setlength{\tabcolsep}{1mm}
    \small
    \centering
    \caption{\small{A comparison with other methods }}
    \label{tab:comp}
    \begin{tabular}{c|c|cccc}
        \toprule
        \multirow{2}*{Dataset} & \multirow{2}*{Method} &  \multicolumn{4}{c}{Label Noise Type} \\
        \cmidrule{3-6}
        & & sy 0.6 & sy 0.8 & asy 0.3 & asy 0.4\\
        \midrule
        \multirow{4}*{CIFAR-10} & CE & 38.70 & 19.55 & 79.32 & 74.67\\
        & Co-teaching & 65.74 & 38.01 & 64.01 & 51.26\\
        & PHuber-CE & 75.44 & 41.18 & 76.06 & 55.78\\
        & \textbf{CE+SR} & \textbf{78.18} & \textbf{51.13} & \textbf{85.63} & \textbf{79.29}\\
        \midrule
        \multirow{4}*{CIFAR-100} & CE & 21.39 & 7.59 & 50.68 & 40.17\\
        & Co-teaching & 34.28 & 7.94 & 42.82 & 33.67\\
        & PHuber-CE & 21.54 & 9.33 & 26.91 & 23.43 \\
        & \textbf{CE+SR} & \textbf{44.95} & \textbf{17.35} & \textbf{59.09} & \textbf{49.51} \\
        \bottomrule
    \end{tabular}
\end{table}

\subsection{Additional Experiments}
\noindent \textbf{Imbalanced Classification.} As shown in Table \ref{webvision}, our method achieves the best result on WebVision that has a certain class imbalance. To better show the performance on class imbalance, we additionally test the ability of SR on the pure imbalanced classification task. We first follow the controllable data imbalance in \cite{maas2011learning} to create the imbalanced CIFAR-10/-100 by reducing the number of training examples and keeping the validation set unchanged. We also consider two imbalance types: long-tailed imbalance\cite{cui2019class} and step imbalance \cite{buda2018systematic}. The results shown in Table \ref{tab:imb} are encouraging, where CE with sparse regularization achieves a very significant improvements compared with CE in all cases. Albeit simple, these additional experiments demonstrate that SR can also mitigate class imbalance well.
\vskip-5pt
\begin{table}[h]
\setlength{\tabcolsep}{0.8mm}
    \small
    \centering
    \caption{\small{Validation accuracy on imbalanced CIFAR-10/-100.}}
    \label{tab:imb}
    \begin{tabular}{c|c|cccc}
        \toprule
        \multirow{2}*{Dataset} & \multirow{2}*{Method} &  \multicolumn{4}{c}{Imbalanced Type} \\
        \cmidrule{3-6}
        & & lt-0.01 & lt-0.1 & step-0.01 & step-0.1\\
        \midrule
        \multirow{2}*{CIFAR-10} & CE & 64.16 & 81.81 & 57.44 & 79.35\\
        & \textbf{CE+SR} & \textbf{69.78} & \textbf{84.49} & \textbf{61.03} & \textbf{82.11}\\
        \midrule
        \multirow{2}*{CIFAR-100} & CE & 35.17 & 51.43 & 37.92 & 53.43\\
        & \textbf{CE+SR} & \textbf{41.24} & \textbf{59.51} & \textbf{40.21} & \textbf{58.42}\\
        \bottomrule
    \end{tabular}
    \vskip-15pt
\end{table}

\section{Conclusion and Future Work}
In this paper, we presented a novel method for learning with noisy labels. We first provided a theoretical conclusion that any loss can be made robust to noisy labels by restricting the output of a network to a permutation set of any fixed vector. According to this principle, subsequently, we proposed a simple but effective strategy for robust learning through sparse regularization, which is the approximation of the constraint of one-hot permutation. The meanings of sparse regularization are two-fold: the network output sharpening operation is designed to enforce the output distribution of a network to be sharp, and the $\ell_p$-norm ($p\le 1$) regularization is tailored to promote the network output to be sparse. Experimental results demonstrated the superior performance of the proposed method over the SOTA methods on both synthetic and real-world datasets.
Moreover, we additionally experiment with imbalanced classification, and the results are encouraging, which demonstrates that sparse regularization can also mitigate class imbalance well.

Overall, this paper has investigated the one-hot constraint, \textit{i.e.}, restricting the output to a permutation set over a one-hot vector. In future research, a promising direction is replacing the one-hot vector with a fixed smoothing vector.

\section*{Acknowledgments}
This work was supported by National Natural Science Foundation of China under Grants 61922027, 62071155 and 61932022.

\newpage
\onecolumn
\begin{center}
    \textbf{\Large{Learning with Noisy Labels via Sparse Regularization:\\ Supplementary Materials}}
\end{center}
% \title{Learning with Noisy Labels via Sparse Regularization: Supplementary Materials}

\vskip10pt
\newtheorem{thm}{Theorem}
\newenvironment{thmbis}[1]
  {\renewcommand{\thethm}{#1}%
   \addtocounter{thm}{-1}%
   \begin{thm}}
  {\end{thm}}
\appendix
\section{Proof of Theorems}
\begin{thmbis}{1}
In a multi-class classification problem, $\forall L\in\mathcal L$, $L$ is noise-tolerant under symmetric label noise if $\eta<1-\frac{1}{k}$ and $f:\mathcal X\rightarrow \mathcal P_{\mathbf v}$, where $\mathbf v$ is a fixed vector, i.e., 
\begin{equation}
    \mathop{\arg\min}_{f:\mathcal X\rightarrow \mathcal P_{\mathbf v}} R_{L}(f)=\mathop{\arg\min}_{f:\mathcal X\rightarrow \mathcal P_{\mathbf v}} R^\eta_{L}(f).
\end{equation}
\end{thmbis}
\begin{proof}
For symmetric label noise, we have
$$
\begin{aligned}
R_L^\eta (f)&=\mathbb E_{\mathbf x,y}\left[(1-\eta)L(f(\mathbf x),y)+\frac{\eta}{k-1}\sum_{i\not = y}L(f(\mathbf x), i)\right]\\
&= \mathbb E_{\mathbf x, y}\left[ (1-\eta)L(f(\mathbf x),y) + \frac{\eta}{k-1}(\sum_{i=1}^kL(f(\mathbf x), i) - L(f(\mathbf x), y)\right]\\
&=(1-\eta)R_L(f)+\frac{\eta}{k-1}(\sum_{i=1}^k L(\mathbf v,i)-R_L(f))\\
&=(1-\frac{\eta k}{k-1})R_L(f)+\frac{\eta}{k-1}\sum_{i=1}^k L(\mathbf v,i)
\end{aligned}
$$
since $1-\frac{\eta k}{k-1}>0$ and $\sum_{i=1}^k L(\mathbf v,i)$ is a constant, then $f^*$ minimizes $R^\eta(f)$ if and only if $f^*$ minimizes $R(f)$.
\end{proof}

\begin{thmbis}{2}
In a multi-class classification problem, we let $f:\mathcal X\rightarrow \mathcal P_{\mathbf v}$, where $\mathbf v$ is a fixed vector. If $R_L(f^*)=0$, $\forall f:\mathcal X\rightarrow \mathcal P_{\mathbf v}$, $\forall L\in\mathcal L$ and $0\le L\le \frac{C}{k-1}$, $L$ is noise-tolerant under asymmetric or class-conditional noise when $\eta_{\mathbf y,i}<1-\eta_y$ with $\sum_{k\not =y}\eta_{y,i}=\eta_y$, $\forall \mathbf x$.
\end{thmbis}
\begin{proof}
For asymmetric or class-conditional noise, we have
\begin{equation}
    \begin{aligned}
     R_L^\eta(f)&=\mathbb E_{\mathbf x,y} (1-\eta_{y}) L(f(\mathbf x),y)+\mathbb E_{\mathbf x,y}\sum_{i\not=y}\eta_{y,i}L(f(\mathbf x),i)\\
     &=\mathbb{E}_{\mathbf x,y}(1-\eta_y)(C-\sum_{i\not=y}L(f(\mathbf x),i))+\mathbb E_{\mathbf x,y}\sum_{i\not=y}\eta_{y,i}L(f(\mathbf x),i)\\
     &= C\mathbb{E}_{\mathbf x,y}(1-\eta_y)-\mathbb E_{\mathbf x,y}\sum_{i\not=y}(1-\eta_y-\eta_{y,i})L(f(\mathbf x),i)\\
    \end{aligned}
\end{equation}
Let $f^*_\eta$ and $f^*$ be the minimizer of $R_L^\eta(f)$ and $R_L(f)$ when $f:\mathcal X\rightarrow \mathcal P_{\mathbf v}$, respectively. We have $R_L^\eta(f^*_\eta)-R_L^\eta (f^*)\le 0$ and hence derive that
\begin{equation}
    \mathbb E_{\mathbf x,y}\sum_{i\not=y}(1-\eta_y-\eta_{y,i})(L(f(\mathbf x),i)-L(f(\mathbf x),i))\le 0
    \label{ss}
\end{equation}
Since we are given $R_L(f^*)=0$, we have $L(f^*(\mathbf x),y)=0$. Given the condition on $L$ in the theorem, this implies $L(f^*(\mathbf x),i)=C/(k-1)$, $i \not=y$. As per the assumption on noise in the theorem, $1-\eta_y-\eta_{y,i}>0$. Moreover, $L$ has to satisfy $L(f^*_\eta(\mathbf x),i)\le C/(k-1)$, $\forall i$. Thus for Eq. \ref{ss} to hold, it must be the case that $L(f^*_\eta(\mathbf x),i)=C/(k-1)$, which implies $L(f^*_\eta(\mathbf x),y)=0$. Thus, the minimizer of true risk is also a minimizer of risk under noisy case.
\end{proof}

\begin{thmbis}{3}

In a multi-class classification problem, if the loss function $L$ satisfies $|\sum_{i=1}^k(L(\mathbf u_1,i)-L(\mathbf u_2,i))|\le \delta$ when $\| \mathbf u_1-\mathbf u_2\|_2\le \varepsilon$, and $\delta \rightarrow 0$ as $\varepsilon\rightarrow 0$, then for symmetric label noise satisfying $\eta<1-\frac{1}{k}$, the risk bound can be expressed as
$$
    R_L(f^*_\eta)-R_L(f^*)\le 2c\delta, 
$$
where $c=\frac{\eta}{(1-\eta)k-1}$, $f^*_\eta$ and $f^*$ denote the minimizer of $R_L^\eta(f)$ and $R_L(f)$ when $f\in\mathcal H_{\mathbf v, \varepsilon}$, respectively.
\end{thmbis}
\begin{proof}
For symmetric label noise, we have
\begin{equation}
    \begin{aligned}
    R_L^\eta(f^*)&=\mathbb E_{\mathbf x,y}\big[(1-\eta)L(f^*(\mathbf x),y)+\frac{\eta}{k-1}\sum_{i\not = y}L(f^*(\mathbf x), i)\big]\\
    &=(1-\frac{\eta k}{k-1})R_L(f^*) + \frac{\eta}{k-1}\mathbb E_{\mathbf x,y}\left[\sum_{i=1}^k L(f^*(\mathbf x),i) \right]\\
    &=(1-\frac{\eta k}{k-1})R_L(f^*) + \frac{\eta}{k-1}\sum_{i=1}^k L(\mathbf v,i) + \frac{\eta}{k-1} \delta_1
    \end{aligned}
\end{equation}
where $\delta_1=\mathbb E_{\mathbf x,y}[\sum_{i=1}^k L(f(\mathbf x),i) - \sum_{i=1}^k L(\mathbf v,i)]$. On the other hand, $f^*\in\mathcal H_{\mathbf v,\varepsilon}$, \textit{i.e.}, $\|f^*(\mathbf x)-\mathbf v\|_2\le \varepsilon$, so we have $|\sum_{i=1}^k L(f(\mathbf x),i) - \sum_{i=1}^k L(\mathbf v,i)|\le \delta$. This means that $\delta_1\in[-\delta, \delta]$. Similarly, we can obtain
\begin{equation}
    R_L^\eta(f^*_\eta) = (1-\frac{\eta k}{k-1})R_L(f^*_\eta) + \frac{\eta}{k-1}\sum_{i=1}^k L(\mathbf v,i) + \frac{\eta}{k-1}\delta_2
\end{equation}

Since $f^*_\eta=\mathop{\arg\min}_{f\in\mathcal H_{\mathbf v,\varepsilon}} R_L^\eta(f)$, and $f^*=\mathop{\arg\min}_{f\in\mathcal H_{\mathbf v,\varepsilon}} R_L(f)$, we have
\begin{equation}
    \begin{aligned}
     &0\ge R_L^\eta(f^*_\eta)-R_L^\eta(f^*)\\
     &=(1-\frac{\eta k}{k-1})(R_L(f^*_\eta) - R_L(f^*))+\frac{\eta}{k-1}(\delta_2-\delta_1)\\
     \Rightarrow & R_L(f^*_\eta)-R_L(f^*)\le \frac{\eta}{(1-\eta)k-1}(\delta_1-\delta_2)\le  \frac{2\eta\delta}{(1-\eta)k-1}
    \end{aligned}
\end{equation}
where we have used the fact that $1-\frac{\eta k}{k-1}>0$, and $\delta_2-\delta_1\le 2\delta$ holds for $\delta_1,\delta_2\in[-\delta,\delta]$.
\end{proof}

\section{Experiments}
In this section, we provide the experimental details.

\noindent \textbf{Datasets.}\quad We verify the effectiveness of our method on benchmark datasets, including MNIST \cite{MNIST}, CIFAR-10/-100 \cite{CIFAR} with synthetic label noise.

Since MNIST, CIFAR-10, and CIFAR-100 are clean, following previous works \cite{SCE, ma2020normalized}, we experiment with two types of label noise: symmetric (uniform) noise and asymmetric (class-conditional) noise. For symmetric noise, we corrupt the training labels by flipping the labels in each class randomly to incorrect labels in other classes with flip probability $\eta\in\{0.2, 0.4, 0.6, 0.8\}$. For asymmetric noise, we flip the labels within a specific set of classes, for example, for MNIST, flipping 2 $\rightarrow$ 7, 7 $\rightarrow $ 1, 5 $\leftrightarrow$ 6, and 3 $\rightarrow$ 8; for CIFAR-10, flipping TRUCK $\rightarrow$ AUTOMOBILE, BIRD $\rightarrow$ AIRPLANE, DEER $\rightarrow $ HORSE, and CAT $\leftrightarrow$ DOG; for CIFAR-100, the 100 classes are grouped into 20 super-classes with each has 5 sub-classes, and each class are flipped within the same super-classes into the next.

\noindent \textbf{Baselines.}\quad We experiment with the following state-of-the art methods, and two effective loss functions CE and Focal Loss (FL) \cite{lin2017focal} for classification. Moreover, we add the proposed sparse regularization mechanism to CE, FL and GCE, i.e., CE+SR, FL+SR and GCE+SR. All the implementations and experiments are based on PyTorch.
\vskip-5pt
\begin{itemize}
    \item GCE \cite{GCE}. The Generalized Cross Entropy (GCE) is defined as $L_{GCE}(\mathbf u,i)=(1-u_i^q)/q$ ($0<q\le1$).
    \vskip-5pt
    \item SCE \cite{SCE}. The Symmetric Cross Entropy (SCE) can be regarded as a weighted loss of CE and RCE (scaled MAE): $L_{SCE}(\mathbf u,i)=\alpha L_{CE}(\mathbf u,i)+\beta L_{RCE}(\mathbf u,i)$.
    \vskip-5pt
    \item NLNL \cite{kim2019nlnl}. NLNL improves robustness with a complementary label.
    \vskip-5pt
    \item APL \cite{ma2020normalized}. The Active Passive Loss (APL) was proposed to combine a robust active loss and a robust passive loss, i.e., $L_{APL}=\alpha L_{Active}+\beta L_{Passive}$.
\end{itemize}

\noindent\textbf{Network Structure and Training Details.}\quad Following the setting in \cite{ma2020normalized}, we use a 4-layer CNN for MNIST, an 8-layer CNN for CIFAR-10 and a ResNet-34 for CIFAR-100. The networks are trained for 50, 120, 200 epochs for MNIST, CIFAR-10, CIFAR-100, respectively. For all the training, we use SGD optimizer with momentum 0.9 and cosine learning rate annealing. Weight decay is set as $1\times 10^{-3}$, $1\times 10^{-4}$, $1\times 10^{-5}$ for MNIST, CIFAR-10, CIFAR-100, respectively. The initial learning rate is set to 0.01 for CIFAR-10 and 0.1 for CIFAR-100. Batch size is set to 128. Typical data augmentations including random width/height shift and horizontal flip are applied.

\noindent\textbf{Parameters Setting.}\quad We set the parameters which match their original papers for all baseline methods. Specifically, for FL, we set $\gamma=0.3$. For GCE, we set $q=0.7$. For SCE, we set $A=-3$, and $\alpha=0.01$, $\beta=1$ for MNIST, $\alpha=0.1$, $\beta=1$ for CIFAR-10, $\alpha=6$, $\beta=0.1$ for CIFAR-100. For APL (NCE+MAE), we set $\alpha=1$, $\beta=100$ for MNIST, $\alpha,\beta=1$ for CIFAR-10, and $\alpha=10$, $\beta=0.1$ for CIFAR-100. For our sparse regularization, we set $(\tau, p, \lambda_0, \rho, r)=(0.1, 0.1, 4, 2, 5)$ for MNIST, $(0.5, 0.1, 1.1, 1.03, 1)$ for CIFAR-10, and $(0.5, 0.01, \cdot, 1.02, 1)$ for CIFAR-100. Otherwise, on CIFAR-100, we set $\lambda_0$ to $10$ and $4$ for symmetric and asymmetric label noise, respectively.

As for the parameter setting for Webvision, we use the suggested $q=0.7$ for GCE, $A=-4$, $\alpha=10$, $\beta=1$ for SCE, while for APL, we set $\alpha=50$, $\eta=0.1$. For our CE+SR and FL+SR, we set $\tau=0.5$, $p=0.01$, $\lambda_0=2$, $\rho=1.02$ and $f=1$.

\noindent \textbf{More experiments about hyperparameter selection.}\quad We offer more experimental results on selecting different $\lambda$ on CIFAR-10 with 0.6 symmetric label noise. We adjust $\tau$ from 1.0 to 0.5. The results are shown in Fig. \ref{lambda}. We found that the output sharpening can benefit the sparse regularization. We can achieve the similar robustness result of $\lambda=20$ ($\tau=1$) by setting $\lambda=5$ and $\tau=0.5$, which demonstrates that the output sharpening also plays the role of sparse regularization when using $\ell_p$-norm. Moreover, smaller $\lambda$ can help maintain the fitting ability of the model with classification loss $L(f(\mathbf x),y)$ (\textit{i.e.}, learning efficiently while keeping robustness). As a evidence, the eventual accuracy ($\tau=0.5$, $\lambda=5$) is higher than the experiments with $\tau=1.0$.

\begin{figure*}[htb]
    \centering
    \begin{minipage}[htbp]{0.19\linewidth}
        \centering
        \subfigure[$\eta=0.0$]{
        \includegraphics[width=1.2in]{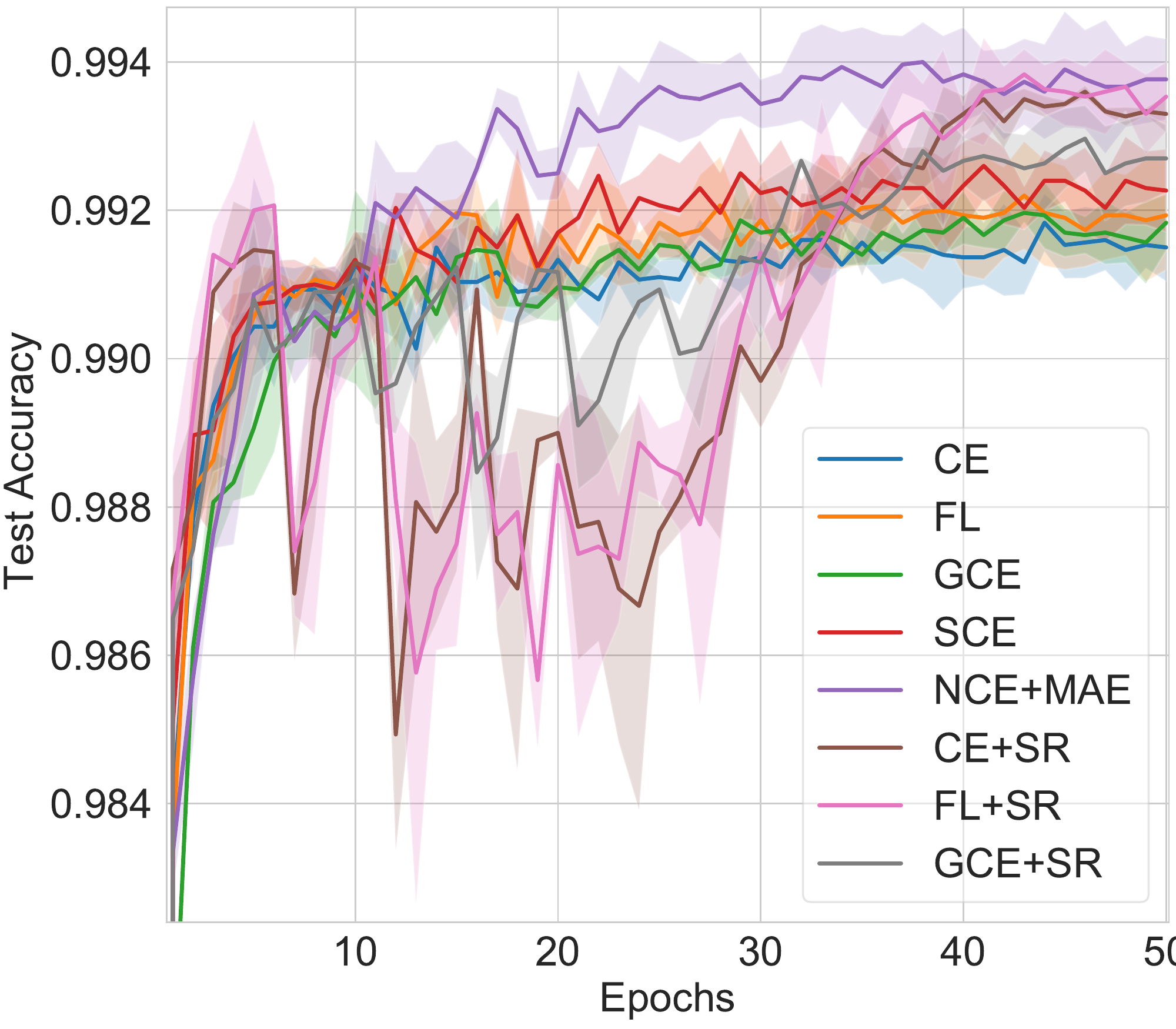}
    }
    \end{minipage}
    \begin{minipage}[htbp]{0.76\linewidth}
        \subfigure[$\eta=0.2$]{
        \includegraphics[width=1.2in]{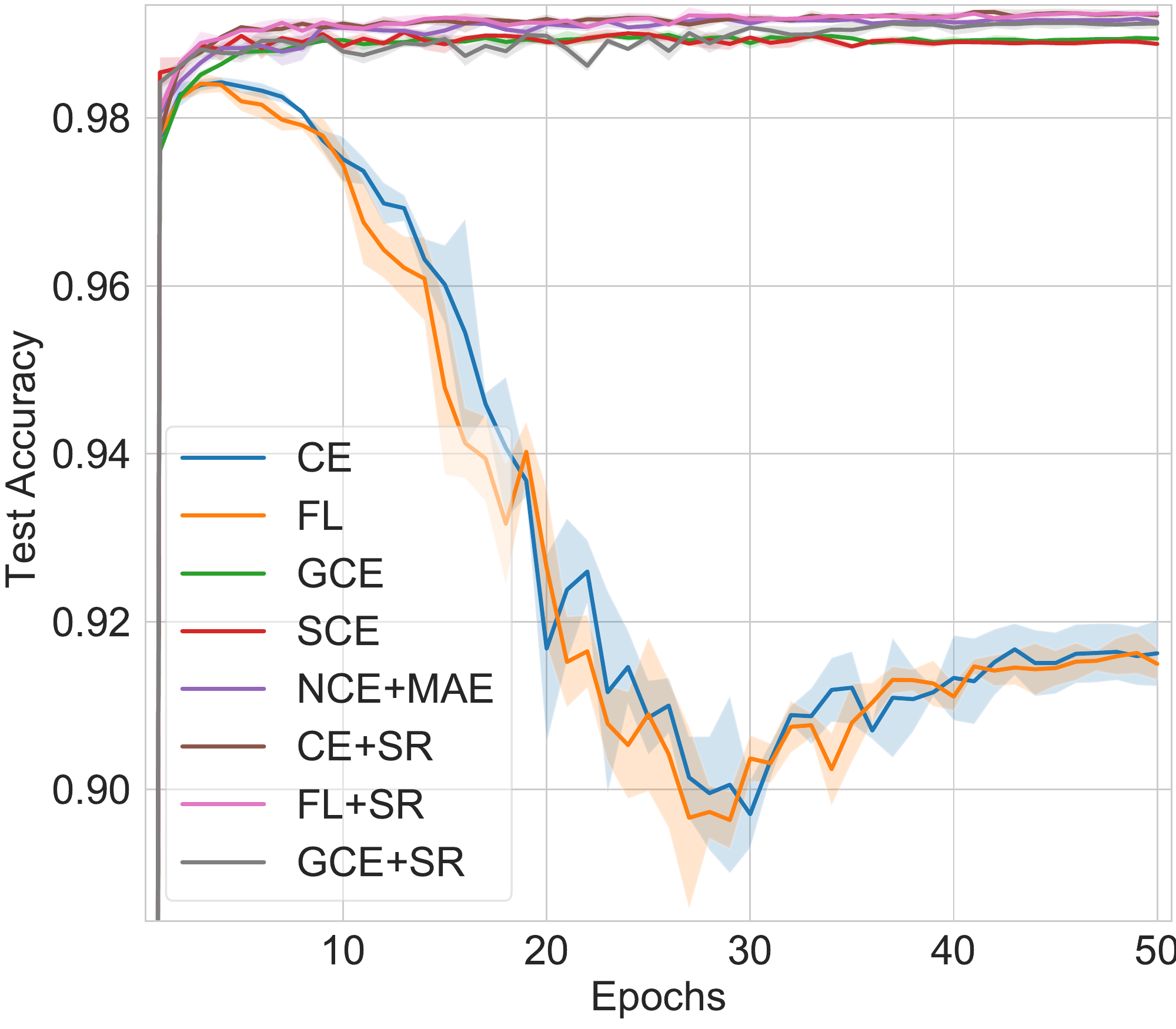}
        }
        \subfigure[$\eta=0.4$]{
        \includegraphics[width=1.2in]{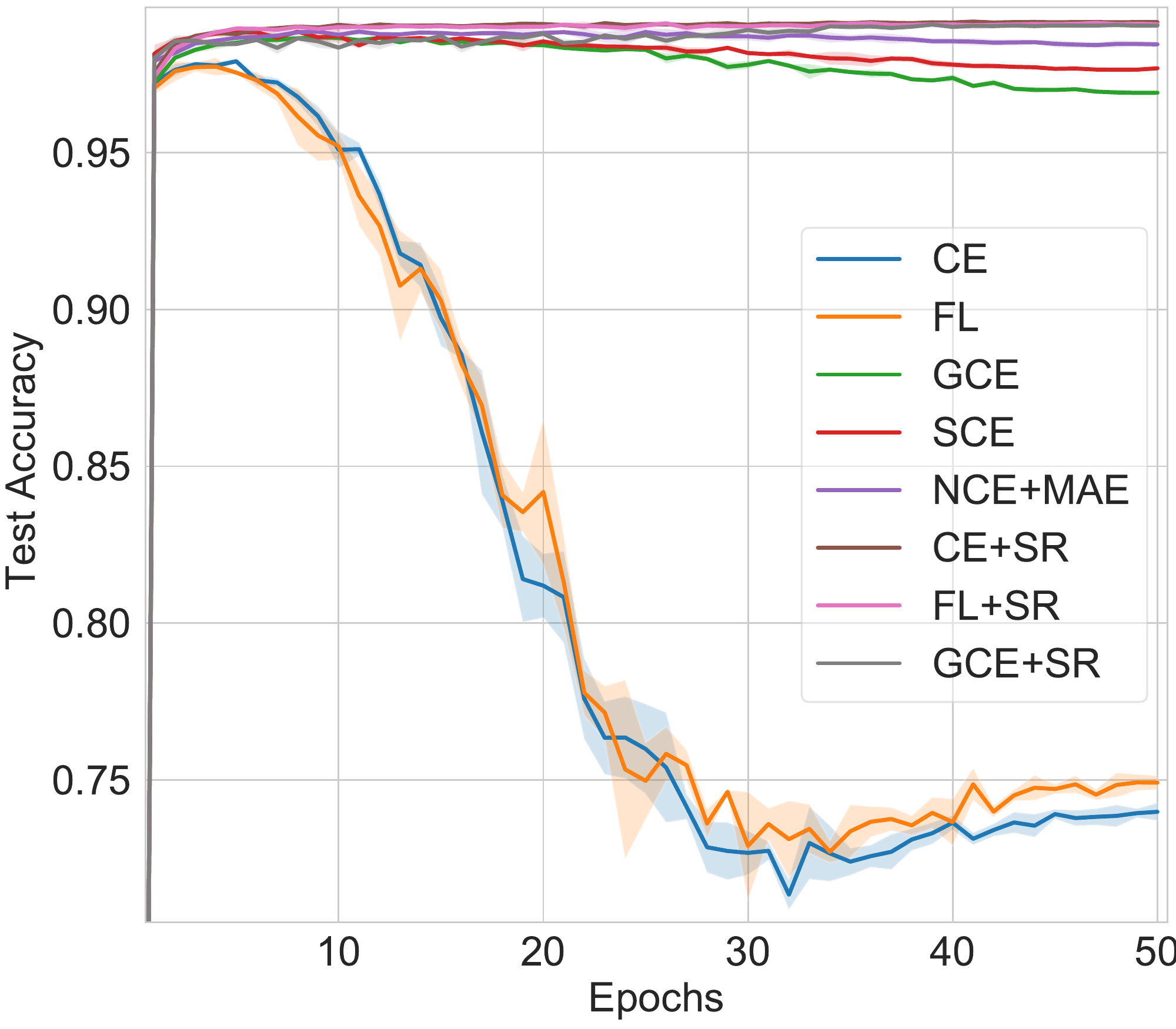}
        }
        \subfigure[$\eta=0.6$]{
        \includegraphics[width=1.2in]{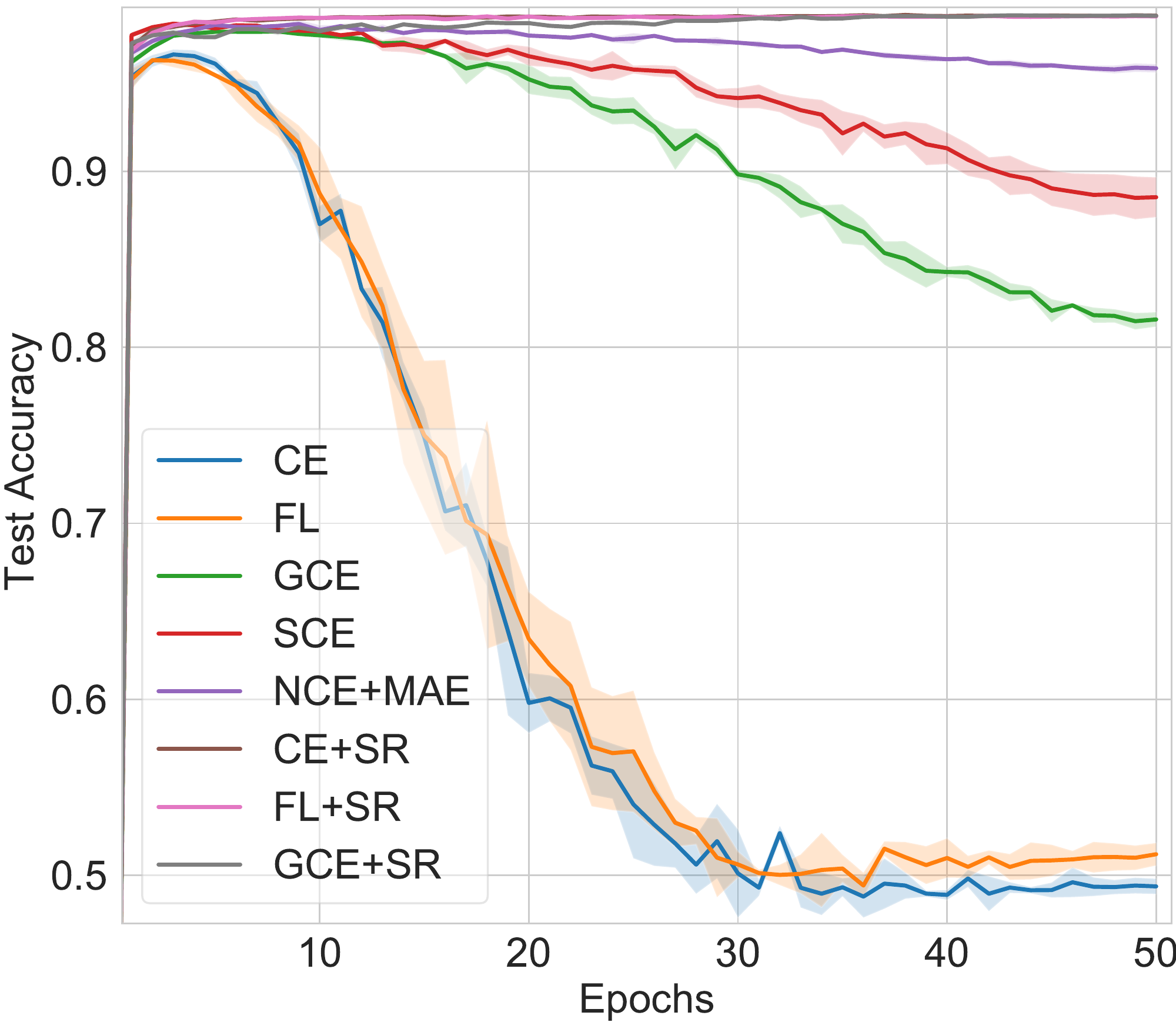}
        }
        \subfigure[$\eta=0.8$]{
        \includegraphics[width=1.2in]{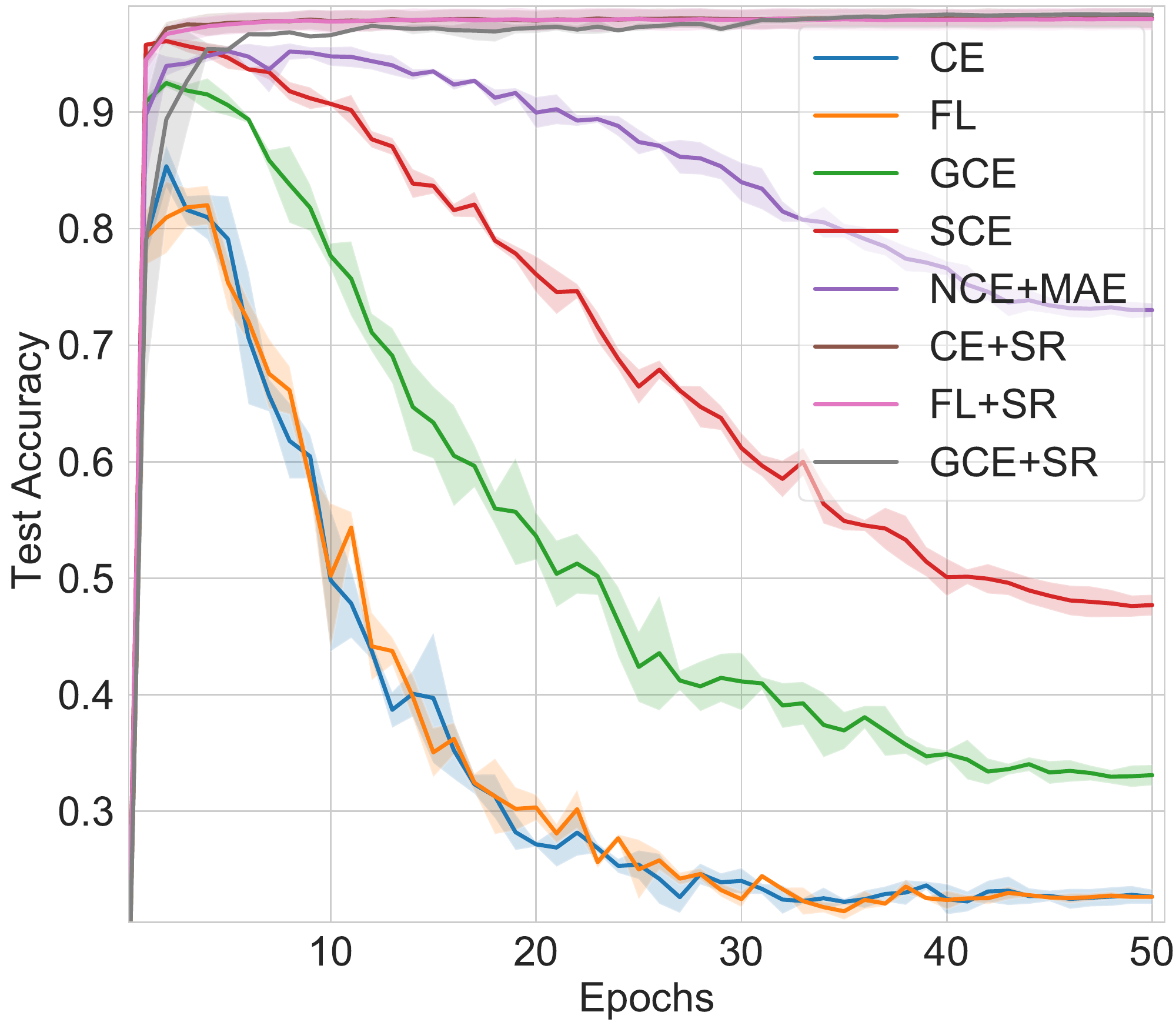}
        }
        \\
        \subfigure[$\eta=0.1$]{
        \includegraphics[width=1.2in]{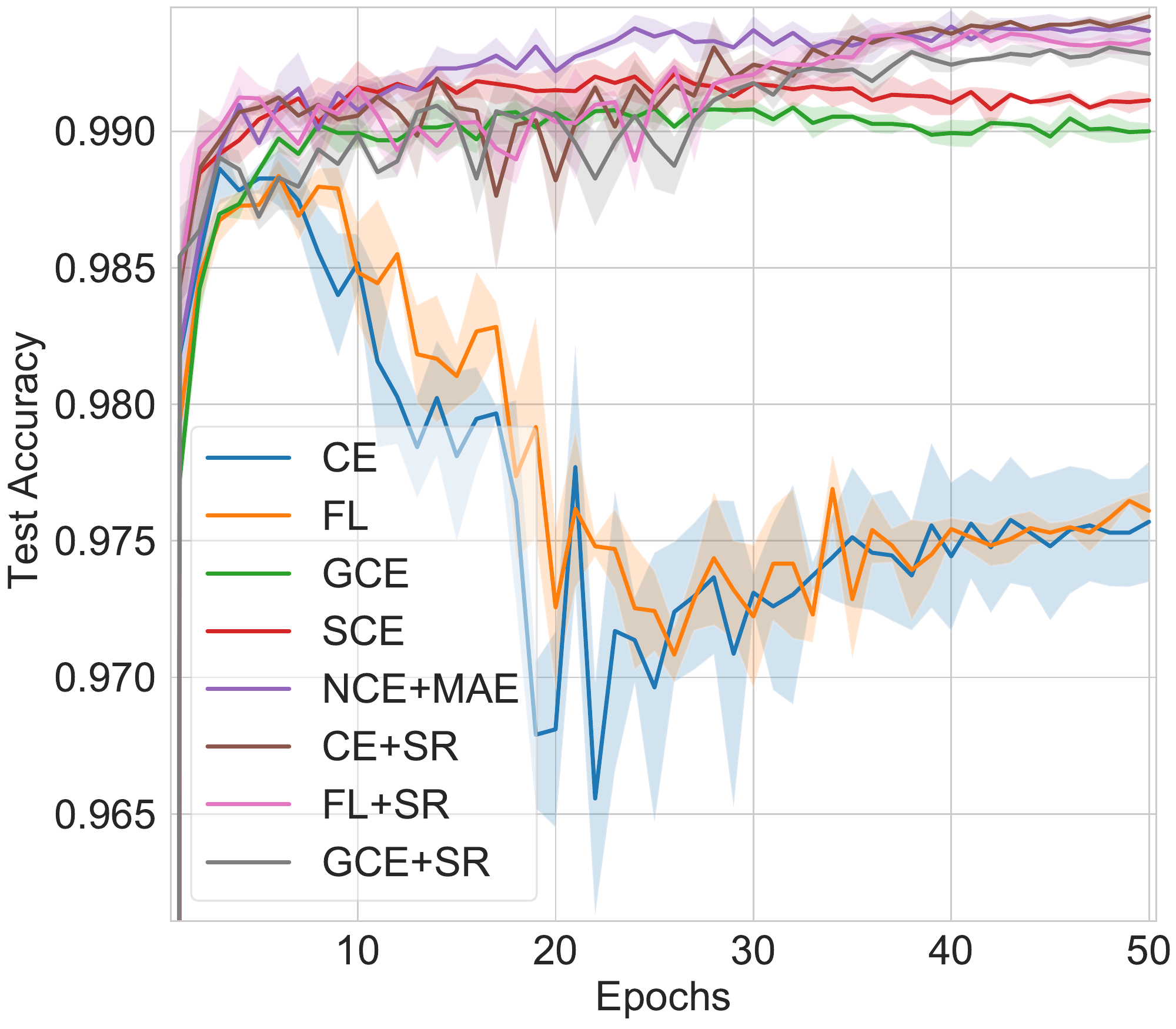}
        }
        \subfigure[$\eta=0.2$]{
        \includegraphics[width=1.2in]{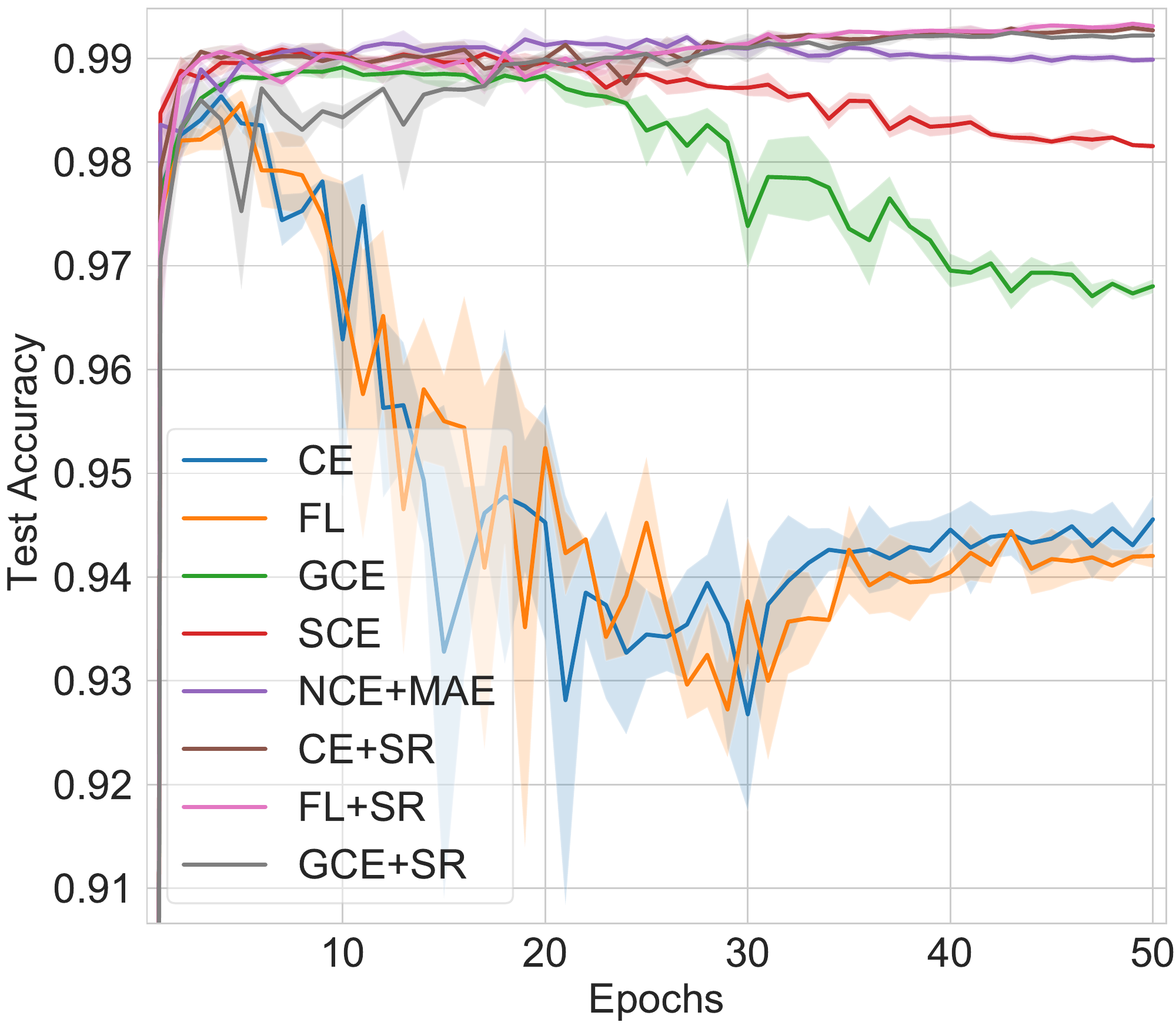}
        }
        \subfigure[$\eta=0.3$]{
        \includegraphics[width=1.2in]{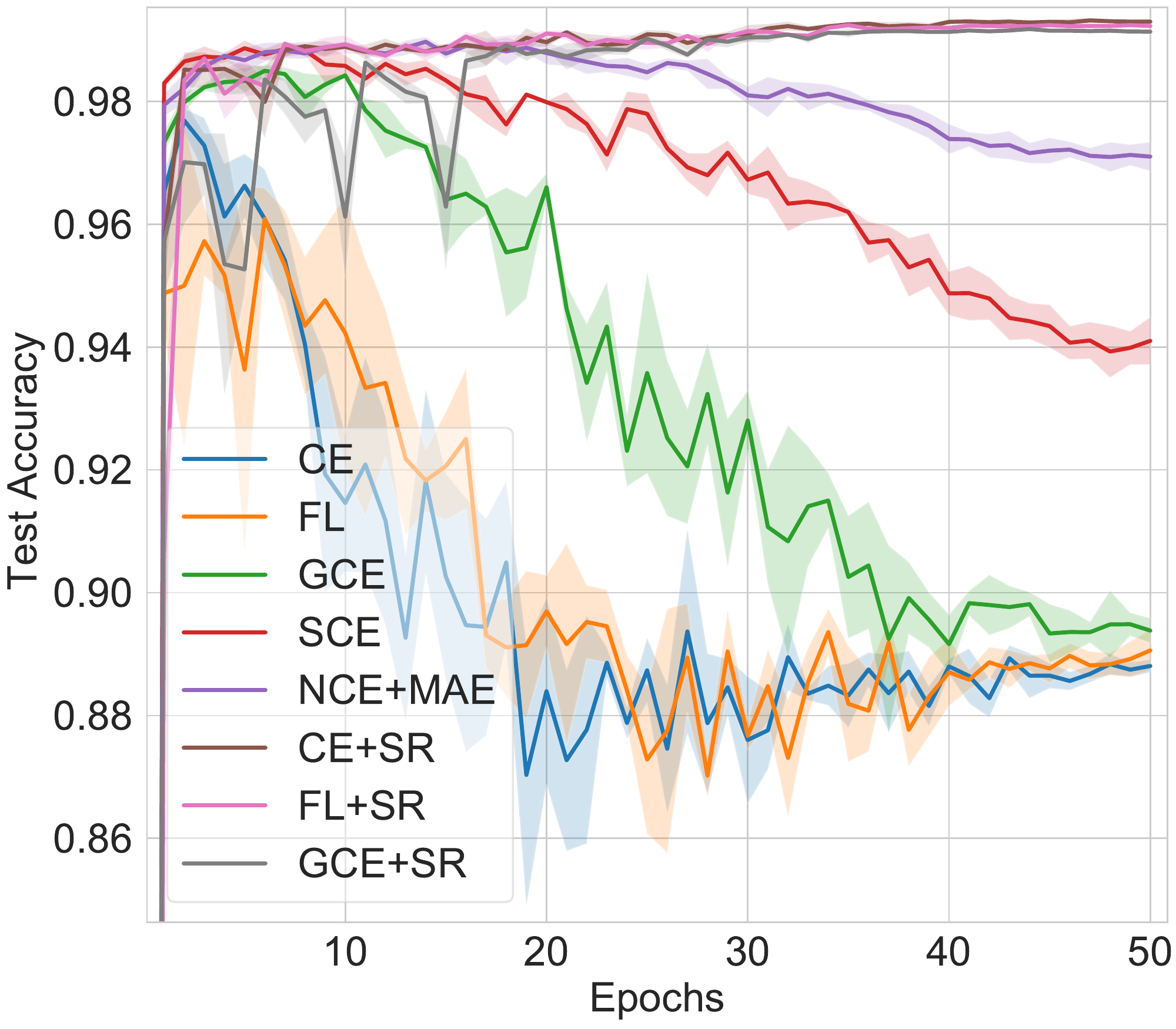}
        }
        \subfigure[$\eta=0.4$]{
        \includegraphics[width=1.2in]{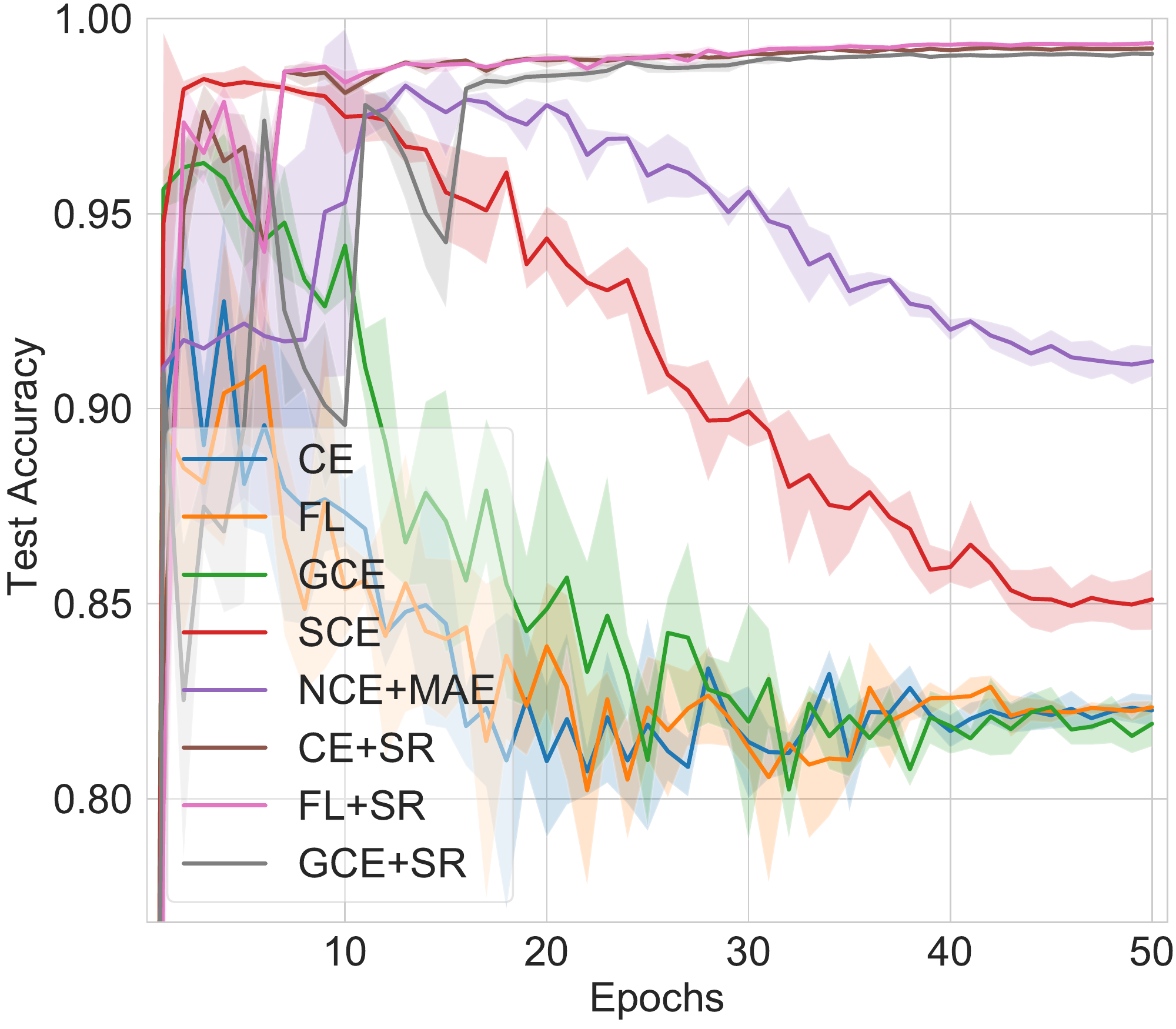}
        }
    \end{minipage}
    \caption{Test accuracies of different methods on MNIST with different label noise, where (a) denotes the clean case, (b-e) denote the symmetric label noise, and (f-i) denote the asymmetric label noise.}
    \label{mnist-acc}
\end{figure*}

\begin{figure*}[htb]
    \centering
    \begin{minipage}[htbp]{0.19\linewidth}
        \centering
        \subfigure[$\eta=0.0$]{
        \includegraphics[width=1.2in]{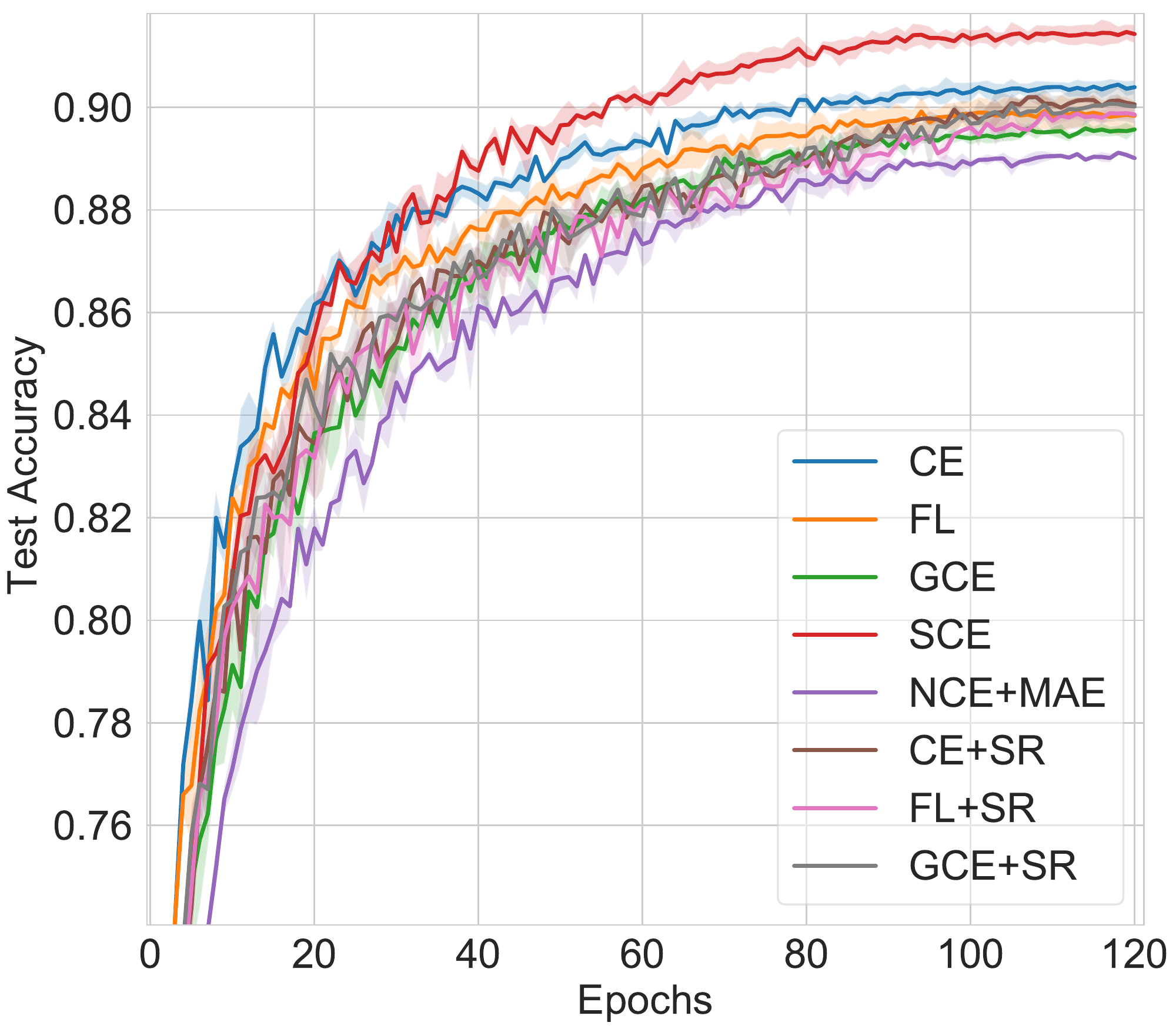}
    }
    \end{minipage}
    \begin{minipage}[htbp]{0.76\linewidth}
        \subfigure[$\eta=0.2$]{
        \includegraphics[width=1.2in]{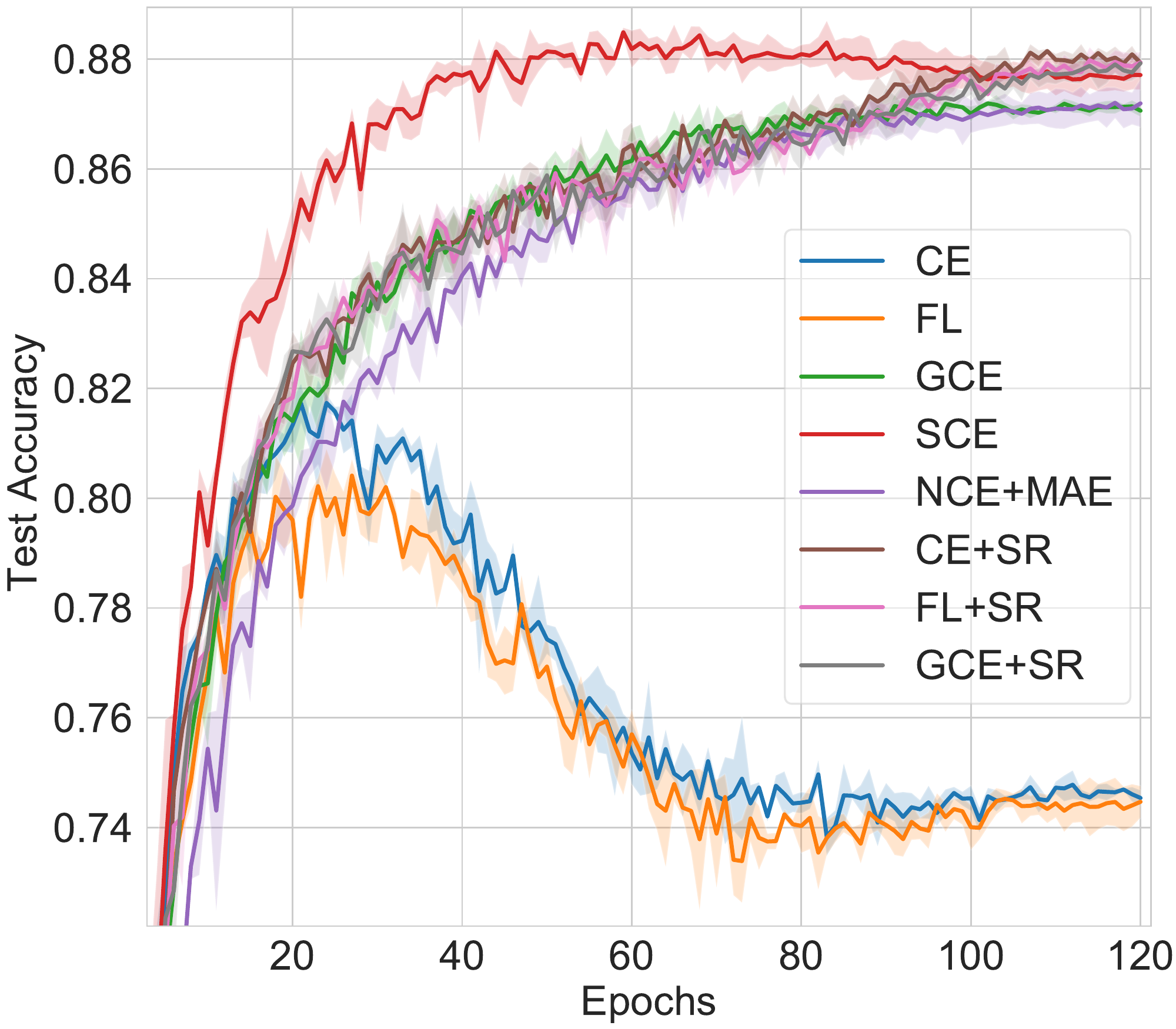}
        }
        \subfigure[$\eta=0.4$]{
        \includegraphics[width=1.2in]{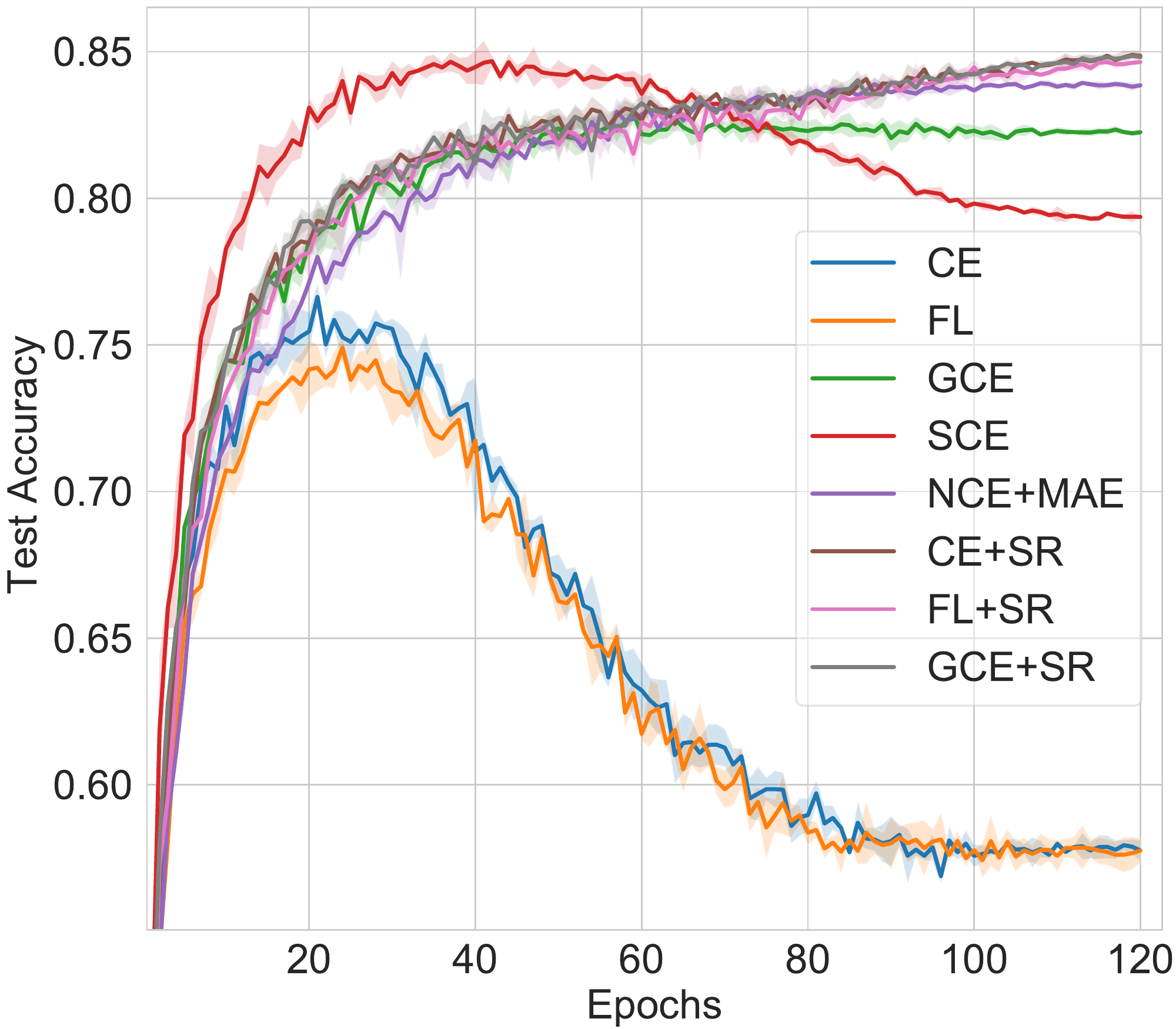}
        }
        \subfigure[$\eta=0.6$]{
        \includegraphics[width=1.2in]{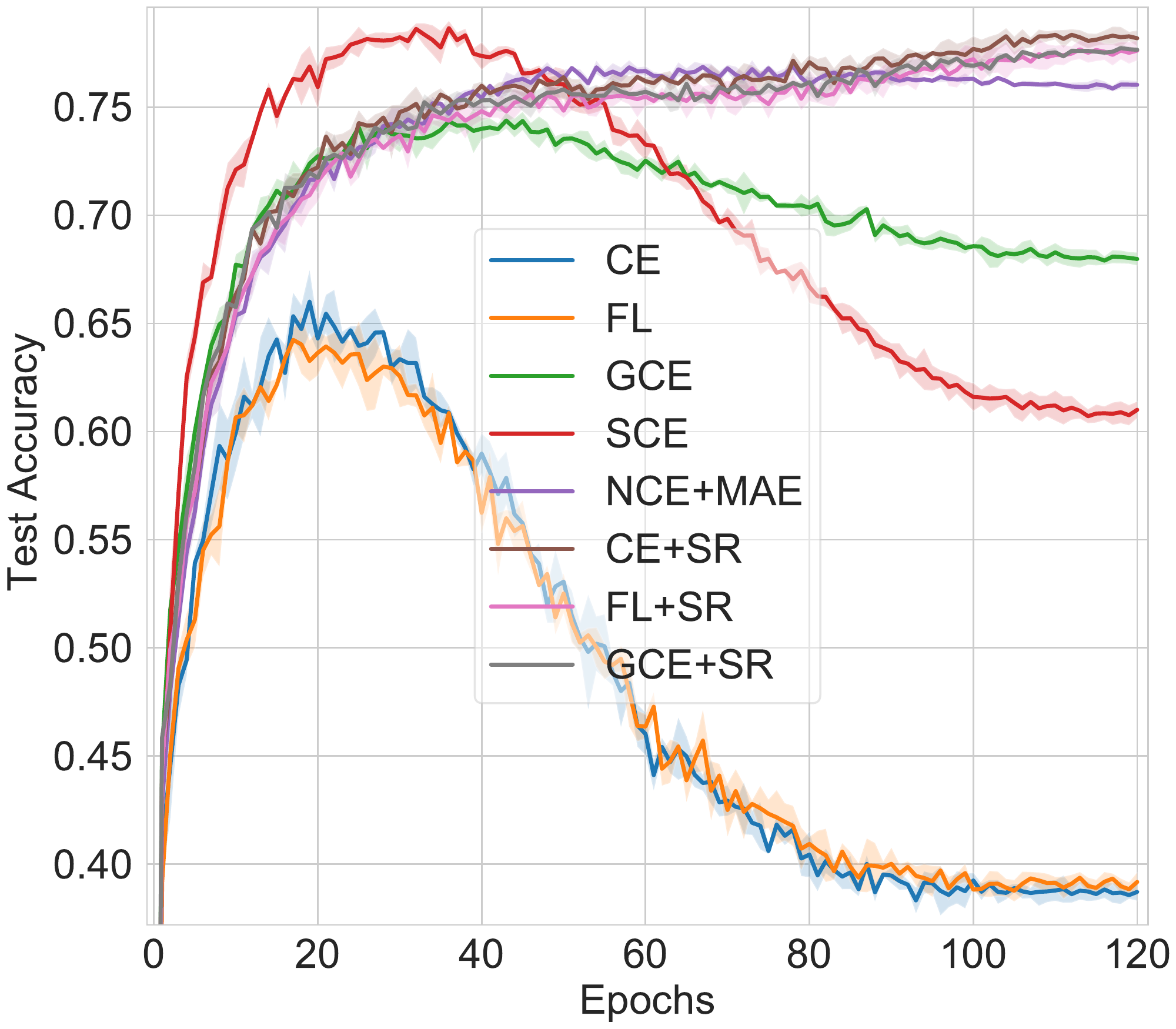}
        }
        \subfigure[$\eta=0.8$]{
        \includegraphics[width=1.2in]{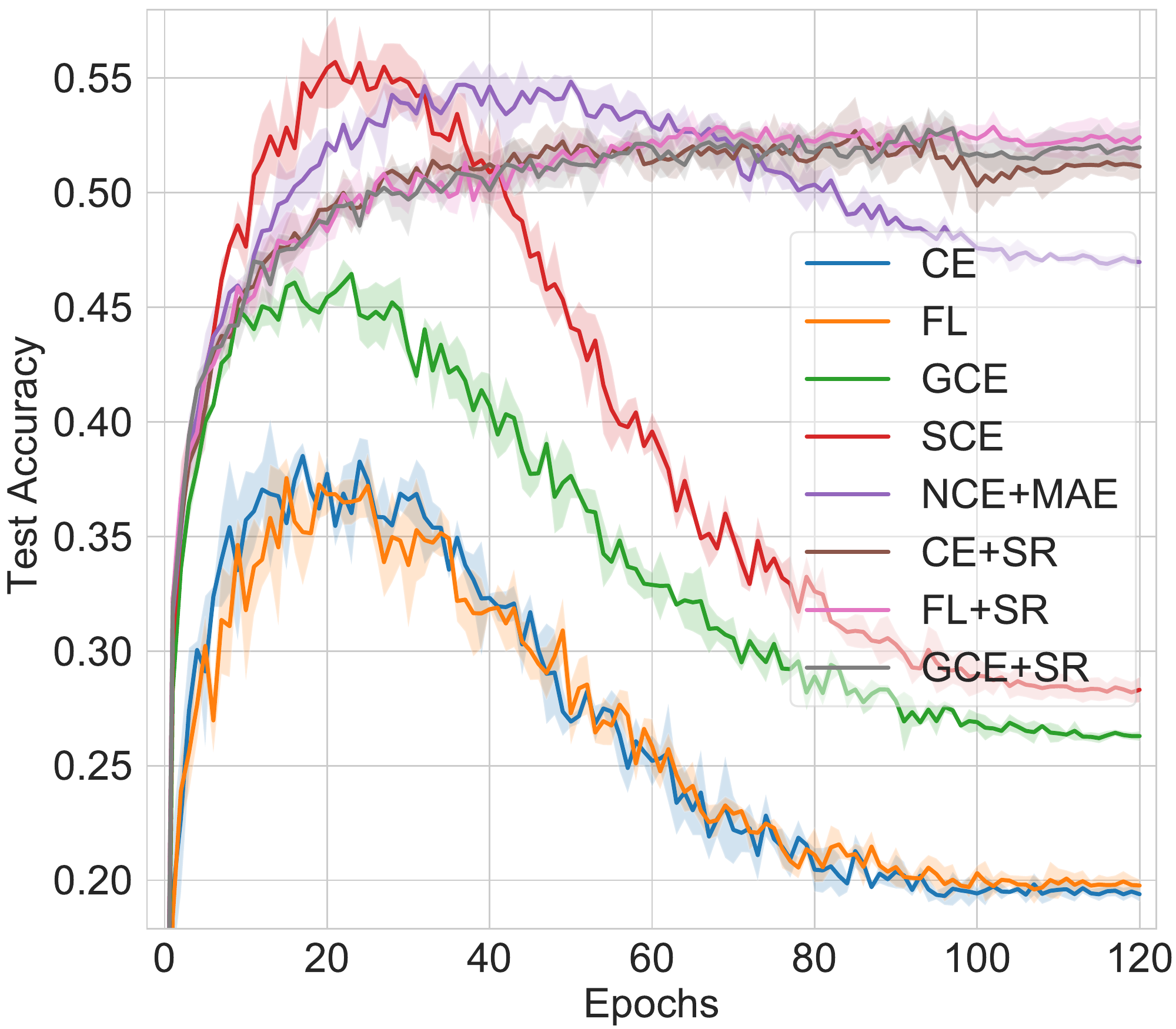}
        }
        \\
        \subfigure[$\eta=0.1$]{
        \includegraphics[width=1.2in]{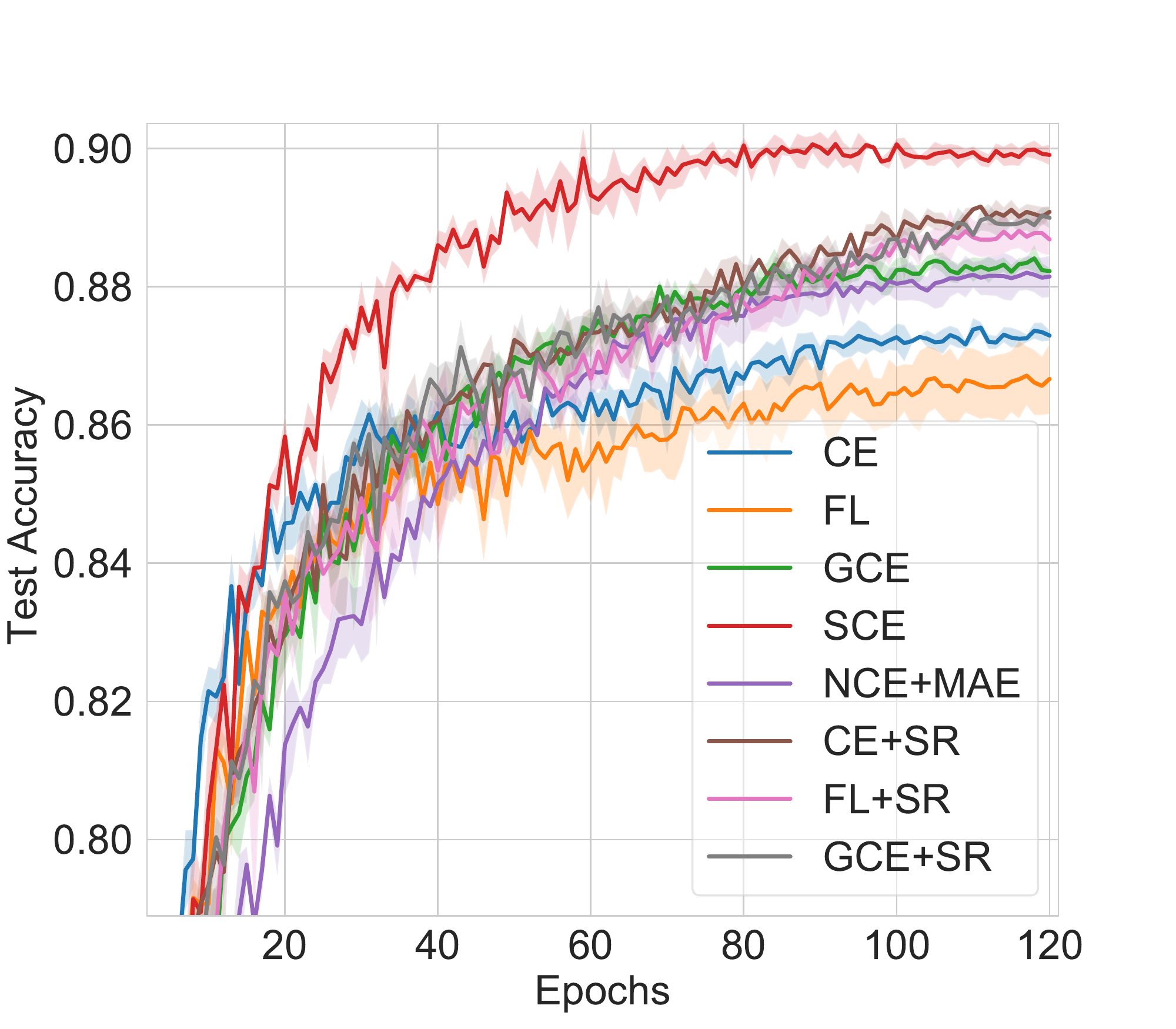}
        }
        \subfigure[$\eta=0.2$]{
        \includegraphics[width=1.2in]{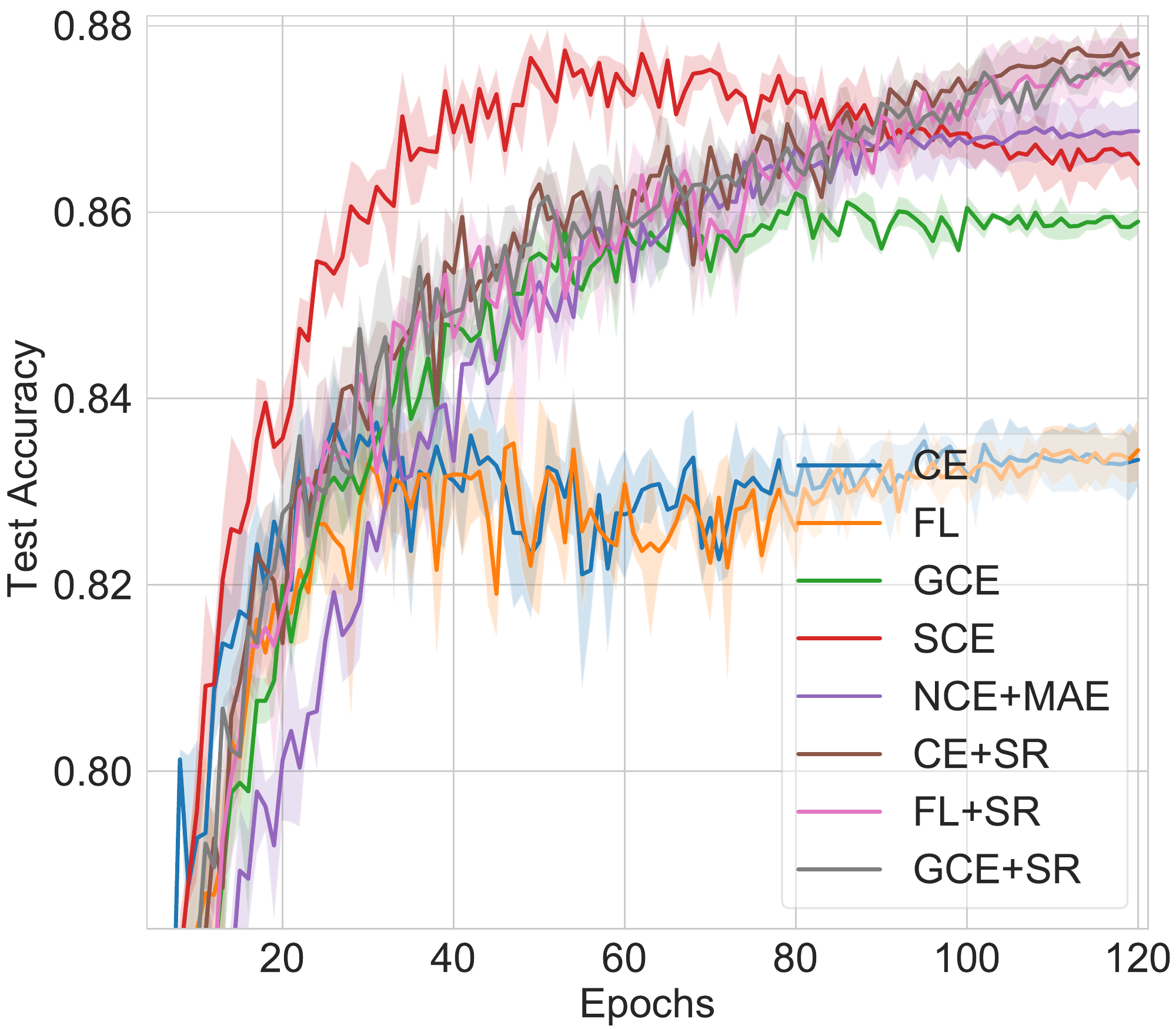}
        }
        \subfigure[$\eta=0.3$]{
        \includegraphics[width=1.2in]{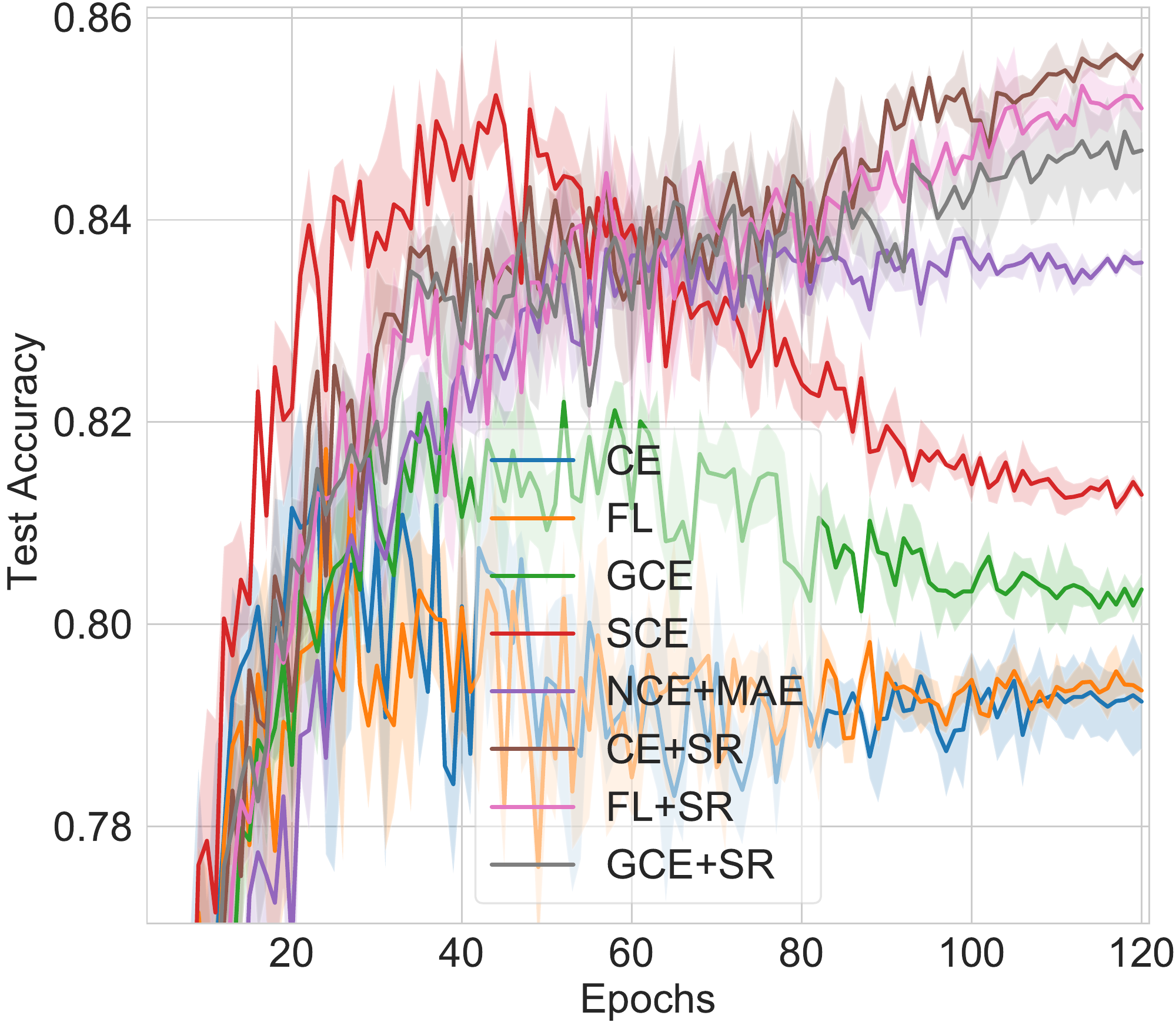}
        }
        \subfigure[$\eta=0.4$]{
        \includegraphics[width=1.2in]{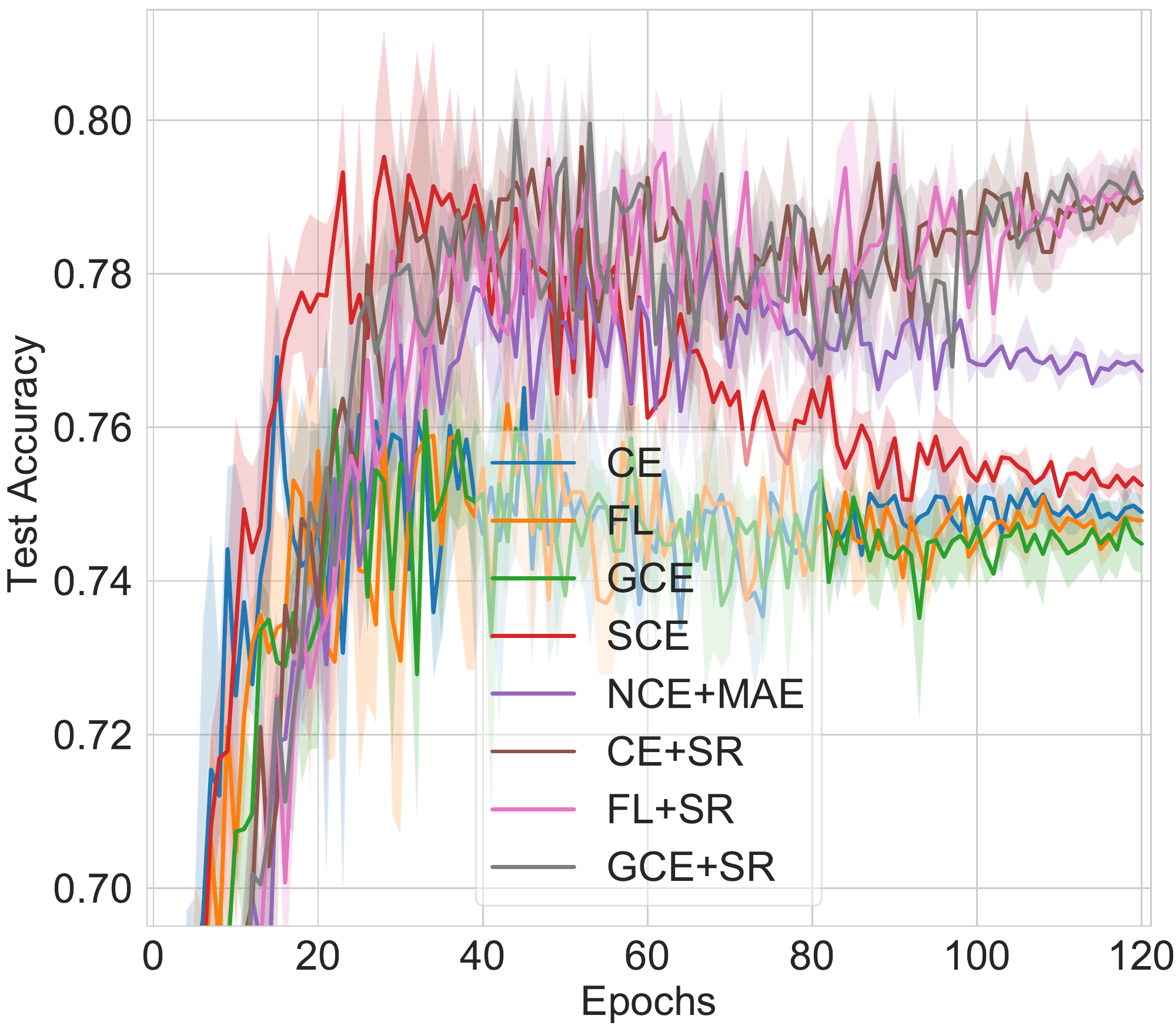}
        }
    \end{minipage}
    \caption{Test accuracies of different methods on CIFAR-10 with different label noise, where (a) denotes the clean case, (b-e) denote the symmetric label noise, and (f-i) denote the asymmetric label noise.}
    \label{cifar10-acc}
\end{figure*}

\begin{figure*}[htb]
    \centering
    \begin{minipage}[htbp]{0.19\linewidth}
        \centering
        \subfigure[$\eta=0.0$]{
        \includegraphics[width=1.2in]{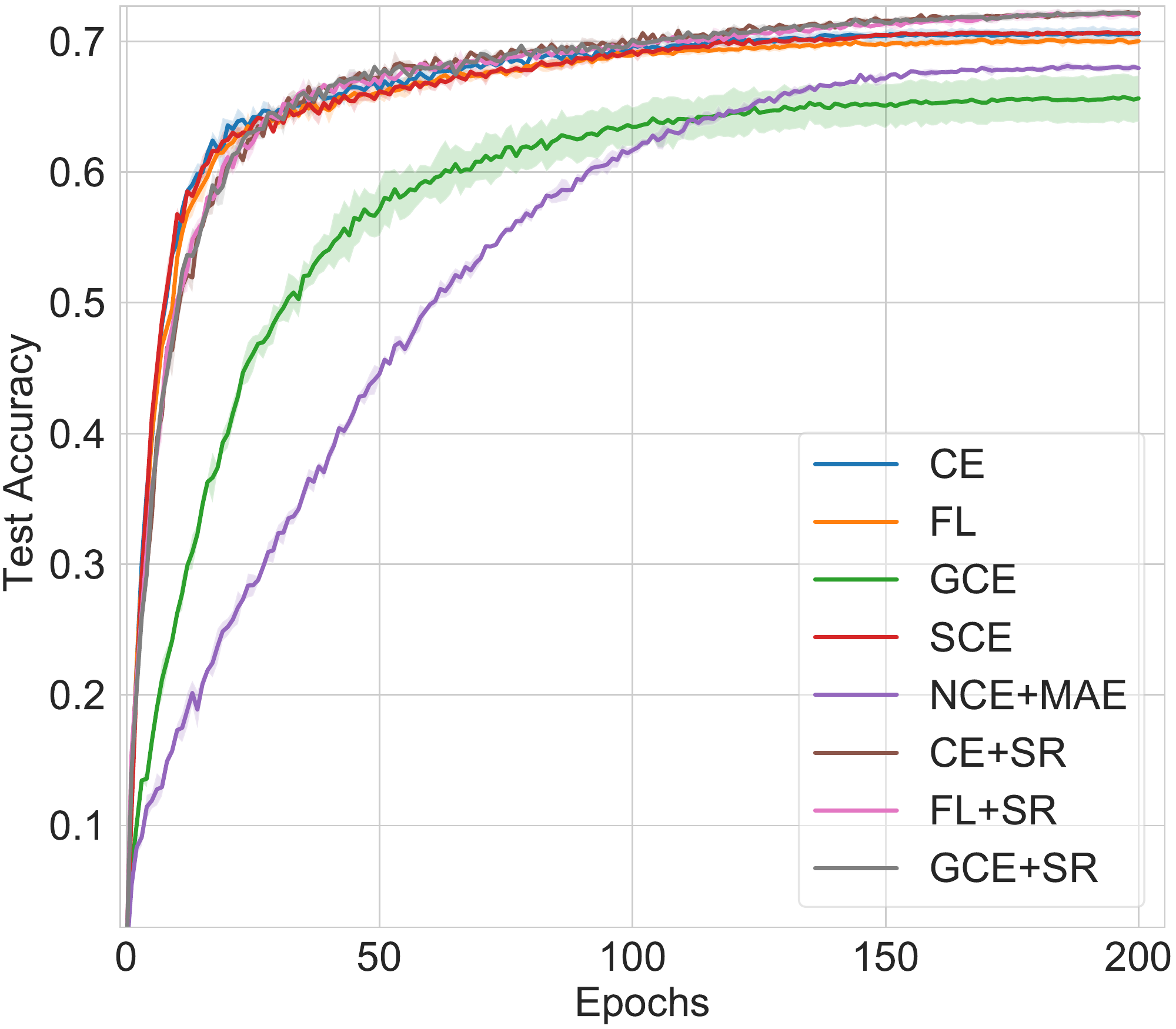}
    }
    \end{minipage}
    \begin{minipage}[htbp]{0.76\linewidth}
        \subfigure[$\eta=0.2$]{
        \includegraphics[width=1.2in]{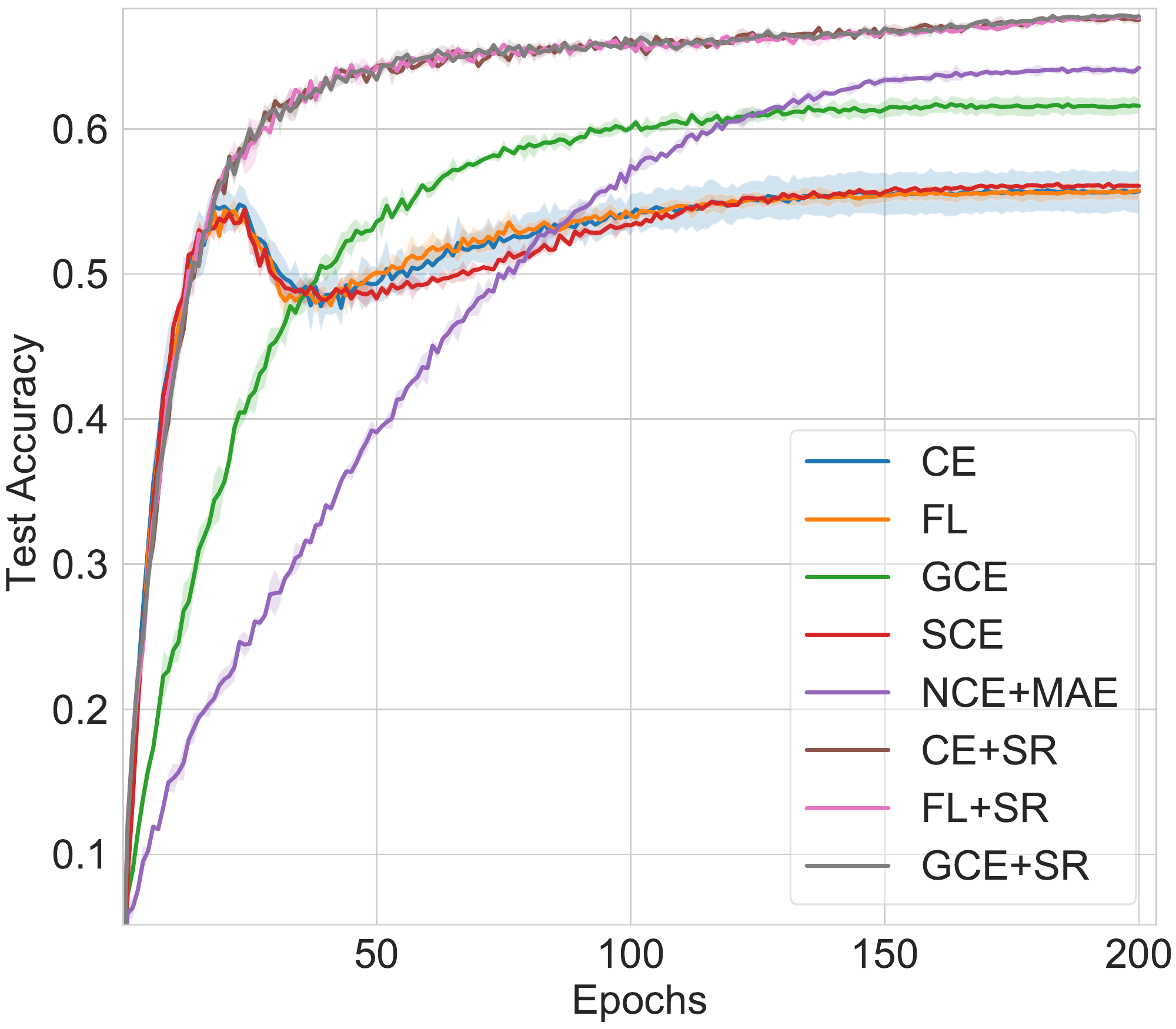}
        }
        \subfigure[$\eta=0.4$]{
        \includegraphics[width=1.2in]{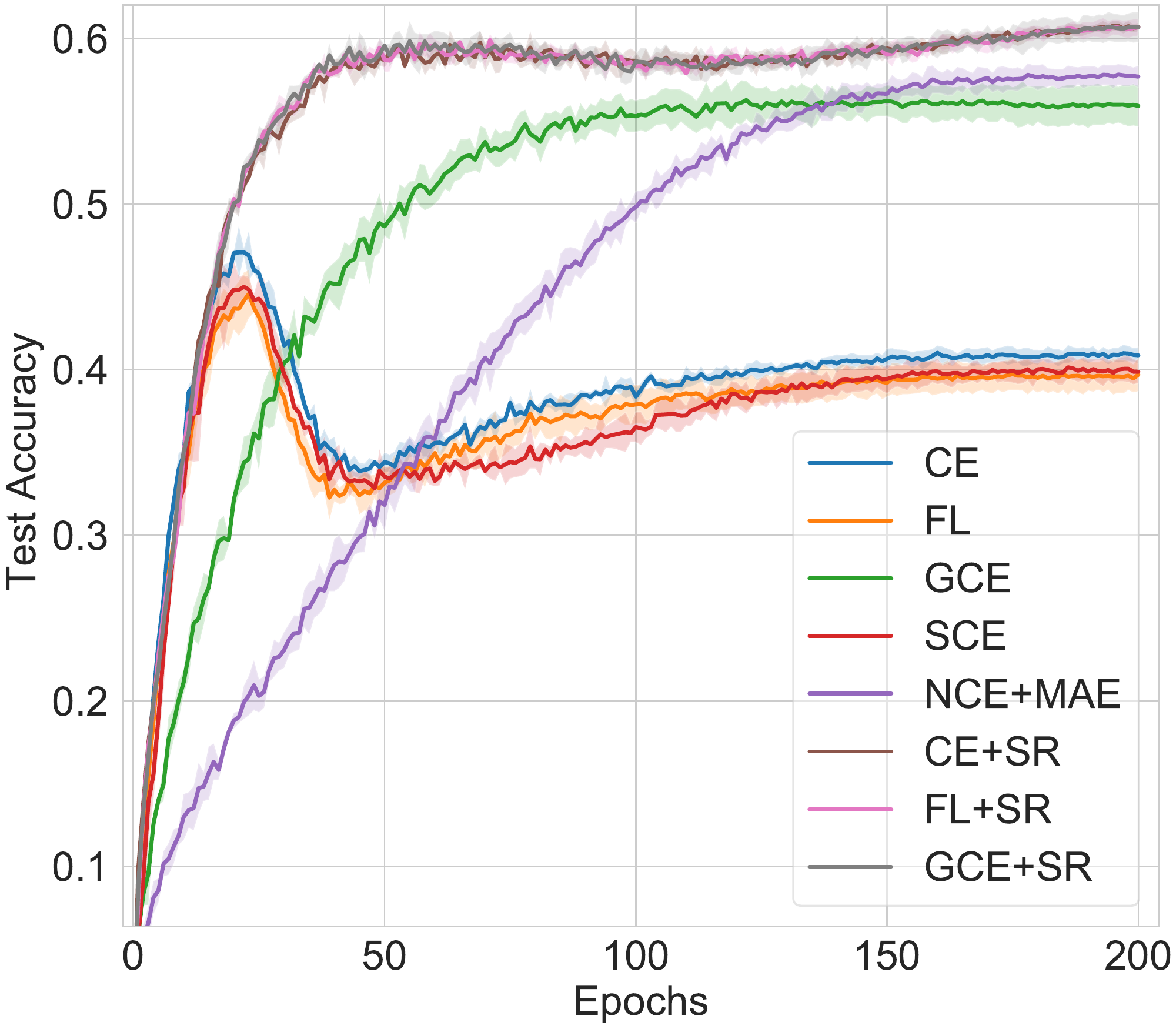}
        }
        \subfigure[$\eta=0.6$]{
        \includegraphics[width=1.2in]{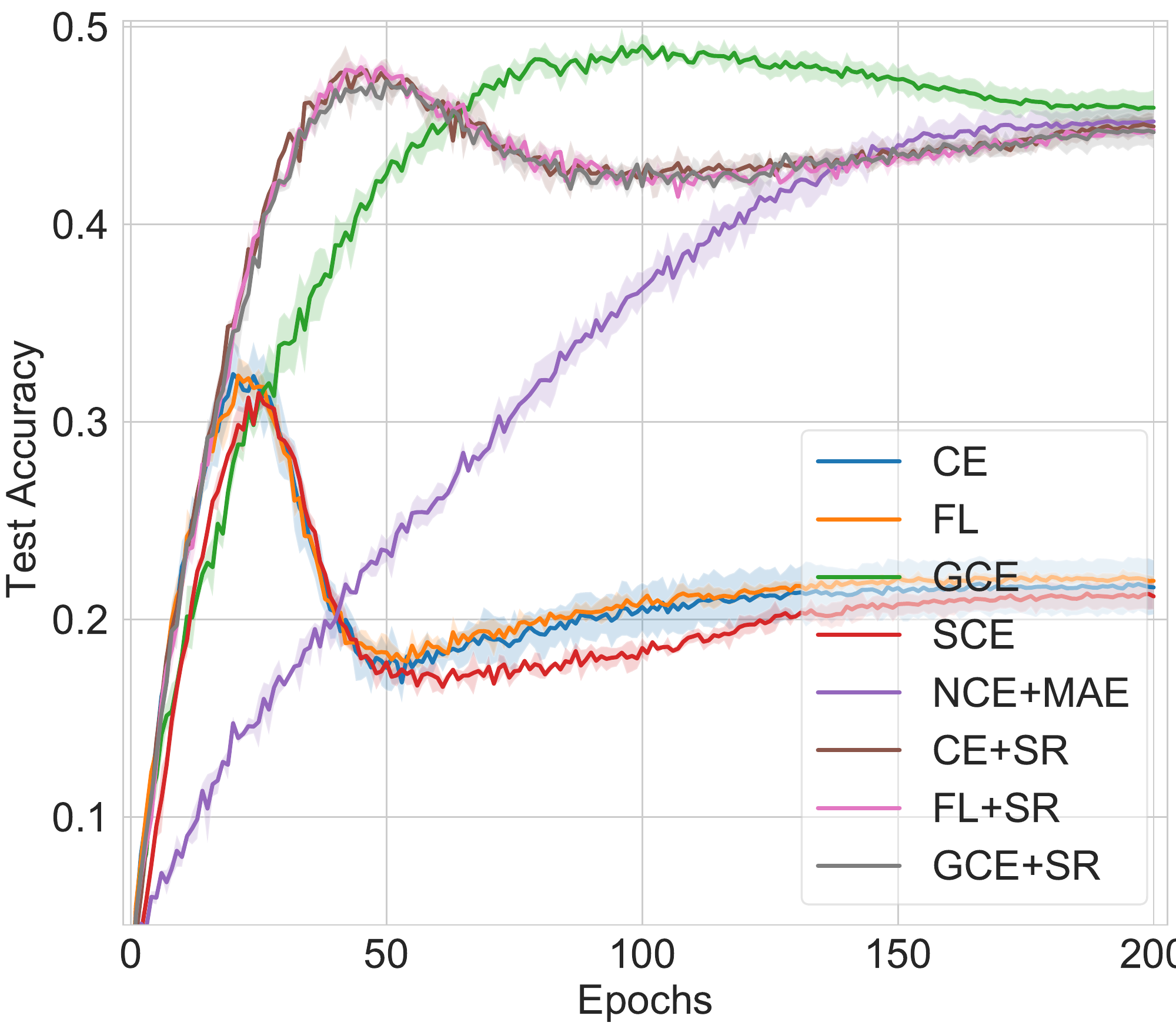}
        }
        \subfigure[$\eta=0.8$]{
        \includegraphics[width=1.2in]{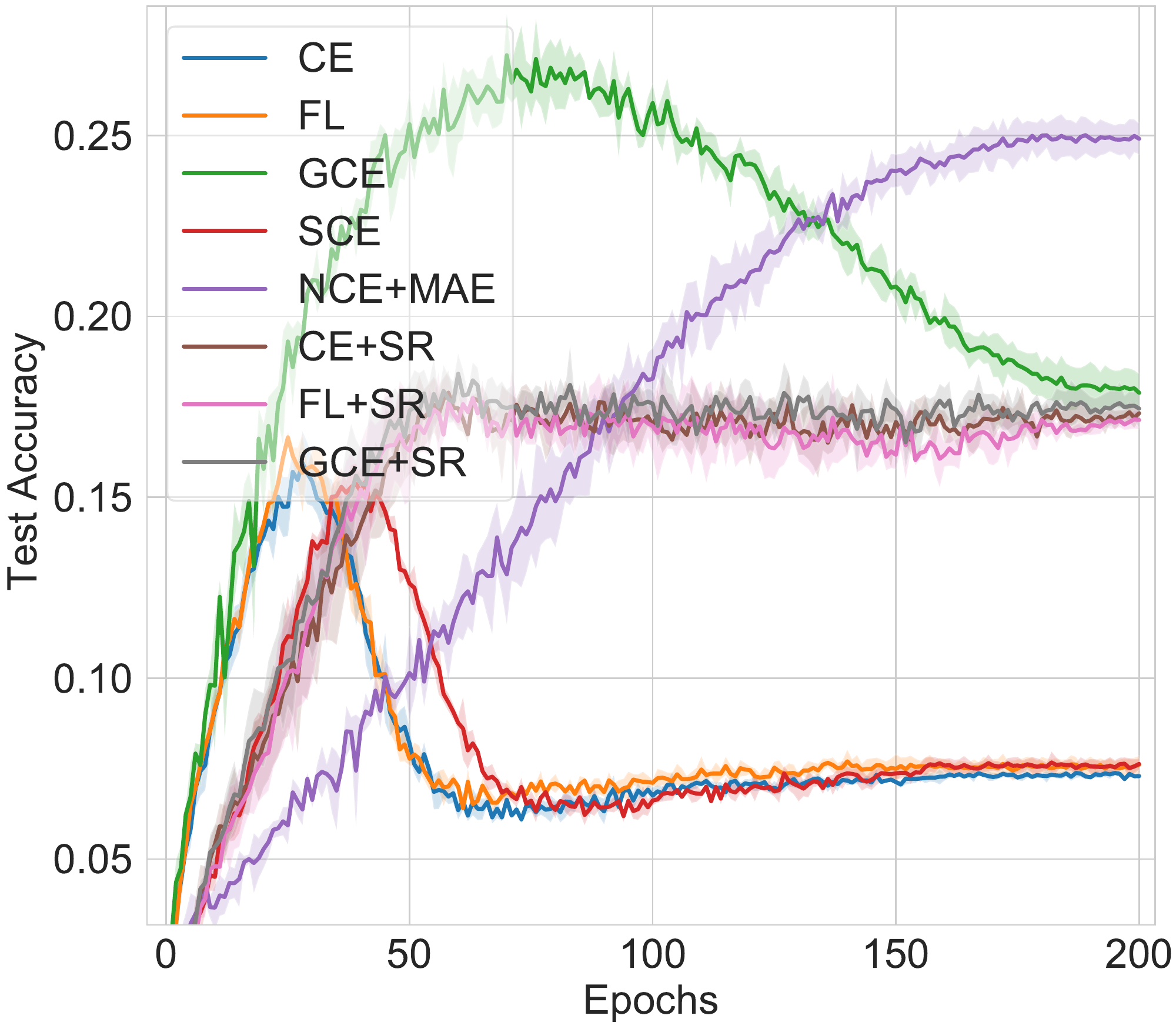}
        }
        \\
        \subfigure[$\eta=0.1$]{
        \includegraphics[width=1.2in]{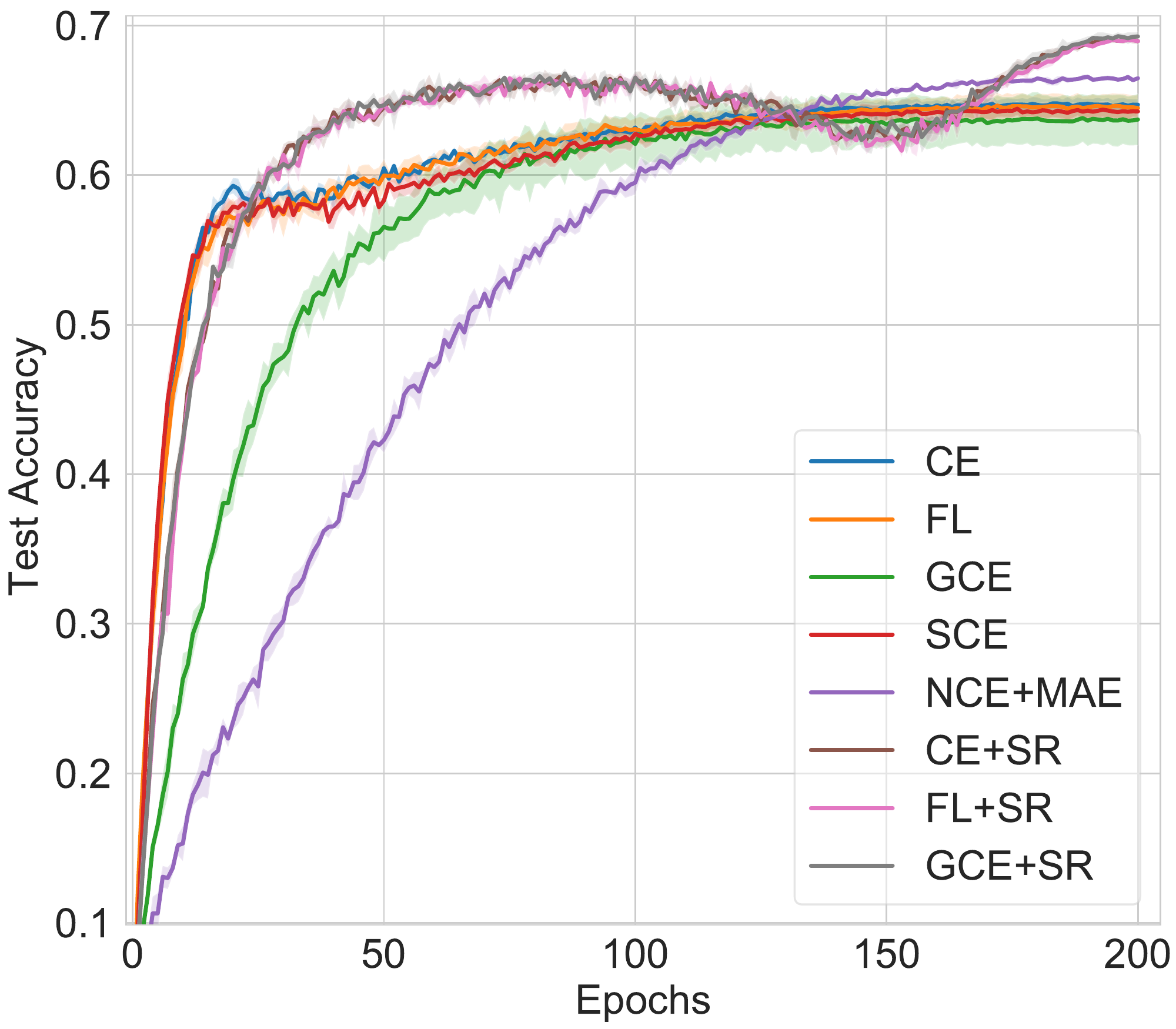}
        }
        \subfigure[$\eta=0.2$]{
        \includegraphics[width=1.2in]{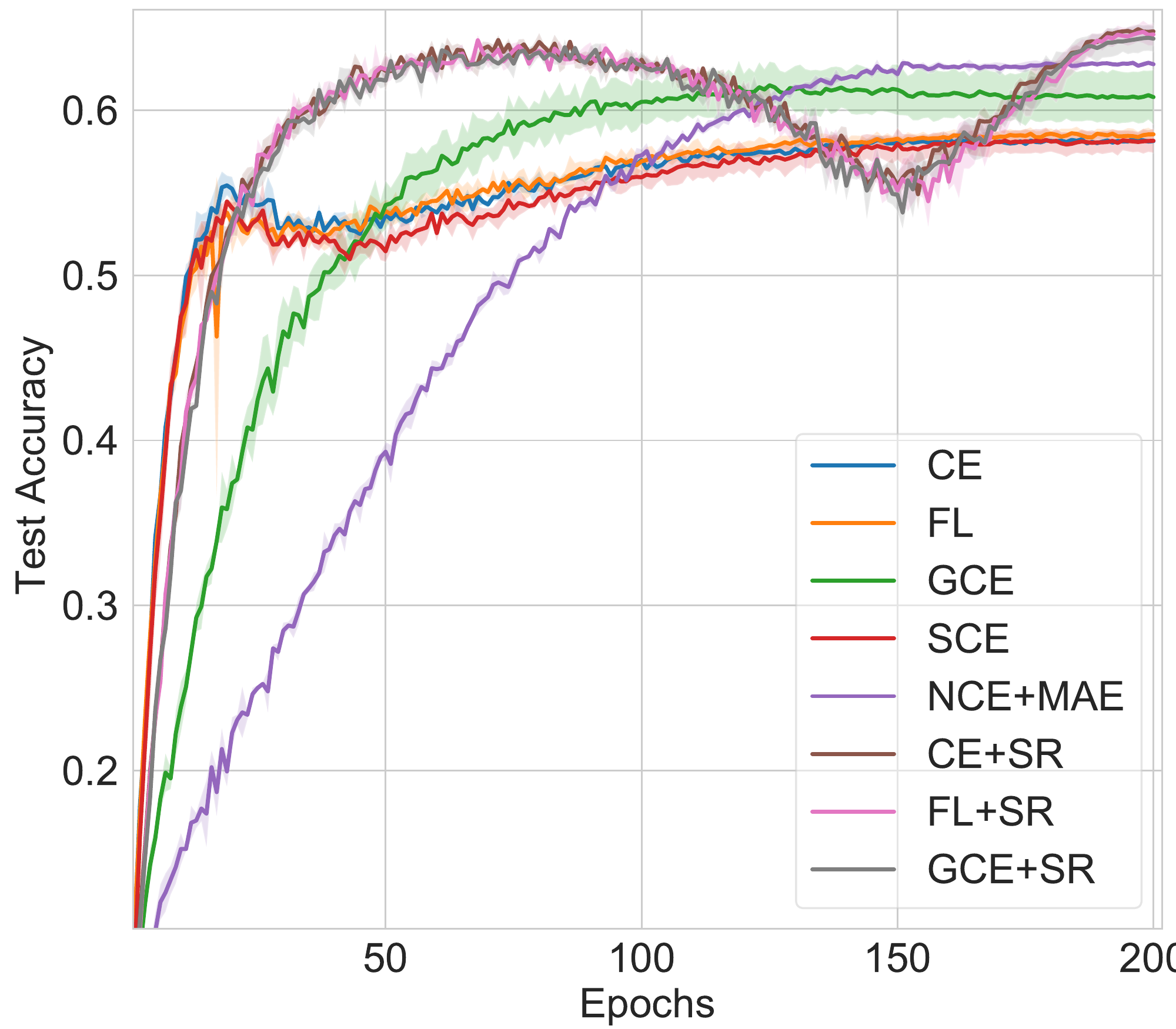}
        }
        \subfigure[$\eta=0.3$]{
        \includegraphics[width=1.2in]{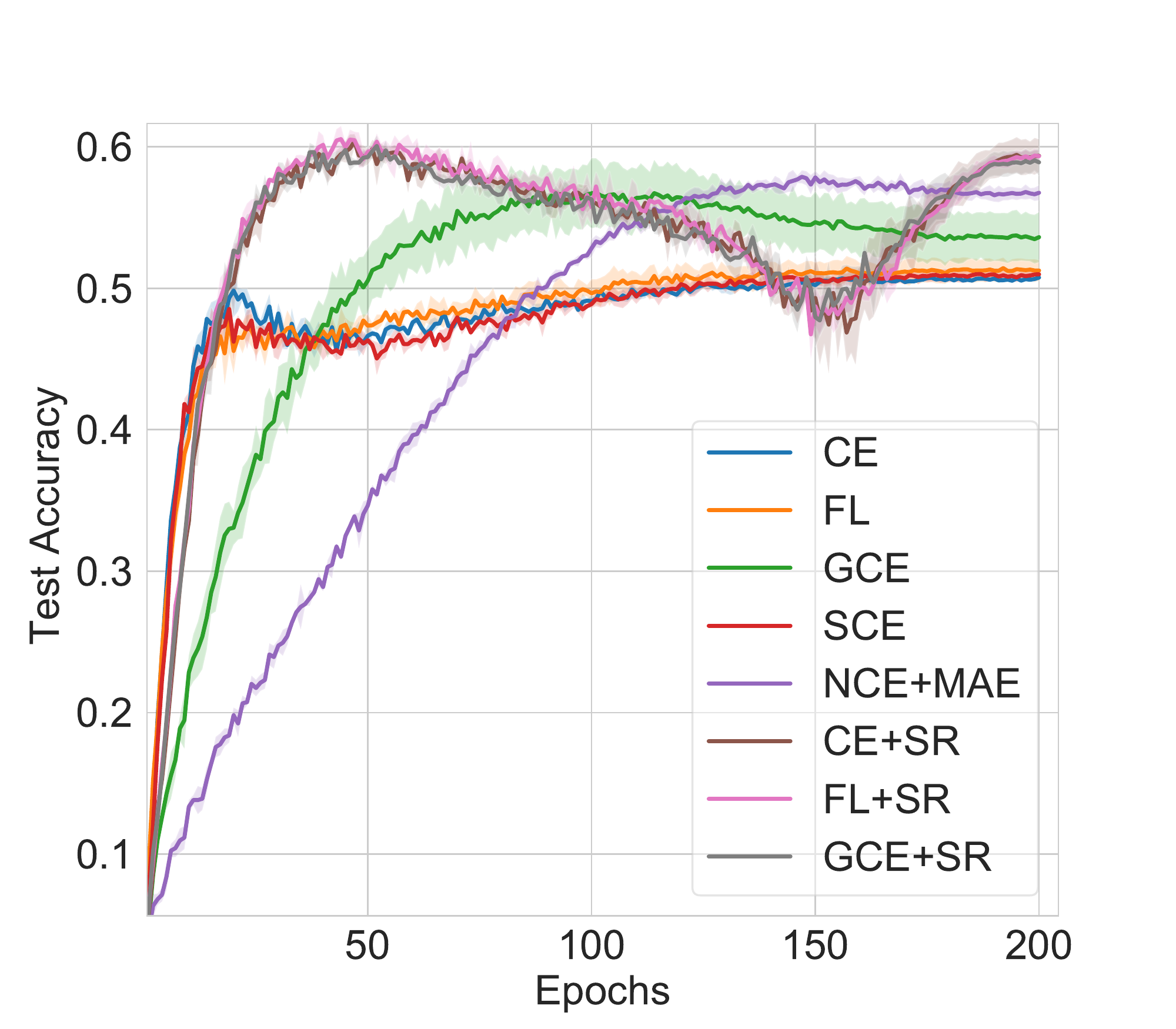}
        }
        \subfigure[$\eta=0.4$]{
        \includegraphics[width=1.2in]{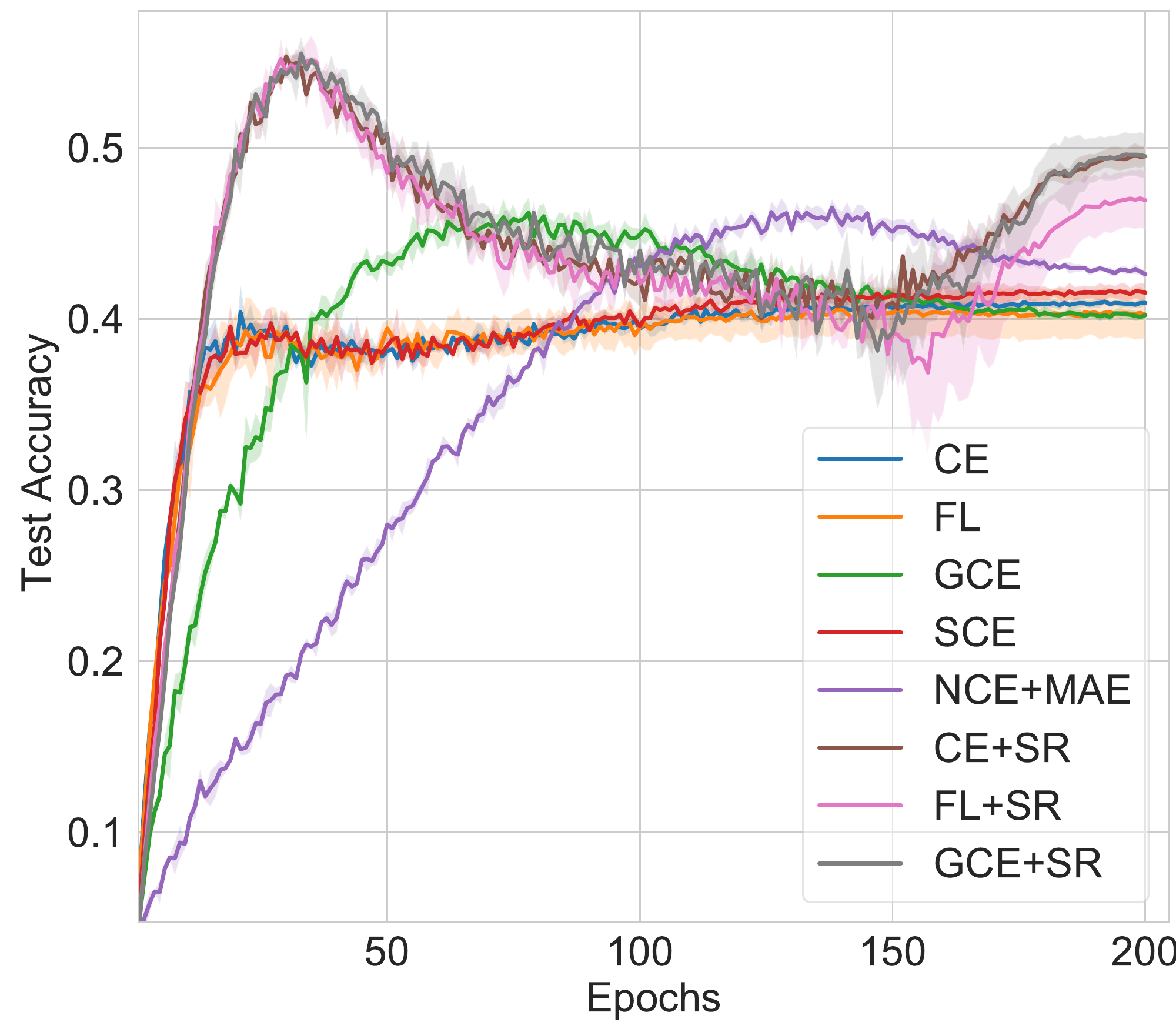}
        }
    \end{minipage}
    \caption{Test accuracies of different methods on CIFAR-100 with different label noise, where (a) denotes the clean case, (b-e) denote the symmetric label noise, and (f-i) denote the asymmetric label noise.}
    \label{cifar100-acc}
\end{figure*}

\noindent \textbf{More results of Comparison study.}\quad Fig. \ref{mnist-acc} shows test accuracy vs. epochs on MNIST. As can be observed, the commonly-used loss functions CE and FL suffer from significant overfitting in all noisy cases. The state-of-the-art methods GCE, SCE and APL show non-trival effectiveness of mitigating label noise, but the effects are crippled when meeting hard label noise. On the contrary, our proposed SR-enhanced methods CE+SR, FL+SR and GCE+SR perform better robustness and more efficiency . Fig. \ref{cifar10-acc} shows test accuracy vs. epochs on CIFAR-10. The results are similar to MNIST, our SR-enhanced methods keep robust and achieve the best accuracy in most cases. Fig. \ref{cifar100-acc} shows test accuracy vs. epochs on CIFAR-100. Our methods are of better fitting ability than commonly-used losses in the clean case, while the state-of-the-art GCE, SCE and APL encounter little underfitting. For 0.2 and 0.4 symmetric label noise, our methods perform the best test accuracy. Interestingly, for all asymmetric label noise, our methods perform overfitting at the beginning, but they later mitigate label noise and outperform other methods.

\begin{table}
\small
\setlength{\tabcolsep}{0.8mm}
\centering
\caption{Test accuracies (\%) of different methods on benchmark datasets with clean or asymmetric label noise ($\eta\in[0.1, 0.2, 0.3, 0.4]$). The results (mean$\pm$std) are reported over 3 random runs and the top 3 best results are \textbf{boldfaced}.}
\label{asymmetric-full-noise}
\begin{tabular}{c|c|cccc}
    \toprule
     \multirow{2}*{Datasets} & \multirow{2}*{Methods} & \multicolumn{4}{c}{Asymmetric Noise Rate ($\eta$)}  \\
     ~ & ~ & 0.1 & 0.2 & 0.3 & 0.4\\
     \hline\hline
     \multirow{9}*{MNIST} & CE & 97.57 $\pm$ 0.22 & 94.56 $\pm$ 0.22 & 88.81 $\pm$ 0.10 & 82.27 $\pm$ 0.40\\
    ~ & FL &97.58 $\pm$ 0.09 &94.25 $\pm$ 0.15 &89.09 $\pm$ 0.25 &82.13 $\pm$ 0.49\\
    ~ & GCE &99.01 $\pm$ 0.04 &96.69 $\pm$ 0.12 &89.12 $\pm$ 0.24 &81.51 $\pm$ 0.19\\
    % ~ & MAE &99.06 $\pm$ 0.05 &98.92 $\pm$ 0.08 &97.01 $\pm$ 0.06 &90.27 $\pm$ 0.44\\
    ~ & SCE &{99.14 $\pm$ 0.04} &98.03 $\pm$ 0.05 &93.68 $\pm$ 0.43 &85.36 $\pm$ 0.17\\
    ~ & NLNL &98.63 $\pm$ 0.06 &98.35 $\pm$ 0.01 &97.51 $\pm$ 0.15 &95.84 $\pm$ 0.26\\
    ~ & APL & \textbf{99.32 $\pm$ 0.09} &98.89 $\pm$ 0.04 &96.93 $\pm$ 0.17 &91.45 $\pm$ 0.40\\
    \cline{2-6}
    ~ & \textbf{CE+SR} & \textbf{99.42 $\pm$ 0.02} & \textbf{99.27 $\pm$ 0.06} & \textbf{99.24 $\pm$ 0.08} & \textbf{99.23 $\pm$ 0.07}\\
    ~ & \textbf{FL+SR} & \textbf{99.34 $\pm$ 0.05} & \textbf{99.31 $\pm$ 0.02} & \textbf{99.23 $\pm$ 0.02} & \textbf{99.36 $\pm$ 0.05}\\
    ~ & \textbf{GCE+SR} & 99.28 $\pm$ 0.06 & \textbf{99.22 $\pm$ 0.02} & \textbf{99.13 $\pm$ 0.05} & \textbf{99.09 $\pm$ 0.02}\\
    \hline\hline
     \multirow{9}*{CIFAR-10} & CE & 87.55 $\pm$ 0.14 & 83.32 $\pm$ 0.12 & 79.32 $\pm$ 0.59 & 74.67 $\pm$ 0.38\\
     ~ & FL & 86.43 $\pm$ 0.30 & 83.37 $\pm$ 0.07 & 79.33 $\pm$ 0.08 & 74.28 $\pm$ 0.44\\
    %  ~ & MAE & 89.26 $\pm$ 0.15 & 80.58 $\pm$ 5.81 & 63.62 $\pm$ 0.44 & 55.97 $\pm$ 0.06\\
     ~ & GCE & 88.33 $\pm$ 0.05 & 85.93 $\pm$ 0.23 & 80.88 $\pm$ 0.38 & 74.29 $\pm$ 0.43\\
     ~ & SCE & \textbf{89.77 $\pm$ 0.11} & 86.20 $\pm$ 0.37 & 81.38 $\pm$ 0.35 & 75.16 $\pm$ 0.39\\
     ~ & NLNL & 88.54 $\pm$ 0.25 & 84.74 $\pm$ 0.08 & 81.26$\pm$ 0.43 & 76.97 $\pm$ 0.52\\
     ~ & APL & 88.31 $\pm$ 0.20 &86.50 $\pm$ 0.31 & 83.34 $\pm$ 0.39 & 77.14 $\pm$ 0.33\\
     \cline{2-6}
     ~ & \textbf{CE+SR} & \textbf{89.08 $\pm$ 0.08} & \textbf{87.70 $\pm$ 0.19} & \textbf{85.63 $\pm$ 0.07} & \textbf{79.29 $\pm$ 0.20}\\
     ~ & \textbf{FL+SR} & 88.68 $\pm$ 0.23	& \textbf{87.56 $\pm$ 0.29}	& \textbf{85.10 $\pm$ 0.23}	& \textbf{79.07 $\pm$ 0.50}\\
     ~ & \textbf{GCE+SR} & \textbf{89.20 $\pm$ 0.23}	& \textbf{87.55 $\pm$ 0.08}	& \textbf{84.69 $\pm$ 0.46}	& \textbf{79.01 $\pm$ 0.18}\\
     \hline\hline
     \multirow{9}*{CIFAR-100} & CE &64.85 $\pm$ 0.37	&58.11 $\pm$ 0.32	&50.68 $\pm$ 0.55	&40.17 $\pm$ 1.31\\
     ~ & FL &64.78 $\pm$ 0.50	&58.05 $\pm$ 0.42	&51.15 $\pm$ 0.84	&41.18 $\pm$ 0.68\\
    %  ~ & MAE &3.53 $\pm$ 1.62	&4.40 $\pm$ 0.21	&3.33 $\pm$ 1.23	&2.73 $\pm$ 0.12\\
     ~ & GCE &63.01 $\pm$ 1.01	&59.35 $\pm$ 1.10	&53.83 $\pm$ 0.64	&40.91 $\pm$ 0.57\\
     ~ & NLNL & 59.55 $\pm$ 1.22 & 50.19 $\pm$ 0.56 & 42.81 $\pm$ 1.13 & 35.10 $\pm$ 0.20\\
     ~ & SCE &64.26 $\pm$ 0.43	&58.16 $\pm$ 0.73	& 50.98 $\pm$ 0.33	&41.54 $\pm$ 0.52\\
     ~ & APL & 66.48 $\pm$ 0.12	&62.80 $\pm$ 0.05	&56.74 $\pm$ 0.53	&42.61 $\pm$ 0.24\\
     \cline{2-6}
     ~ & \textbf{CE+SR} & \textbf{68.96 $\pm$ 0.22} & \textbf{64.79 $\pm$ 0.01} & \textbf{59.09 $\pm$ 2.10} & \textbf{49.51 $\pm$ 0.59}\\
     ~ & \textbf{FL+SR} & \textbf{68.96 $\pm$ 0.17} & \textbf{64.61 $\pm$ 0.67} & \textbf{58.94 $\pm$ 0.33} & \textbf{46.94 $\pm$ 1.68}\\
      ~ & \textbf{GCE+SR} & \textbf{69.27 $\pm$ 0.31} & \textbf{64.35 $\pm$ 0.78} &  \textbf{57.22 $\pm$ 0.80} & \textbf{49.51 $\pm$ 1.31}\\
     \bottomrule
\end{tabular}
\vskip-5pt
\end{table}

\begin{figure}
    \centering
    \subfigure[CE+SR ($\tau=1.0$)]{
        \label{lambda-cifar10-CE}
        \includegraphics[width=1.6in]{CE-l-t=1.pdf}
    }
    \subfigure[FL+SR ($\tau=1.0$)]{
        \label{lambda-cifar10-FL}
        \includegraphics[width=1.6in]{FL-l-t=1.pdf}
    }
    \subfigure[CE+SR ($\tau=0.5$)]{
        \label{lambda-cifar10--CE-0.5}
        \includegraphics[width=1.6in]{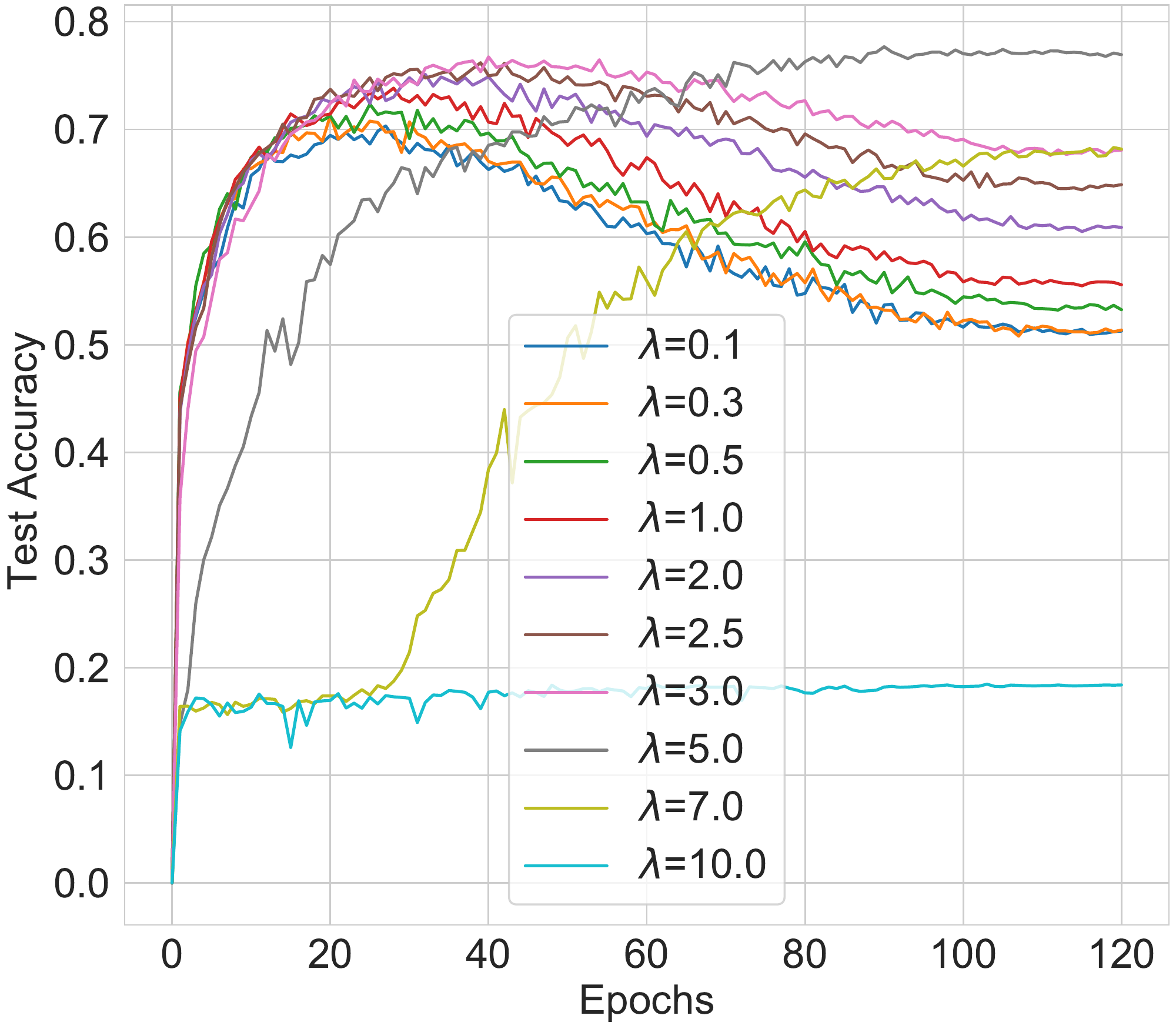}
    }
    \subfigure[FL+SR ($\tau=0.5$)]{
        \label{lambda-cifar10-FL-0.5}
        \includegraphics[width=1.6in]{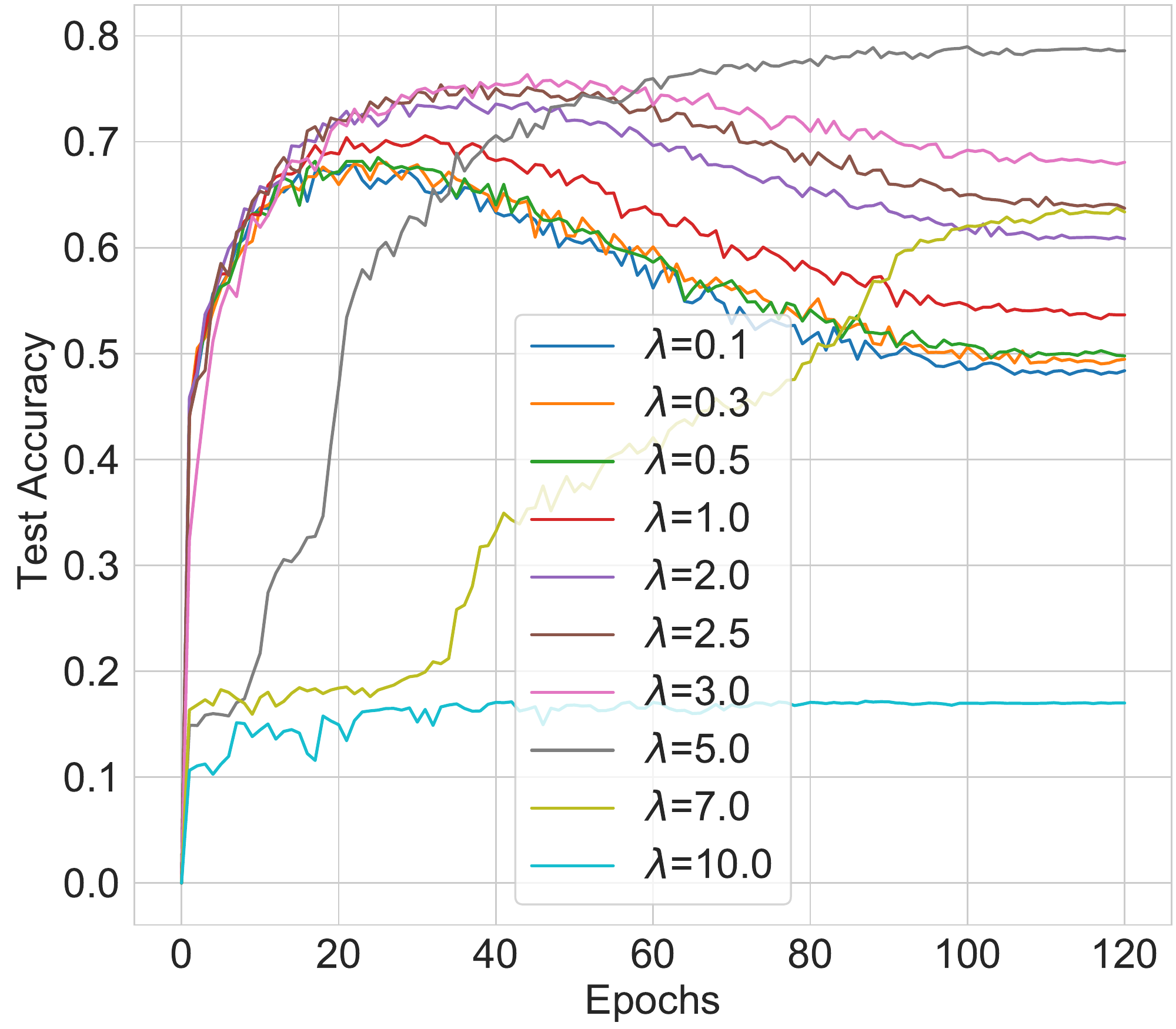}
    }
    \caption{Test accuracy curve of different $\lambda$ on CIFAR-10 with 0.6 symmetric label noise.}
    \label{lambda-cifar10}
    \vskip-10pt
\end{figure}

\begin{figure*}[htb]
    \centering
    \subfigure[CE+SR with $\eta=0.1$]{
    \label{fig:ce+sr0.0}
    \includegraphics[width=1.2in]{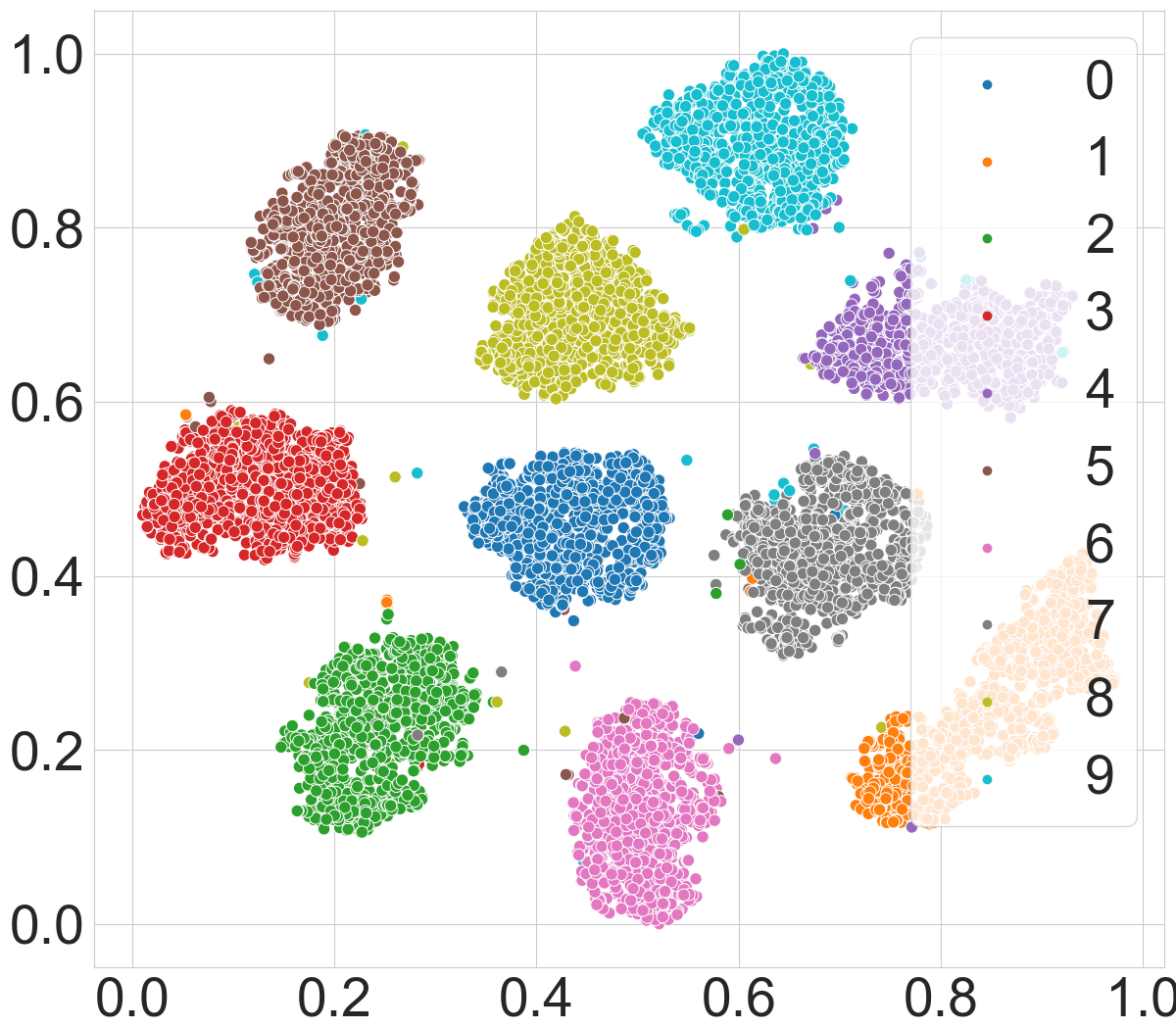}
    }
    \subfigure[CE+SR with $\eta=0.2$]{
    \label{fig:ce+sr0.2}
    \includegraphics[width=1.2in]{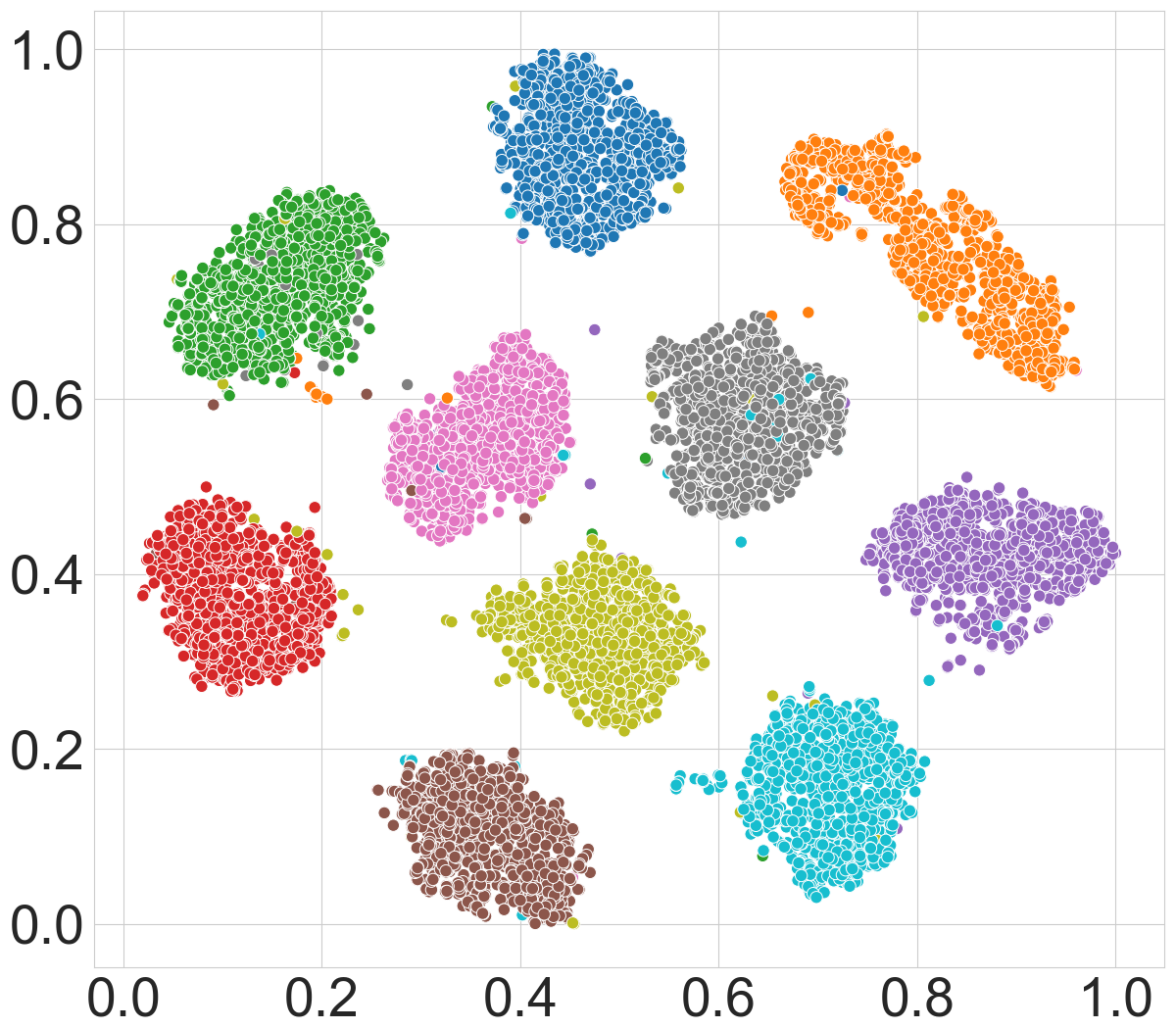}
    }
    \subfigure[CE+SR with $\eta=0.3$]{
    \label{fig:ce+sr0.4}
    \includegraphics[width=1.2in]{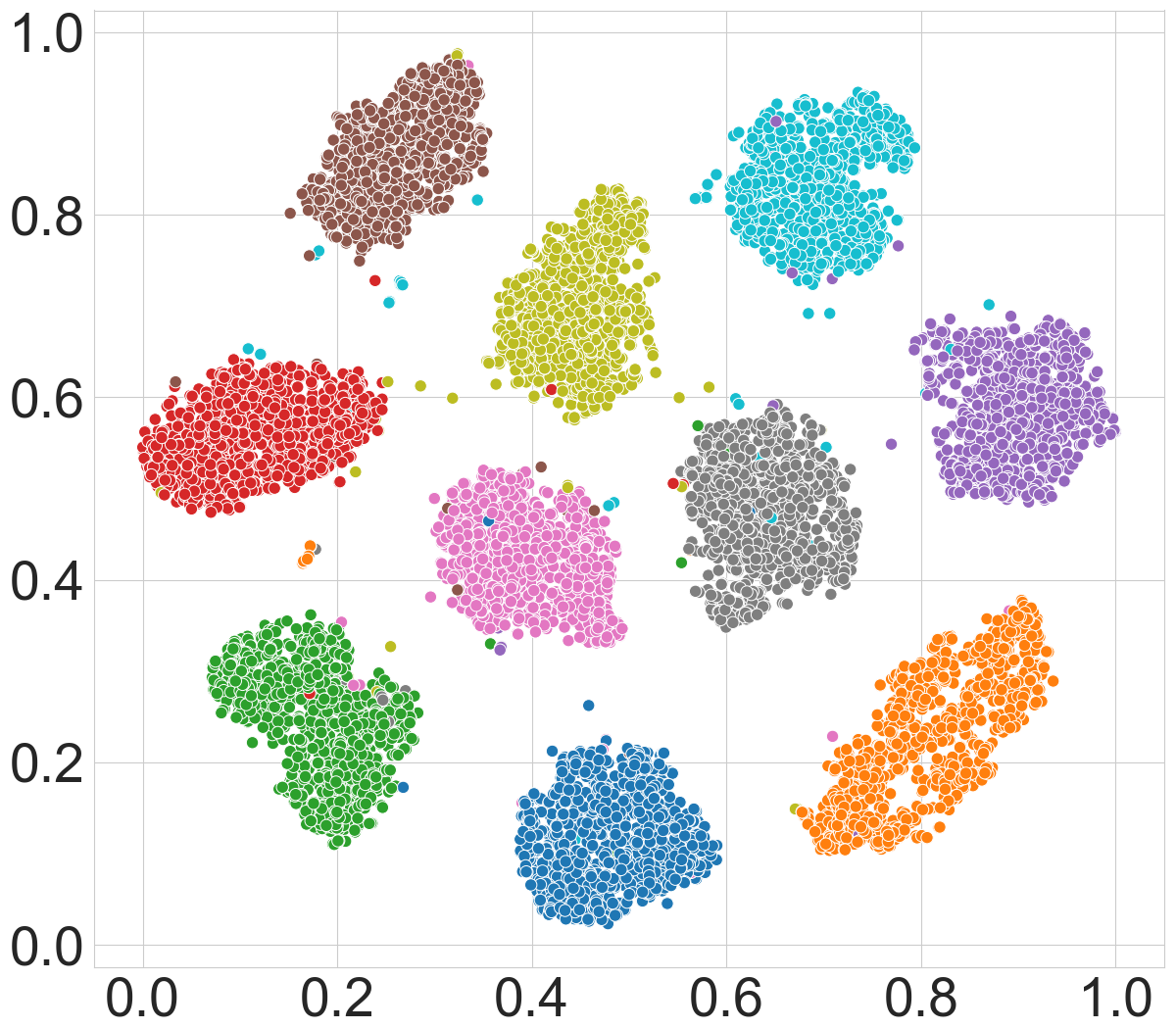}
    }
    \subfigure[CE+SR with $\eta=0.4$]{
    \label{fig:ce+sr0.6}
    \includegraphics[width=1.2in]{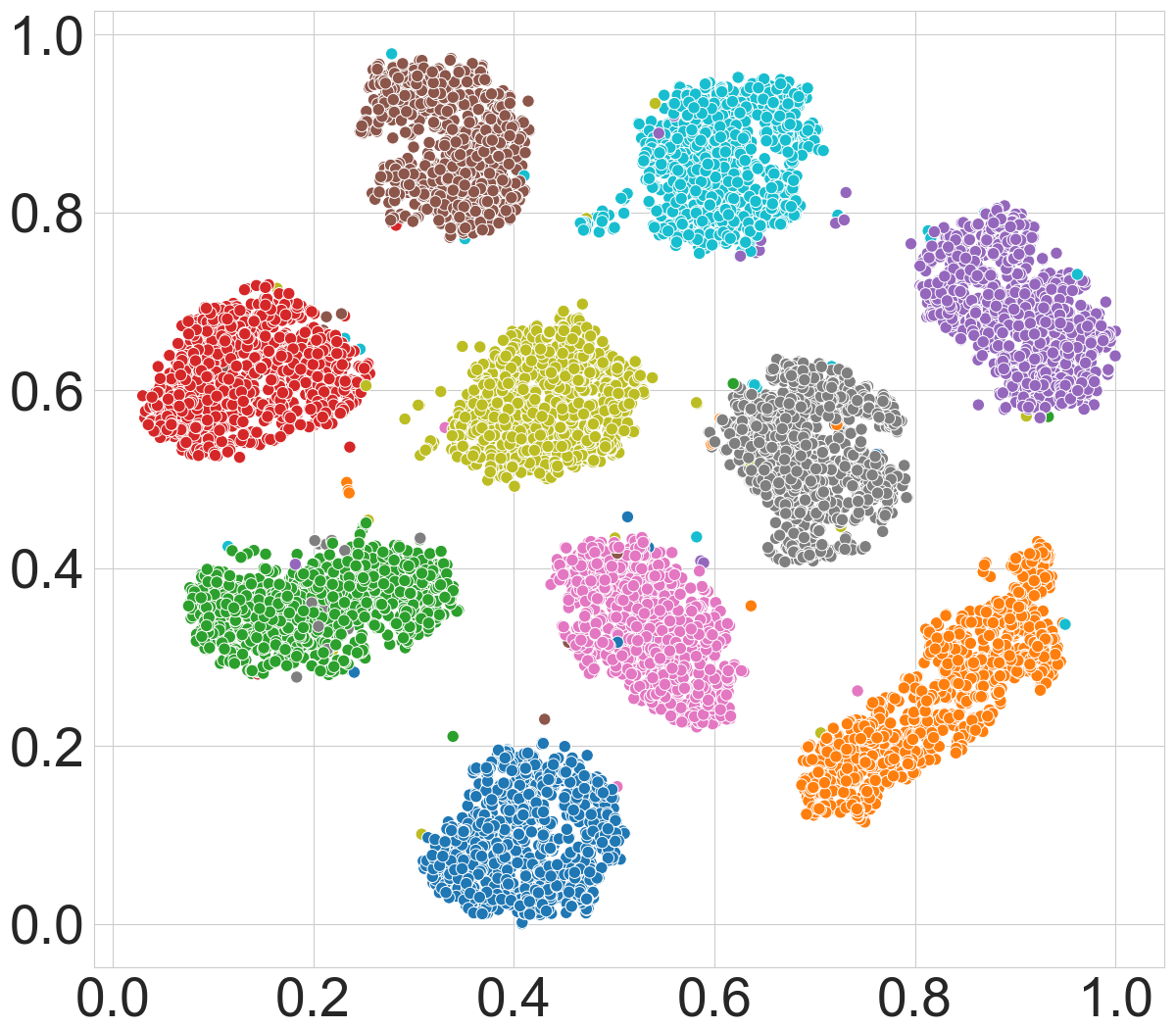}
    }\\
    \subfigure[FL+SR with $\eta=0.1$]{
    \label{fig:fl+sr0.0}
    \includegraphics[width=1.2in]{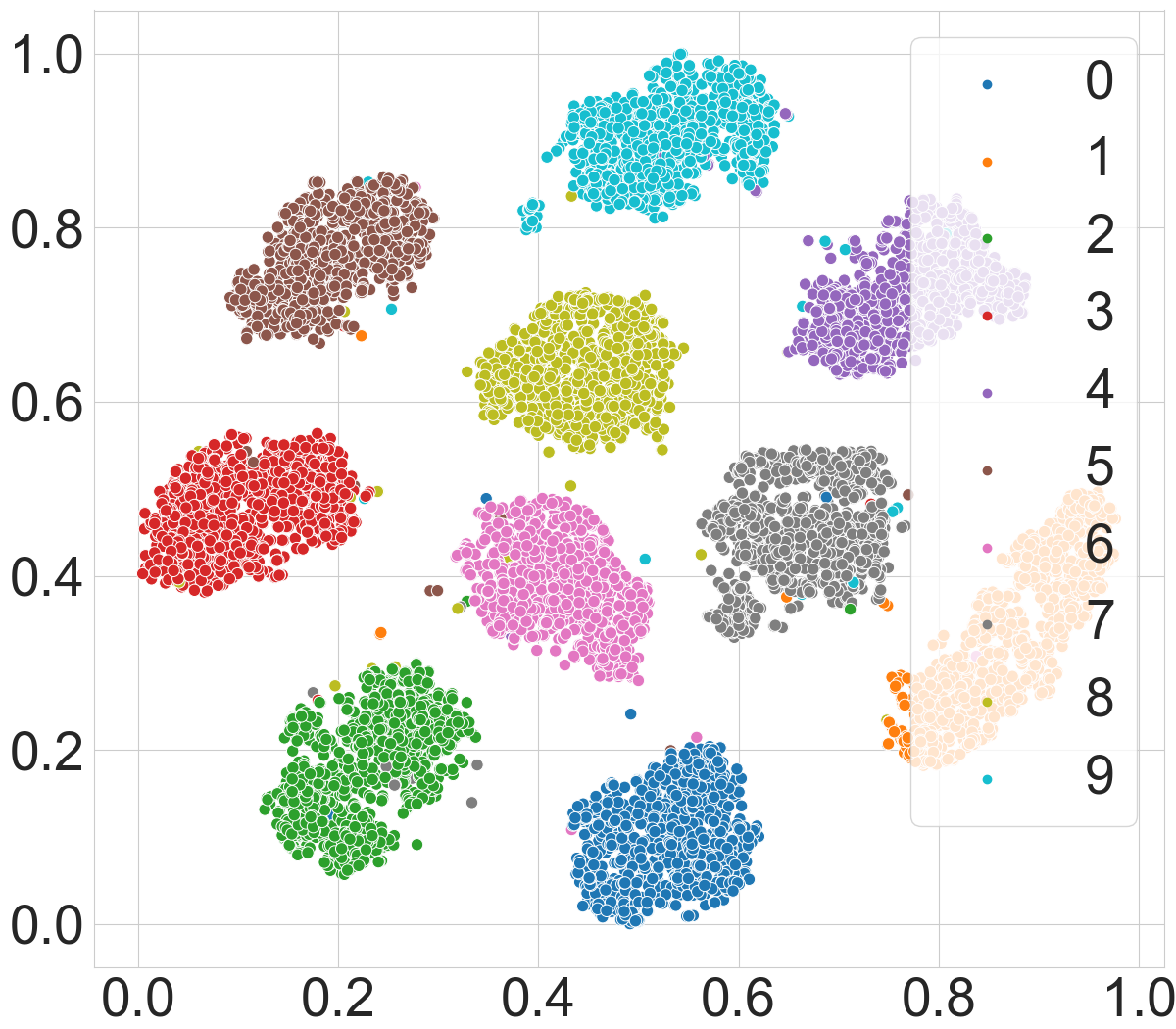}
    }
    \subfigure[FL+SR with $\eta=0.2$]{
    \label{fig:fl+sr0.2}
    \includegraphics[width=1.2in]{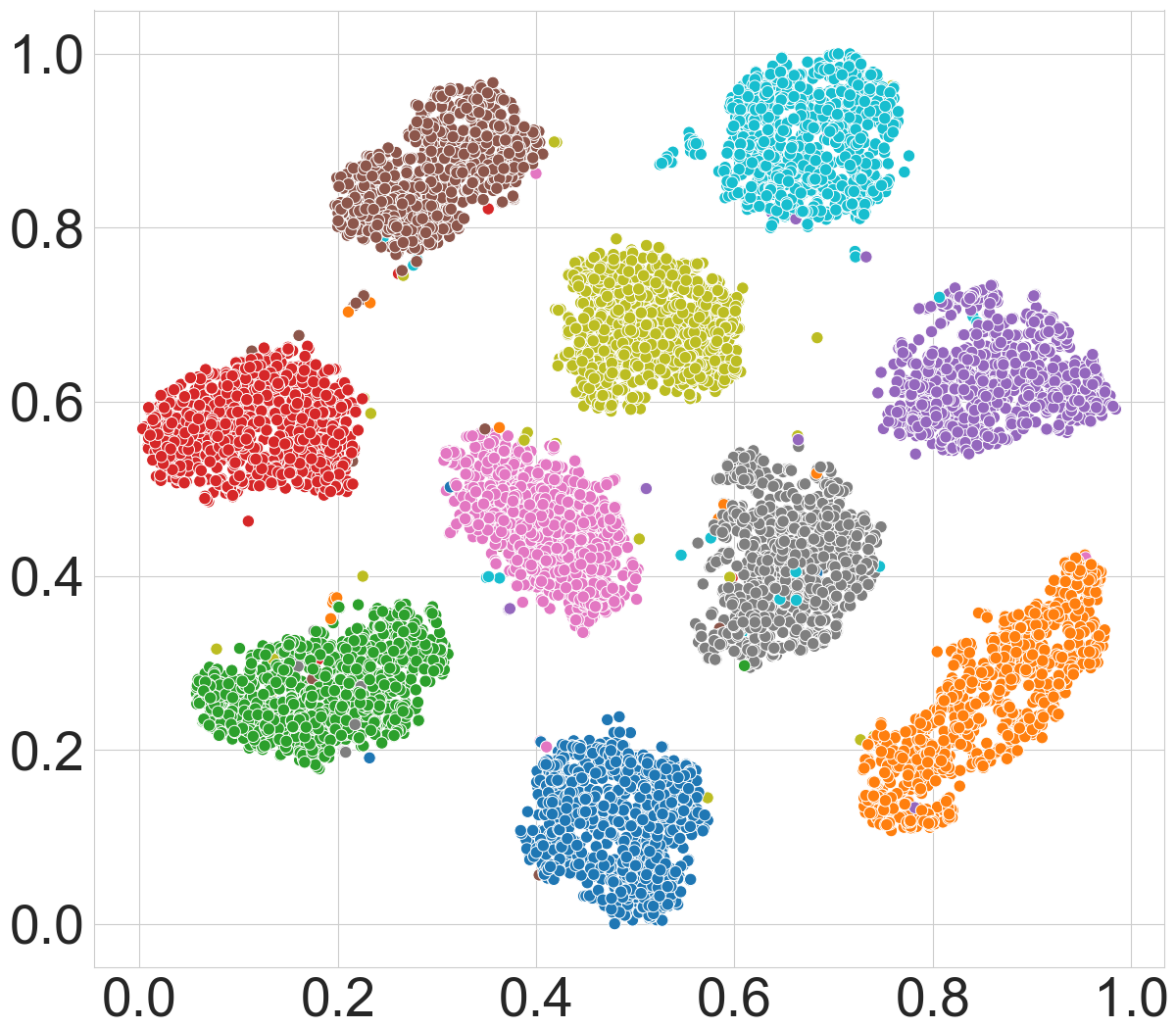}
    }
    \subfigure[FL+SR with $\eta=0.3$]{
    \label{fig:fl+sr0.4}
    \includegraphics[width=1.2in]{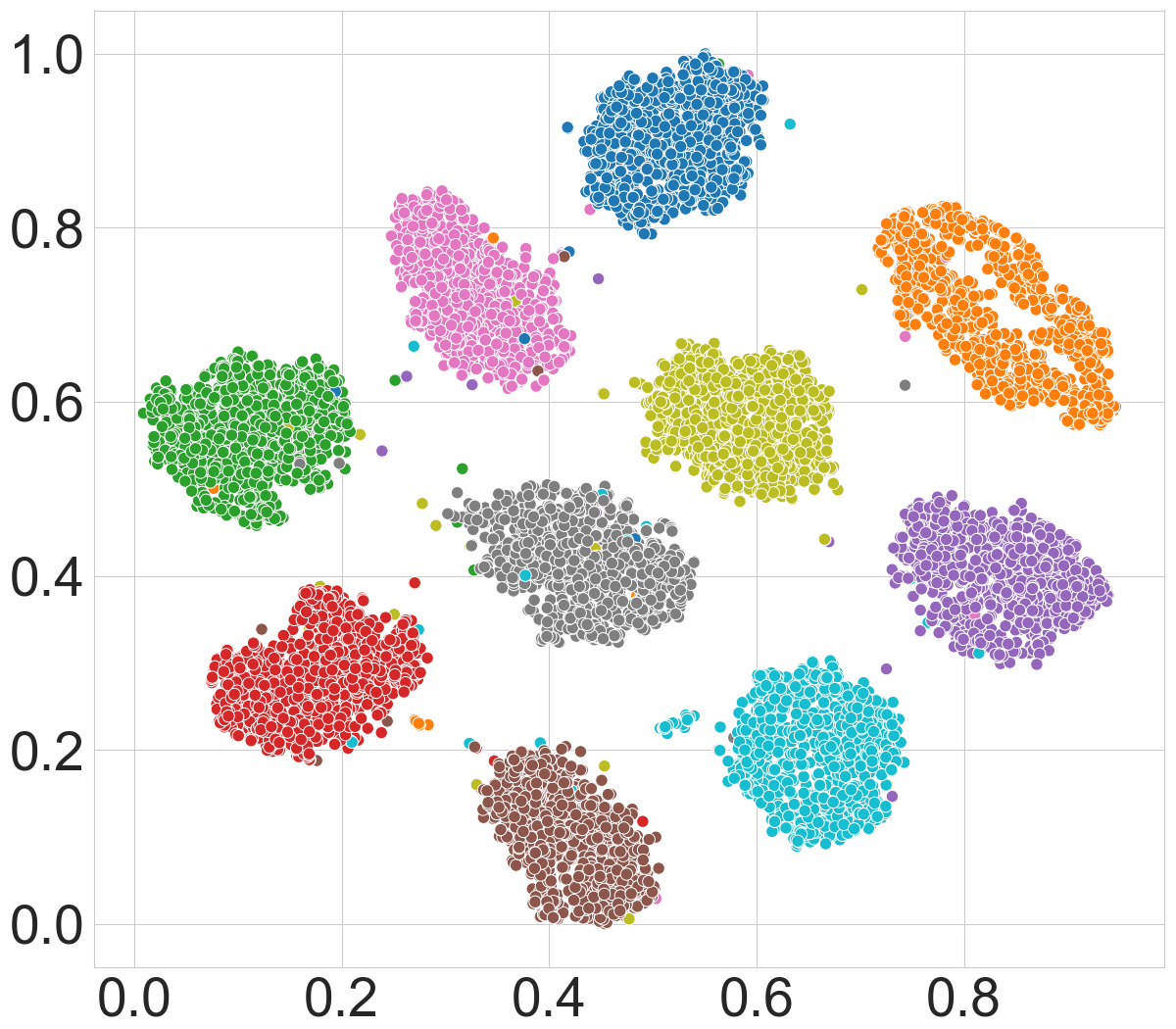}
    }
    \subfigure[FL+SR with $\eta=0.4$]{
    \label{fig:fl+sr0.6}
    \includegraphics[width=1.2in]{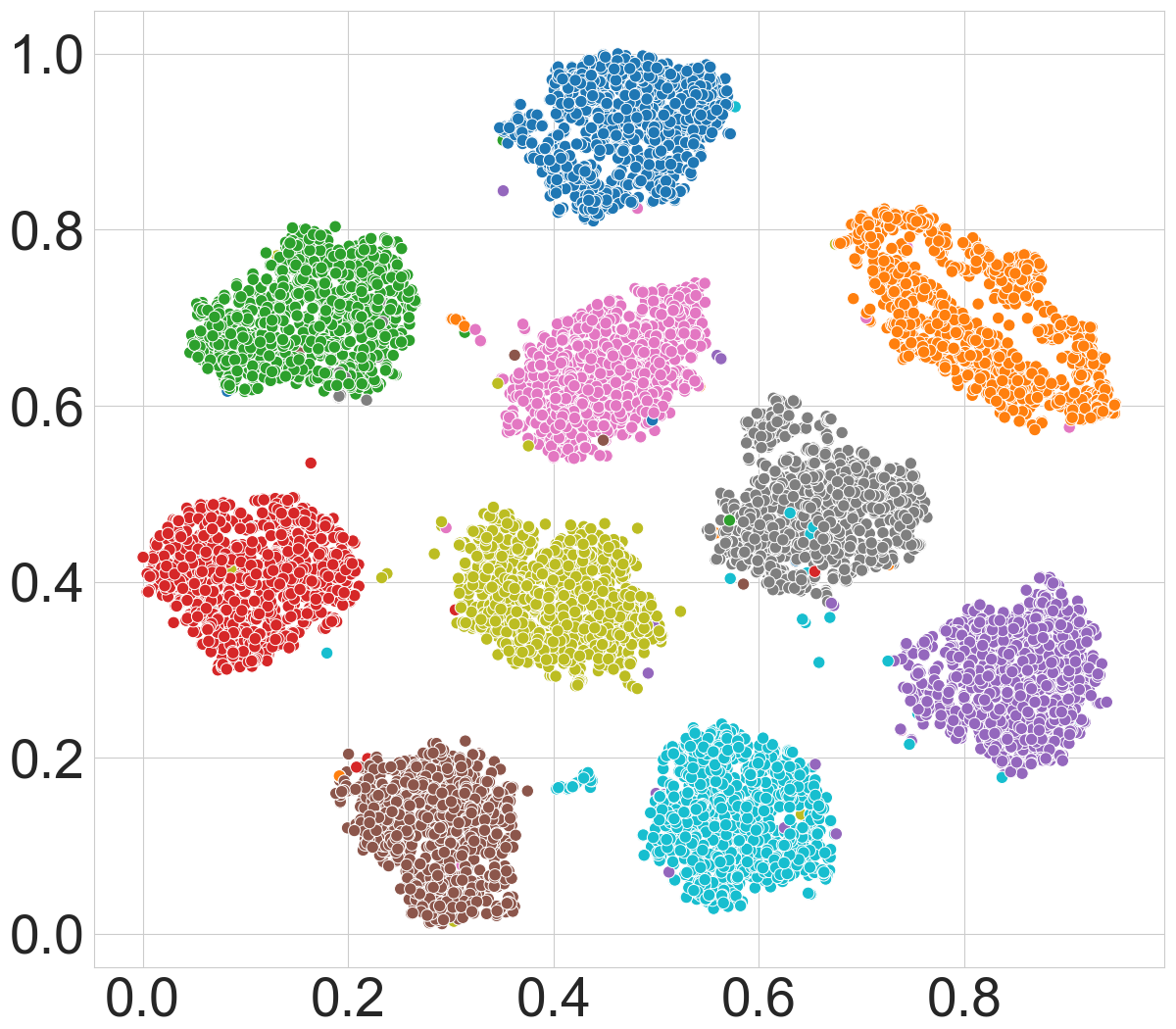}
    }
    \\
    \subfigure[GCE+SR with $\eta=0.1$]{
    \label{fig:gce+sr0.0}
    \includegraphics[width=1.2in]{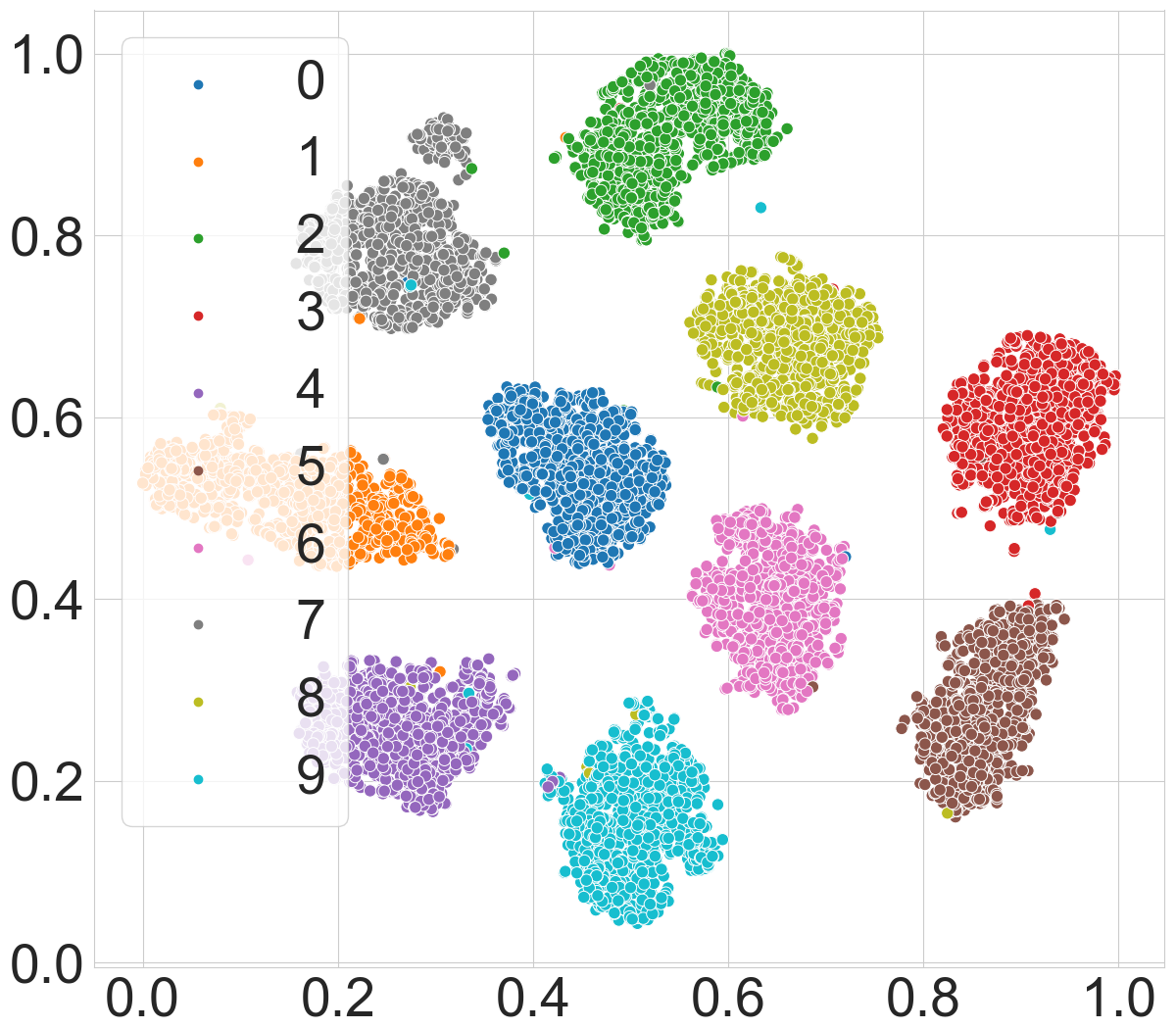}
    }
    \subfigure[GCE+SR with $\eta=0.2$]{
    \label{fig:gce+sr0.2}
    \includegraphics[width=1.2in]{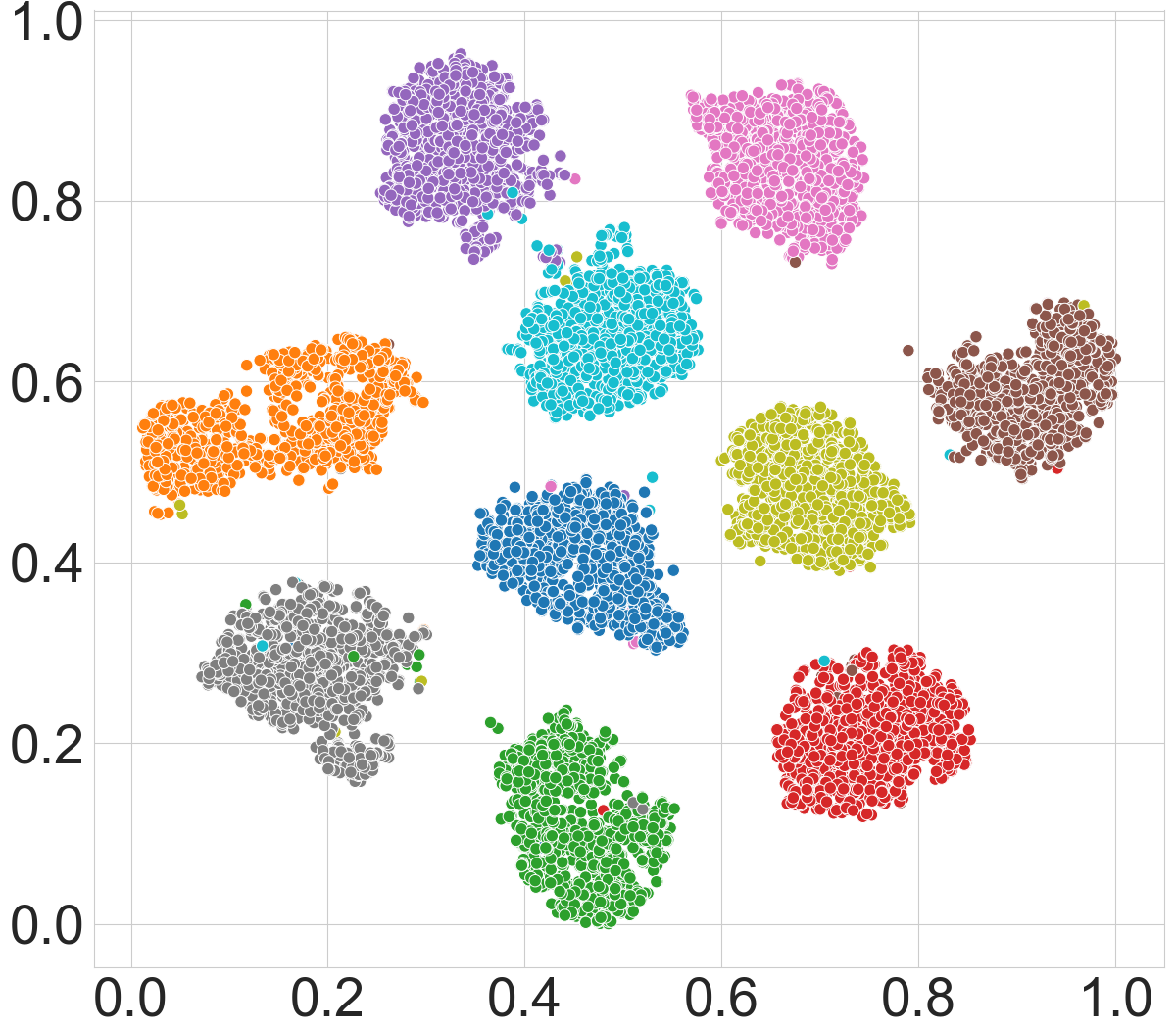}
    }
    \subfigure[GCE+SR with $\eta=0.3$]{
    \label{fig:gce+sr0.4}
    \includegraphics[width=1.2in]{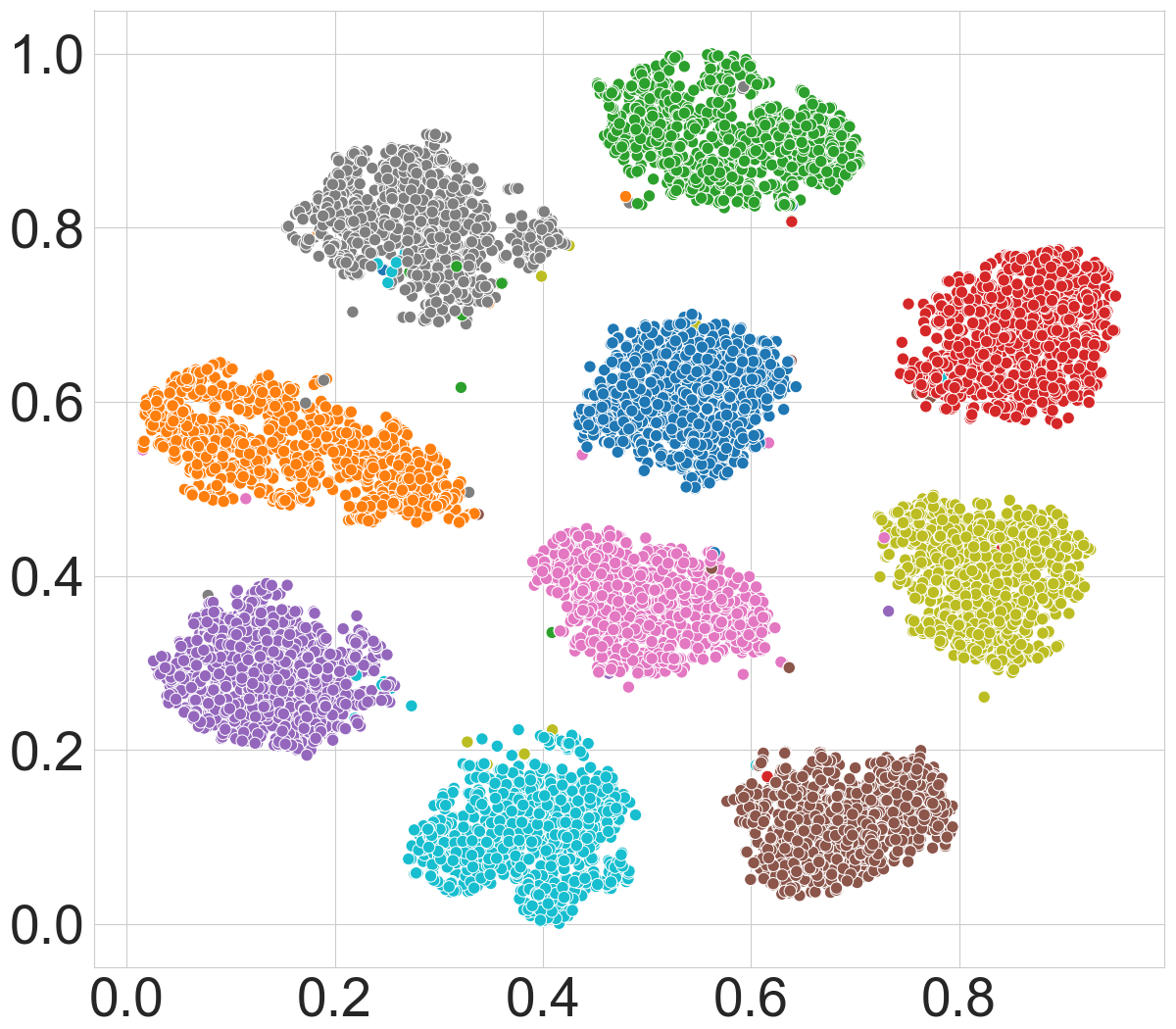}
    }
    \subfigure[GCE+SR with $\eta=0.4$]{
    \label{fig:gce+sr0.6}
    \includegraphics[width=1.2in]{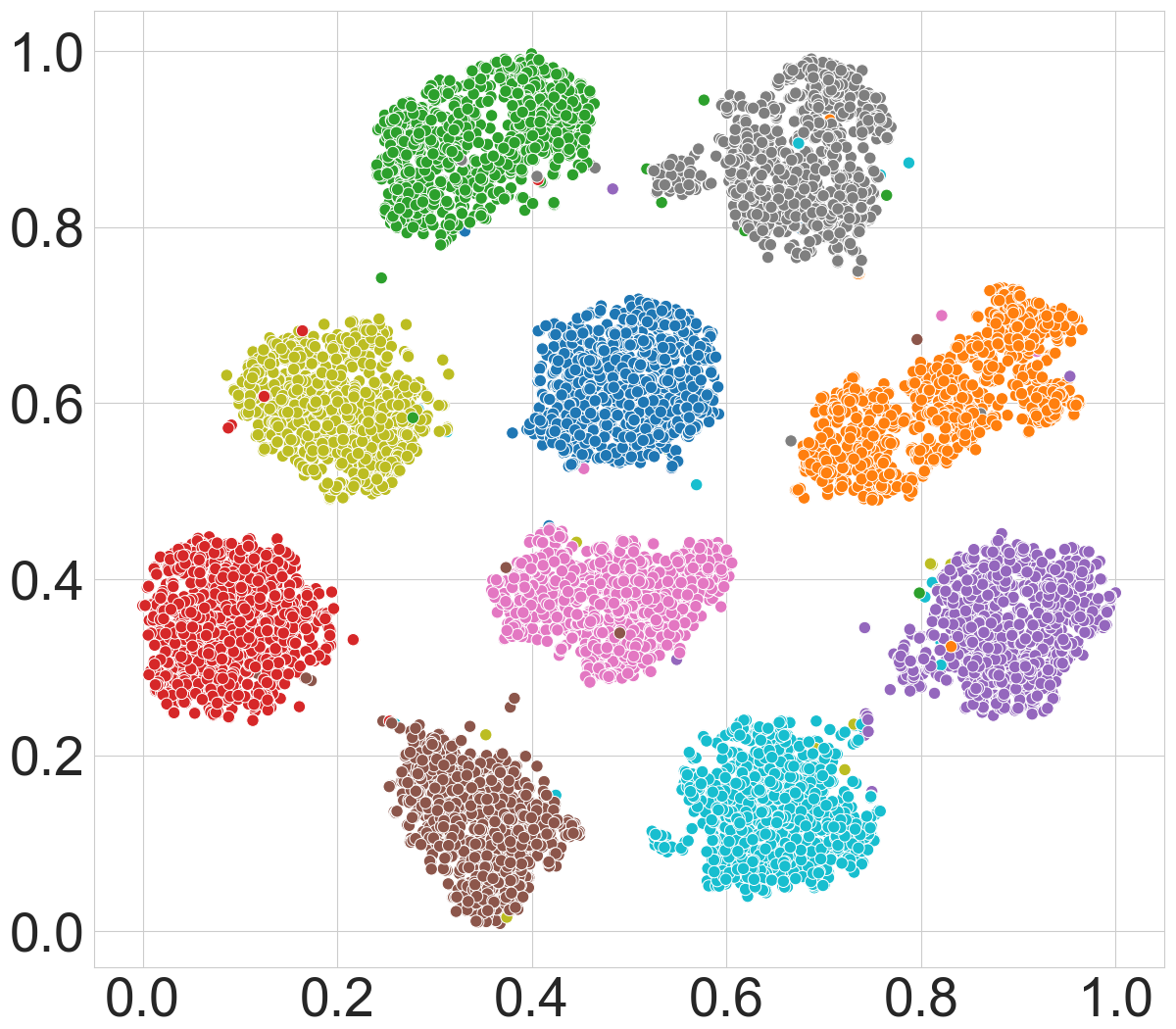}
    }
    
    \caption{Features visualization for CE+SR (top) and FL+SR (bottom) on MNIST with different asymmetric label noise ($\eta\in[0.1, 0.2, 0.3, 0.4]$) by t-SNE \cite{tsne} 2D embeddings at the last second full-connected layer.}
    \label{fig:mnist-a-tsne}
\end{figure*}

\begin{figure*}[htb]
    \centering
    \subfigure[CE with $\eta=0.2$]{
    \label{fig:CE-cifar10-0.2}
    \includegraphics[width=1.5in]{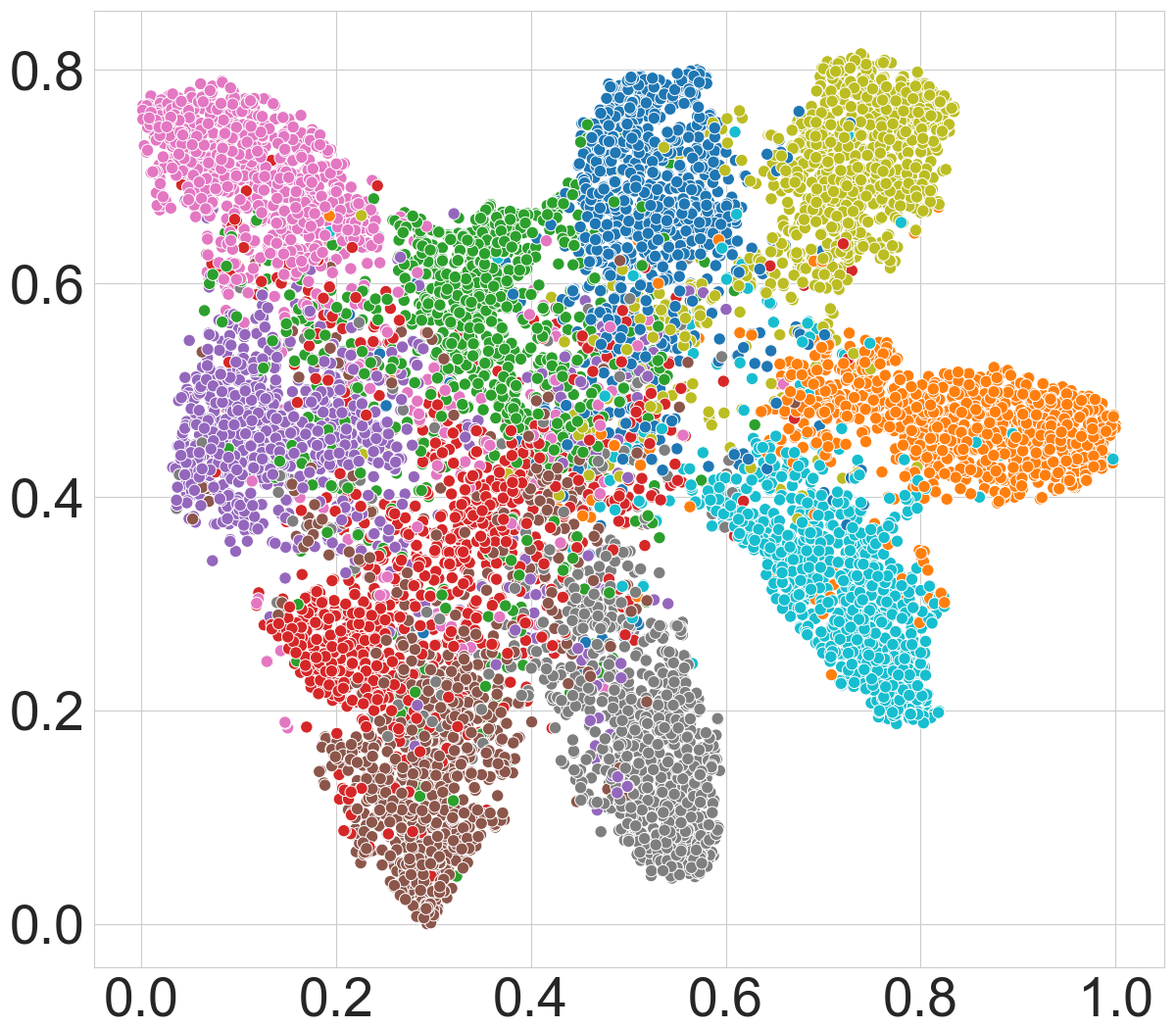}
    }
    \subfigure[FL with $\eta=0.2$]{
    \label{fig:FL-cifar10-0.2}
    \includegraphics[width=1.5in]{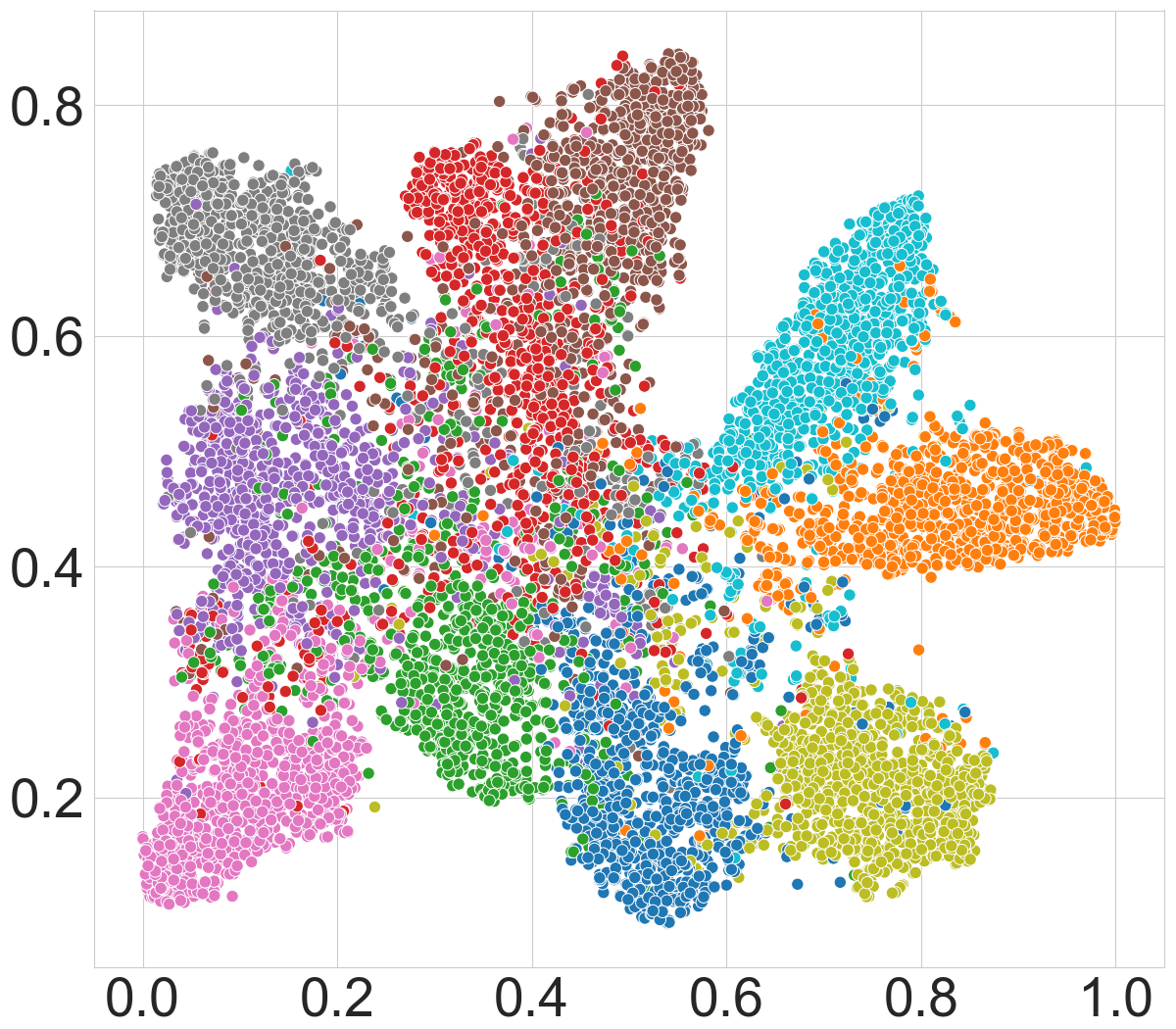}
    }
    \subfigure[GCE with $\eta=0.2$]{
    \label{fig:GCE-cifar10-0.2}
    \includegraphics[width=1.5in]{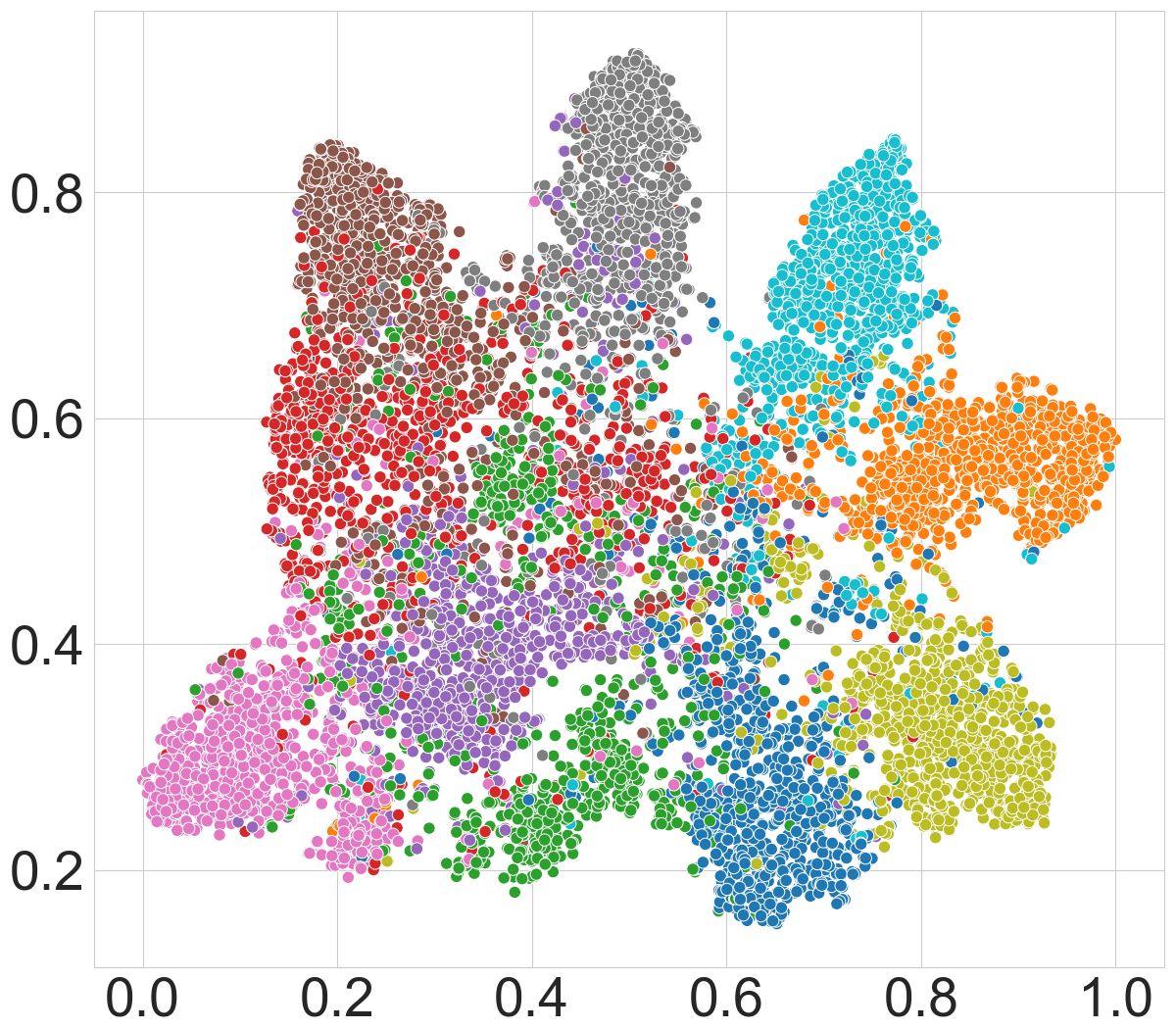}
    }
    \\
    \subfigure[CE+SR with $\eta=0.2$]{
    \label{fig:CESR-cifar10-0.2}
    \includegraphics[width=1.5in]{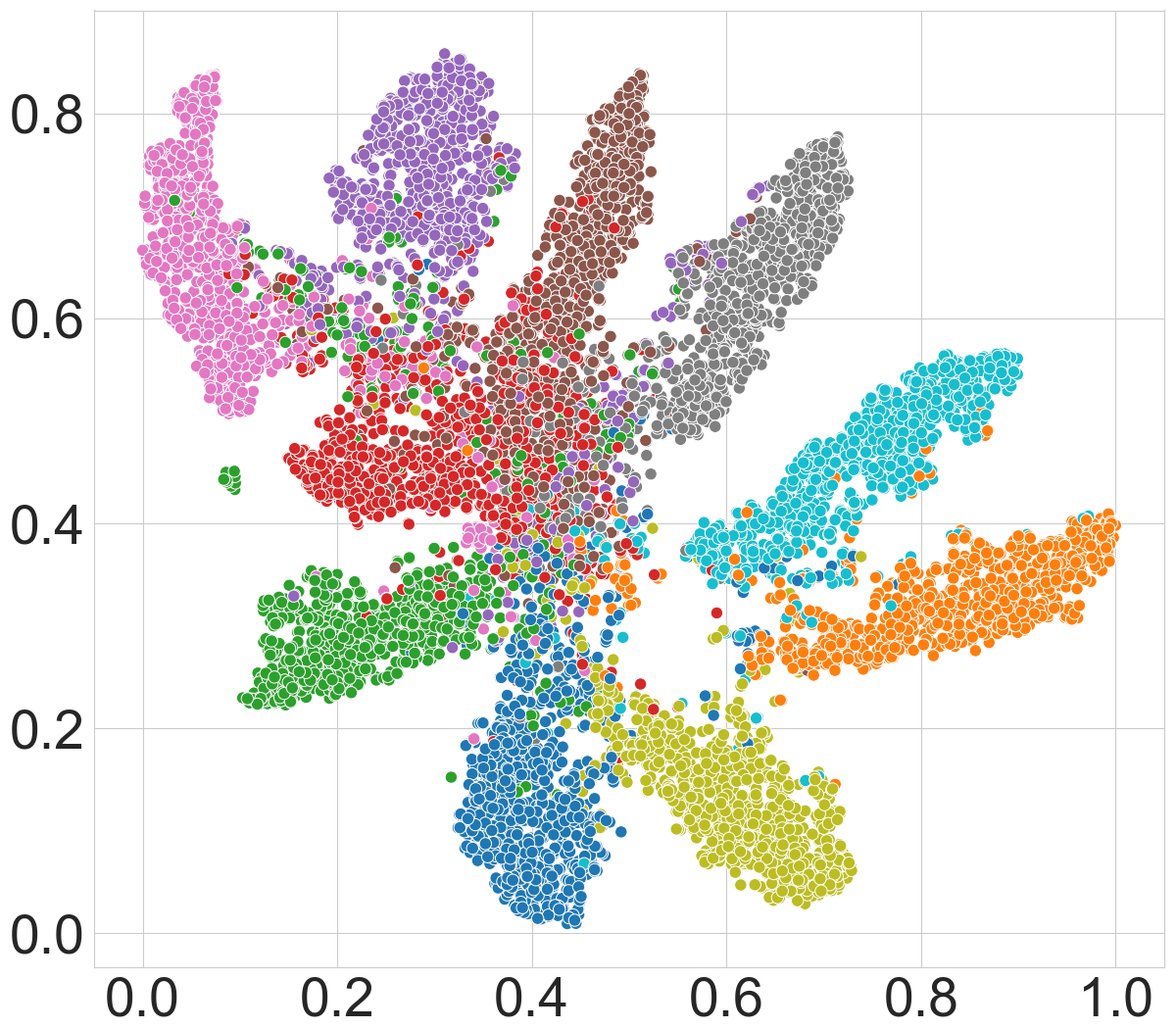}
    }
    \subfigure[FL+SR with $\eta=0.2$]{
    \label{fig:FLSR-cifar10-0.2}
    \includegraphics[width=1.5in]{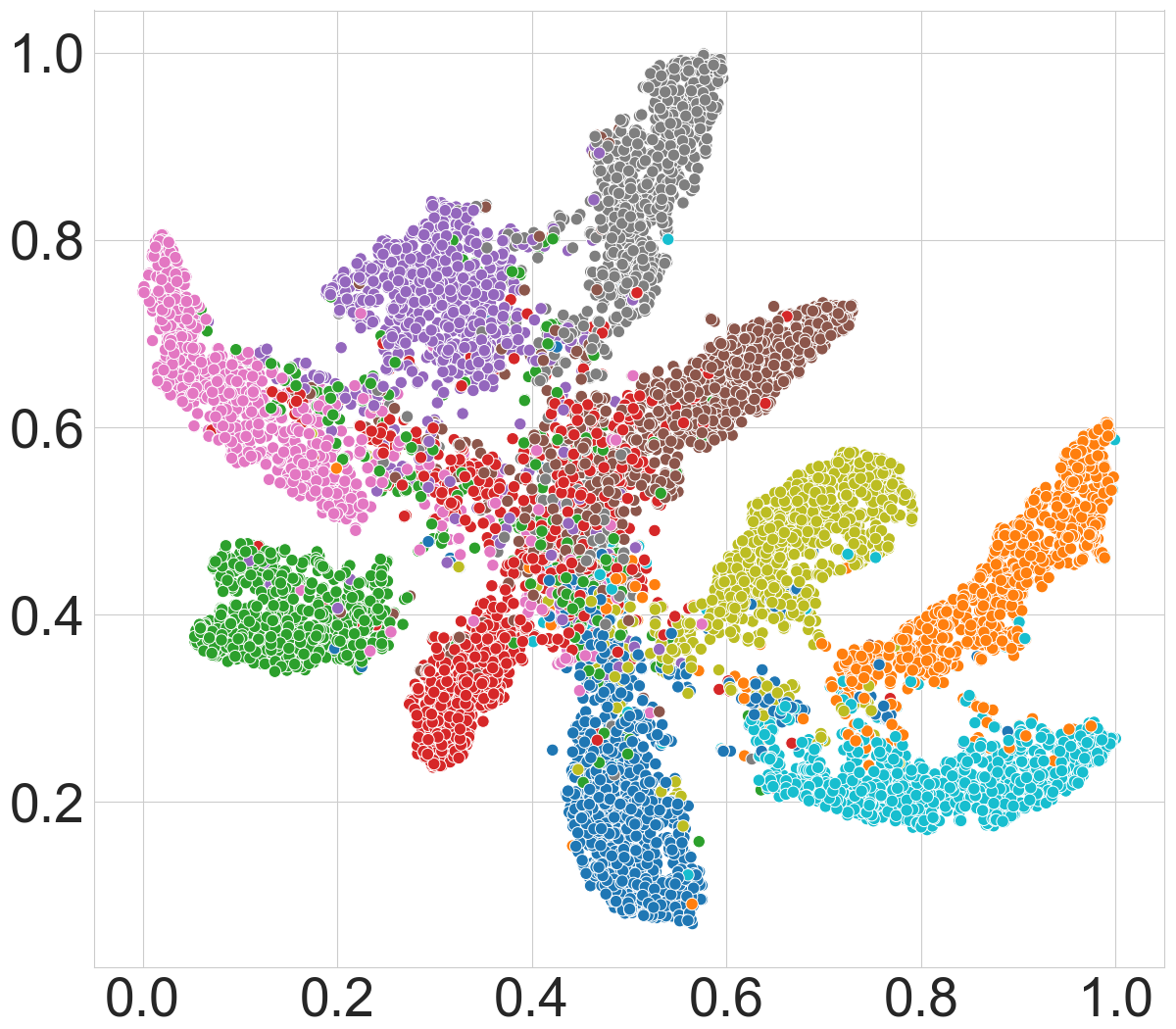}
    }
    \subfigure[GCE+SR with $\eta=0.2$]{
    \label{fig:GCESR-cifar10-0.2}
    \includegraphics[width=1.5in]{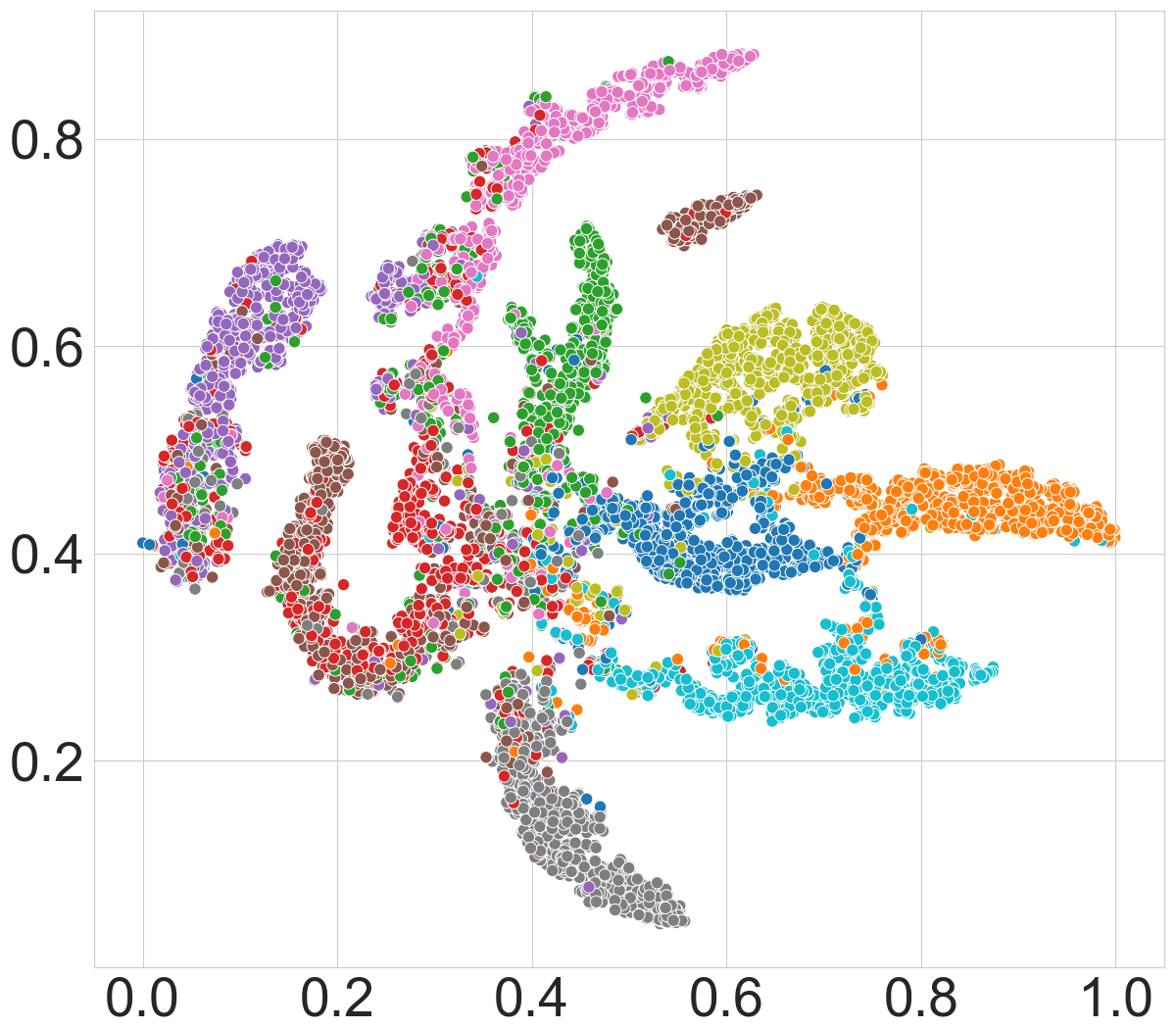}
    }
    
    \caption{Features visualization for CE (top) and CE+SR (bottom) on CIFAR10 with 0.2 symmetric label noise by t-SNE \cite{tsne} 2D embeddings at the last second full-connected layer.}
    \label{fig:0.2}
\end{figure*}

\begin{figure*}[htb]
    \centering
    \subfigure[CE with $\eta=0.8$]{
    \label{fig:CE-cifar10-0.8}
    \includegraphics[width=1.5in]{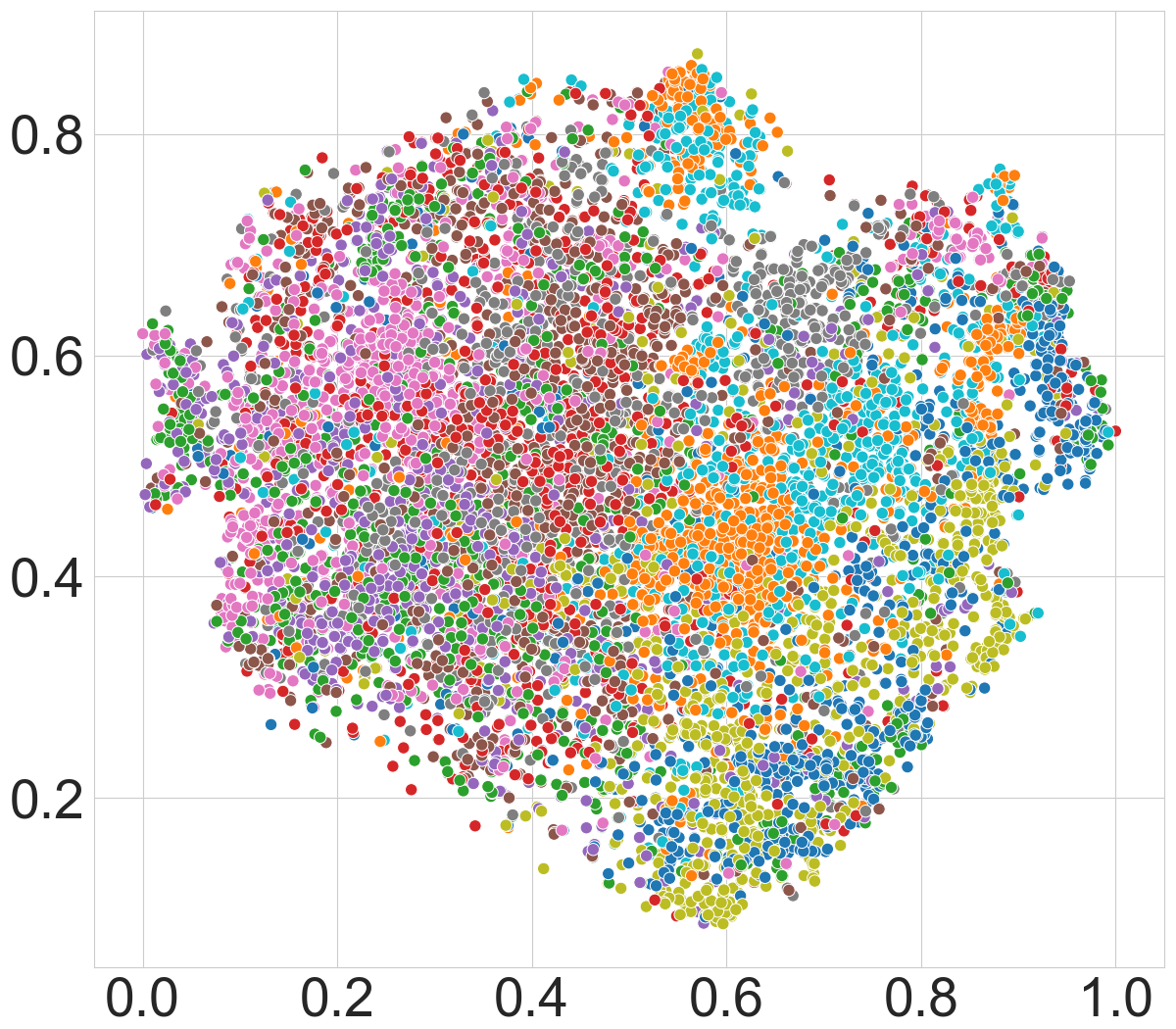}
    }
    \subfigure[FL with $\eta=0.8$]{
    \label{fig:FL-cifar10-0.8}
    \includegraphics[width=1.5in]{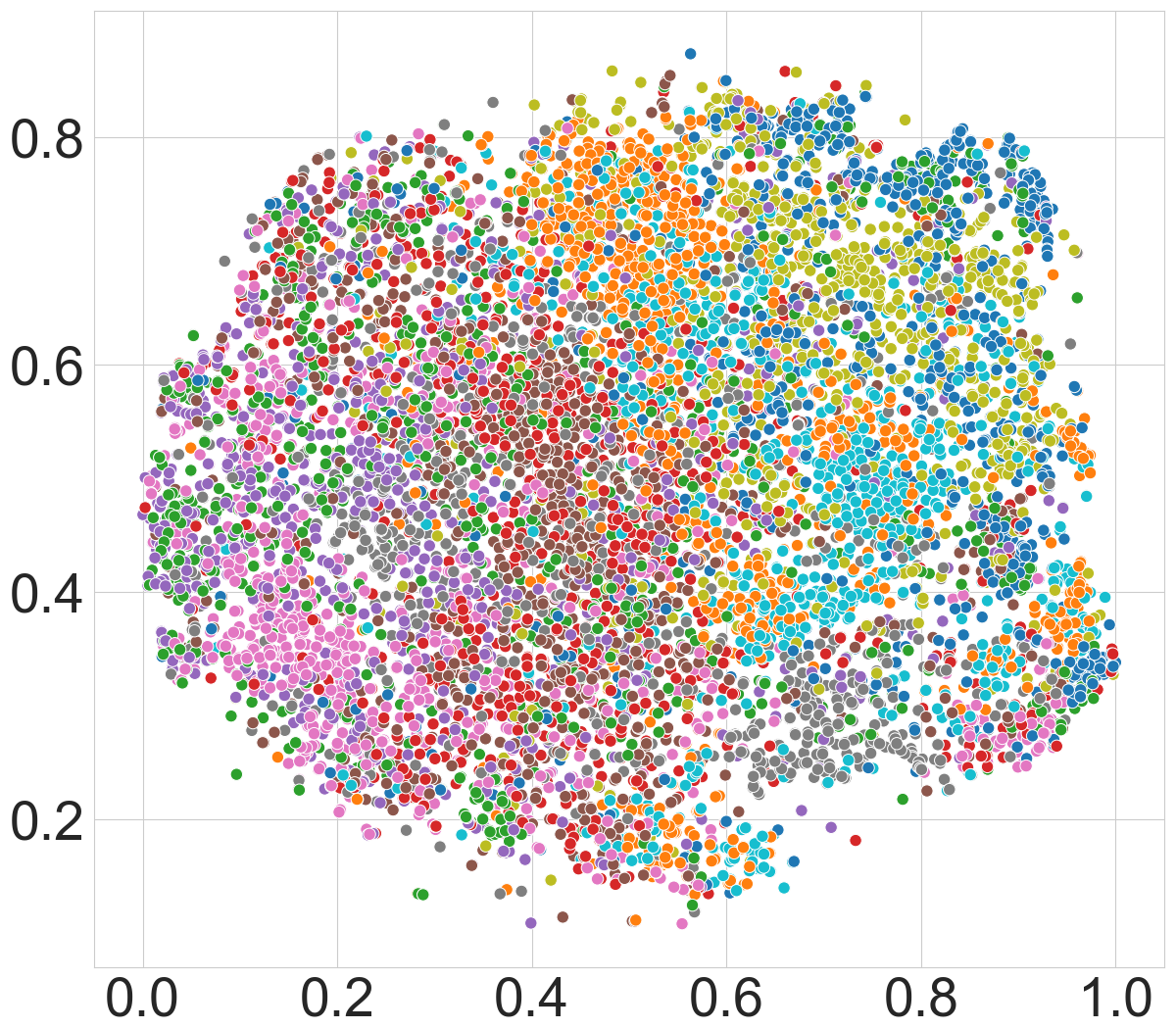}
    }
    \subfigure[GCE with $\eta=0.8$]{
    \label{fig:GCE-cifar10-0.8}
    \includegraphics[width=1.5in]{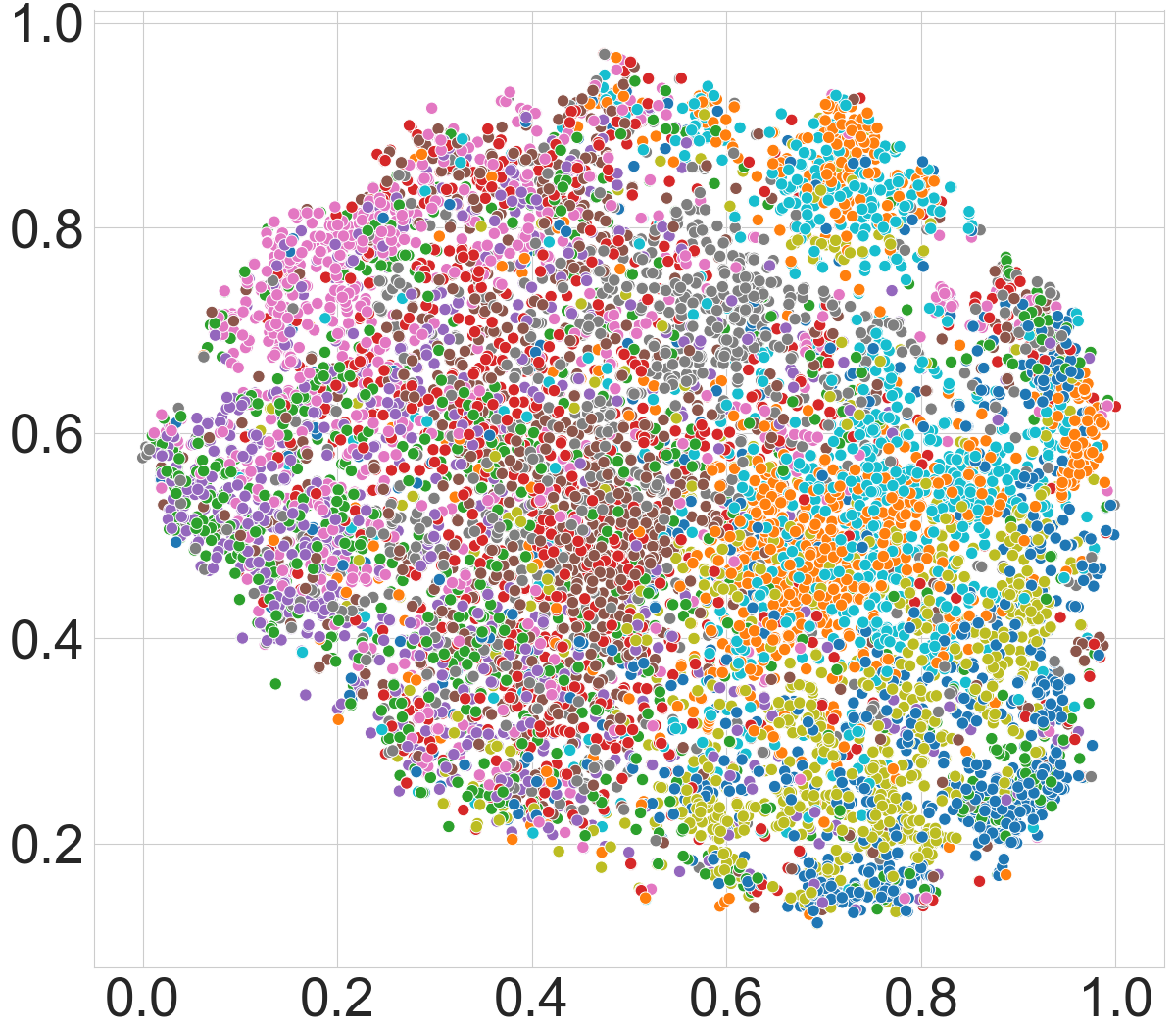}
    }
    \\
    \subfigure[CE+SR with $\eta=0.8$]{
    \label{fig:CESR-cifar10-0.8}
    \includegraphics[width=1.5in]{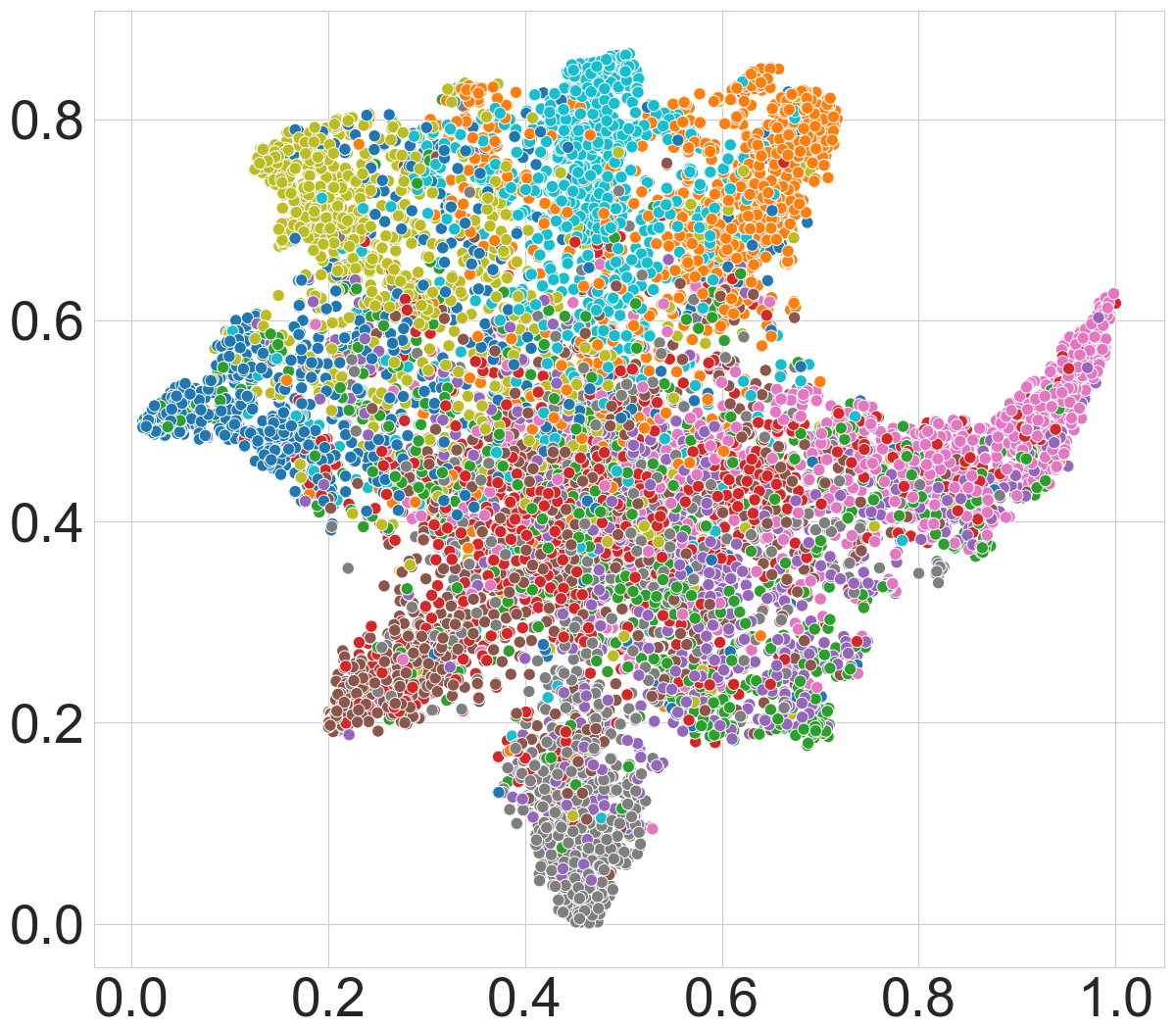}
    }
    \subfigure[FL+SR with $\eta=0.8$]{
    \label{fig:FLSR-cifar10-0.8}
    \includegraphics[width=1.5in]{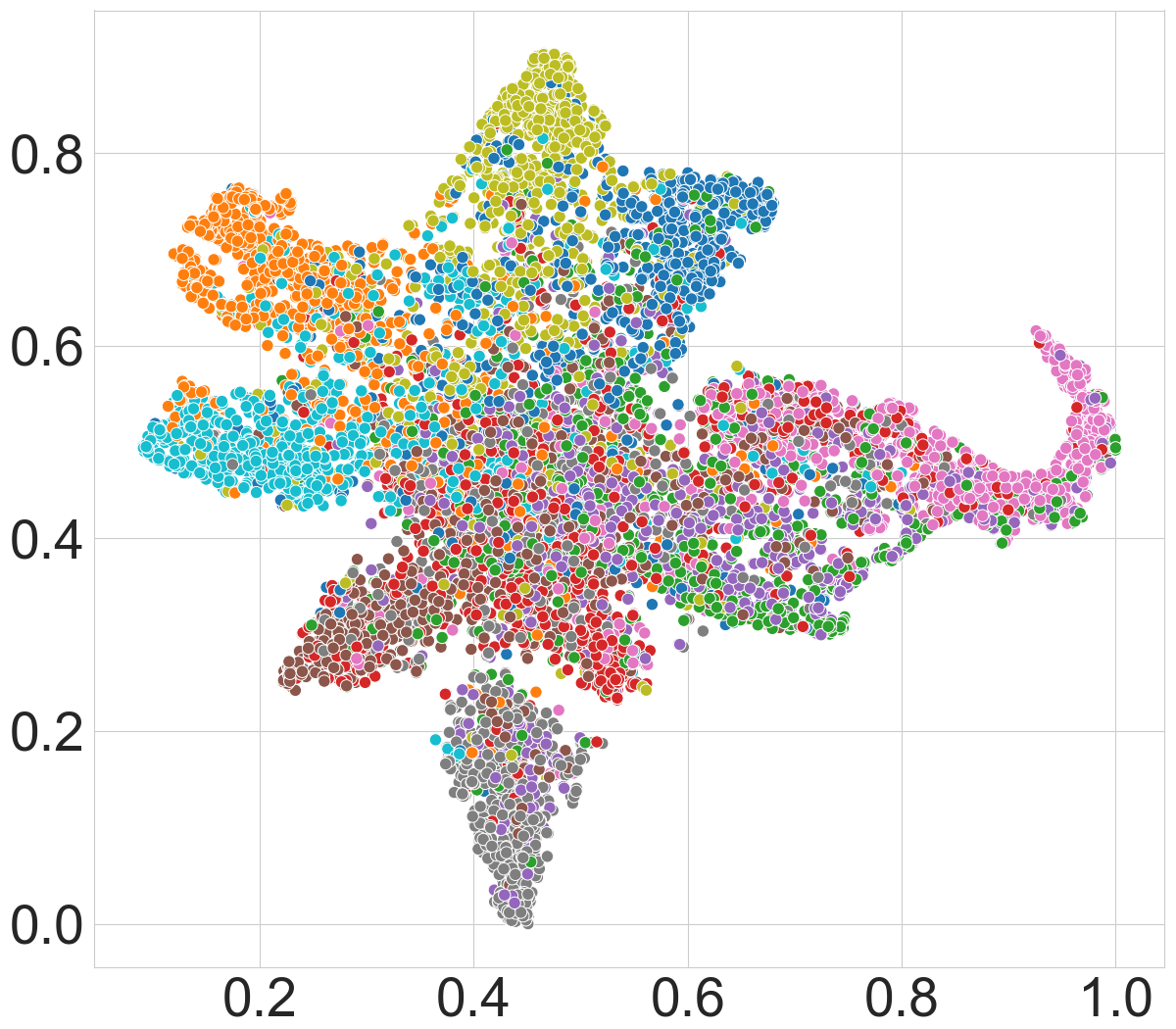}
    }
    \subfigure[GCE+SR with $\eta=0.8$]{
    \label{fig:GCESR-cifar10-0.8}
    \includegraphics[width=1.5in]{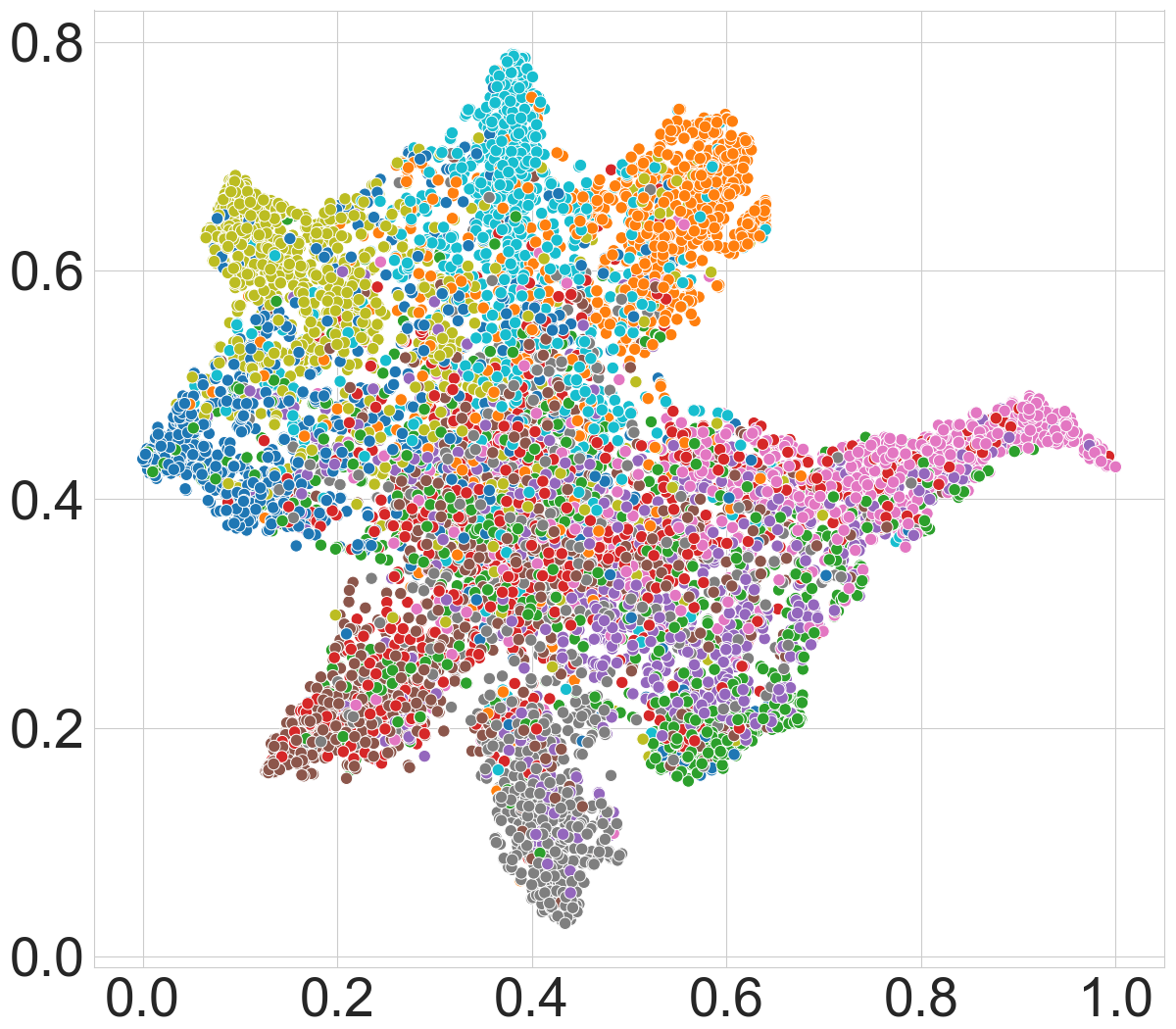}
    }
    
    \caption{Features visualization for CE (top) and CE+SR (bottom) on CIFAR10 with 0.8 symmetric label noise by t-SNE \cite{tsne} 2D embeddings at the last second full-connected layer.}
    \label{fig:0.8}
\end{figure*}

\noindent \textbf{More results of visualizations.}\quad  More visualizations of representations on different datasets are shown in Fig. \ref{fig:mnist-a-tsne}, \ref{fig:0.2} and \ref{fig:0.8}. As can be seen, the representations learned by the proposed sparse regularization (SR)-enhanced methods are more discriminative than those learned by original losses, which are with a more separated and clearly bound margin.

\end{document}